\documentclass[11pt,english,authoryear]{article}
\usepackage[a4paper]{geometry}
\usepackage[export]{adjustbox}
\usepackage[latin9]{inputenc}
\usepackage[T1]{fontenc}
\usepackage{microtype}
\usepackage{mathptmx}
\usepackage{stmaryrd}
\usepackage{stackrel}
\usepackage{booktabs}
\usepackage{makecell}
\usepackage{graphicx}
\usepackage{setspace}
\usepackage{amsmath}
\usepackage{amssymb}
\usepackage{accents}
\usepackage{dcolumn}
\usepackage{diagbox}
\usepackage{amsthm}
\usepackage{subfig}
\usepackage{float}
\usepackage{babel}
\usepackage{tikz}
\usepackage{natbib}

\makeatletter
\def\ps@pprintTitle{
  \let\@oddhead\@empty
  \let\@evenhead\@empty
  \let\@oddfoot\@empty
  \let\@evenfoot\@oddfoot
}
\makeatother

\theoremstyle{definition}
\newtheorem{defn}{Definition}\theoremstyle{plain}
\newtheorem{lem}{Lemma}\theoremstyle{plain}
\newtheorem{thm}{Theorem}\theoremstyle{plain}
\newtheorem{cor}{Corollary}\theoremstyle{plain}
\newtheorem{prop}{Proposition}\theoremstyle{plain}

\floatstyle{ruled}
\newfloat{algorithm}{tbp}{loa}
\providecommand{\algorithmname}{Algorithm}
\floatname{algorithm}{\protect\algorithmname}

\DeclareMathOperator*{\argmin}{argmin}

\newcolumntype{.}{D{.}{.}{-1}}

\begin{document}

\title
{Generalization error minimization: a new approach to model evaluation and selection with an application to penalized regression\footnote{The authors would like to thank Mike Bain, Colin Cameron, Peter Hall and Tsui Shengshang for valuable comments on an earlier draft. We would also like to acknowledge participants at the 12th International Symposium on Econometric Theory and Applications and the 26th New Zealand Econometric Study Group as well as seminar participants at Utah, UNSW, and University of Melbourne for useful questions and comments. Fisher would like to acknowledge the financial support of the Australian Research Council, grant DP0663477.}}

\author
{Ning Xu \\ School of Economics, University of Sydney \\  n.xu@sydney.edu.au
\and
Jian Hong \\ School of Economics, University of Sydney \\  jian.hong@sydney.edu.au
\and
Timothy C.G. Fisher \\ School of Economics, University of Sydney \\  tim.fisher@sydney.edu.au
}

\maketitle

\begin{abstract}%%%%%%%%%%%%%%%%%%%%%%%%%%%%%%%%%%%%%%%%%%%%%%%%%%%%%%%%%%%%%%%%%%%%%%%%%

We study model evaluation and model selection from the perspective of generalization ability (GA): the ability of a model to predict outcomes in new samples from the same population. We believe that GA is one way formally to address concerns about the external validity of a model. The GA of a model estimated on a sample can be measured by its empirical out-of-sample errors, called the generalization errors (GE). We derive upper bounds for the GE, which depend on sample sizes, model complexity and the distribution of the loss function. The upper bounds can be used to evaluate the GA of a model, ex ante. We propose using generalization error minimization (GEM) as a framework for model selection. Using GEM, we are able to unify a big class of penalized regression estimators, including lasso, ridge and bridge, under the same set of assumptions. We establish finite-sample and asymptotic properties (including $\mathcal{L}_2$-consistency) of the GEM estimator for both the $n \geqslant p$ and the $n < p$ cases. We also derive the $\mathcal{L}_2$-distance between the penalized and corresponding unpenalized regression estimates. In practice, GEM can be implemented by validation or cross-validation. We show that the GE bounds can be used for selecting the optimal number of folds in $K$-fold cross-validation. We propose a variant of $R^2$, the $GR^2$, as a measure of GA, which considers both both in-sample and out-of-sample goodness of fit. Simulations are used to demonstrate our key results.


\end{abstract}%%%%%%%%%%%%%%%%%%%%%%%%%%%%%%%%%%%%%%%%%%%%%%%%%%%%%%%%%%%%%%%%%%%%%%%%%
\renewcommand{\baselinestretch}{1.2}
\setcounter{page}{1}

%%%%%%%%%%%%%%%%%%%%%%%%%%%%%%%%%%%%%%%%%%%%%%%%%%%%%%%%%%%%%%%%%%%%%%%%%%%%%%%%%%%%%%%%%%%%%%%%%%%%%%%%%%%
%%%%%%%%%%%%%
%%%%%%%%%%%%%            SECTION 1
%%%%%%%%%%%%%
%%%%%%%%%%%%%%%%%%%%%%%%%%%%%%%%%%%%%%%%%%%%%%%%%%%%%%%%%%%%%%%%%%%%%%%%%%%%%%%%%%%%%%%%%%%%%%%%%%%%%%%%%%%

\section{Introduction}

Traditionally in econometrics, a statistical inference about a population is formed using a model imposed and estimated on a sample. Put another way, the goal of inference is to see whether a model estimated on a sample may be generalized to the population. Deciding whether the estimated model is valid for generalization is referred to as the process of model evaluation. The performance of a model is usually evaluated using the sample data at hand, referred to as the `internal validity' of the model. However, internal validity may not be a useful indicator of model performance in many different scenarios. A perspective of increasing interest in applied econometrics considers the performance of a model on out-of-sample data. In this paper, we focus on the ability of a model estimated from a given sample to fit new samples, referred to as the \textbf{generalization ability} (GA) of the model.

Generalization ability is one aspect of \textbf{external validity}, the extent to which the results from a study generalize to other settings. Researchers evaluating pilot programs and randomized trials in a variety of settings from labour economics to development economics are increasingly focused on the external validity of their findings.\footnote{In the literature, external validity sometimes refers to whether results based on a sample from one population generalize to another population. The concept of GA may be adapted to this interpretation provided the heterogeneity across the two populations is adequately controlled.} For instance, \citet{heckman2007econometric}, \citet{ludwig2011mechanism} and \citet{allcott2012external} discuss the importance of externally valid results in policy and program evaluation. In a similar vein, \citet{guala2005experiments} and \citet{list2011economists} emphasize the importance of external validity in determining whether experiments can offer a robust explanation of causes and effects in the field. While external validity encompasses many issues from experimental design, sample selection, and economic theory, it clearly also raises econometric issues around model estimation and evaluation.

In order to explore the properties of estimators from the perspective of GA, three questions need to be addressed. How do we measure GA? Given a useful measure, what are its properties? Given its properties, how can GA be exploited for estimation? These questions have received only tangential interest in the literature. \citet{roe2009internal} discuss the trade-off between internal and external validity in empirical data, pointing out that while internal validity (or in-sample fit) is well studied, there is much less research on how to control external validity (out-of-sample fit). Some work on external validity has focussed on the properties of specific estimation methods. \citet{angrist2010extrapolate}, for example, studies the external validity of instrument variable estimation in a labor market setting. In this paper, we answer these questions, starting by proposing a measure of GA that is straightforward to implement empirically.

With a new sample at hand, GA is easily measured by using validation or cross-validation to quantify the goodness-of-fit of the estimated model on new (out-of-sample) data. Without a new sample, however, it can be difficult to measure GA \emph{ex ante}. In this paper, when only a single sample is available, we quantify the GA of the in-sample estimates by deriving upper bounds on the empirical out-of-sample errors, which we call the \textbf{empirical generalization errors} (eGE). The upper bounds reveal the properties of the eGE as well as insight into other important statistical properties, such as the trade-off between in-sample and out-of-sample fit in estimation. Furthermore, the upper bounds can be extended to analyze the performance of validation, $K$-fold cross-validation and specific estimators such as penalized regression. Thus, the GA approach yields insight into the finite-sample and asymptotic properties of model estimation and evaluation from several different perspectives.

Essentially, GA measures the performance of a model from an external point-of-view: the greater the GA of an estimator, the better its predictions on out-of-sample data. Furthermore, we show that GA also serves as a natural criterion for model selection, throwing new light on the model selection process. We propose the criterion of minimizing eGE (maximizing GA), or \textbf{generalization error minimization} (\textbf{GEM}), as a framework for model selection. Using penalized regression as an example of a specific model selection scenario, we show how the traditional bias-variance trade-off is connected to GEM and to the trade-off between in-sample and out-of-sample fit. Moreover, the GEM framework allows us to establish additional properties for penalized regression implicit in the bias-variance trade-off.

\subsection{Approaches to model selection}

Given the increasing prevalence of high-dimensional data in economics, model selection is coming to the forefront in empirical work. Researchers often desire a smaller set of predictors in order to gain insight into the most relevant relationships between outcomes and covariates. Without explicitly introducing the concept of GA, the classical approach to model selection focusses on the bias-variance trade-off, yielding methods such as the information criteria (IC), cross-validation, and penalized regression. Consider the standard linear regression model
\[
	Y = X \beta + u
\]
where $Y\in\mathbb{R}^{n}$ is a vector of outcome variables, $X\in\mathbb{R}^{n\times p}$
is a matrix of covariates and $u\in\mathbb{R}^{n}$ is a vector of i.i.d.\ random errors. The parameter vector $\beta\in\mathbb{R}^{p}$ may be sparse in the sense that many of its elements are zero. Model selection typically involves using a score or penalty function that depends on the data \citep{heckerman95}, such as the Akaike information criterion \citep{akaike73}, Bayesian information criterion \citep{schwarz78}, cross-validation errors \citep{stone74,stone77} or the mutual information score among variables \citep{friedman97,friedman00}.

An alternative approach to model selection is penalized regression, implemented through the objective function:
\begin{equation}
	\underset{b_{\lambda}}{\min}~~
	\frac{1}{n} \left(\left\Vert Y - Xb_{\lambda}\right\Vert _{2}\right)^{2}
	+ \lambda\Vert b_{\lambda}\Vert_{\gamma}
\label{penreg}
\end{equation}
where $\Vert\cdot\Vert_{\gamma}$ is the $\mathcal{L}_{\gamma}$-norm and $\lambda\geqslant0$ is a penalty parameter. Note in eq.~(\ref{penreg}) that if $\lambda=0$, the OLS estimator is obtained. The IC can be viewed as special cases with $\lambda=1$ and $\gamma=0$. The lasso \citep{tibshirani96} corresponds to the case with $\gamma=1$ (an $\mathcal{L}_{1}$ penalty). When $\gamma=2$ (an $\mathcal{L}_{2}$ penalty), we have the ridge estimator \citep{hoerlkennard70}. For any $\gamma>1$, we have the bridge estimator \citep{frankfriedman93}, proposed as a generalization of the ridge.

One way to derive the penalized regression estimates $b_{\lambda}$ is using the \textbf{cross-validation} approach, summarized in Algorithm~1. As shown in Algorithm~1, cross-validation solves the constrained minimization problem in eq.~(\ref{penreg}) for each value of the penalty parameter $\lambda$ to derive a $b_{\lambda}$. When the feasible range of $\lambda$ is exhausted, the estimate that produces the smallest out-of-sample error among all the estimated $\{b_{\lambda}\}$ is chosen to be the penalized regression estimate, $b^*$.

\begin{table}[ht]
\caption*{Algorithm 1: Penalized regression estimation under cross-validation}

\begin{tabular}{lp{137mm}}
\toprule
    1. & Set the penalty parameter $\lambda = 0$. \\
    2. & Partition the sample into a training set $T$ and a test set $S$. Standardize all variables (to ensure the penalized regression residual $e$ satisfies $\mathbb{E}(e)=0$ in $T$ and $S$). \\
    3. & Compute the penalized regression estimate $b_{\lambda}$ on $T$. Use $b_{\lambda}$ to calculate the prediction error on $S$. \\
    4. & Increase the penalty parameter $\lambda$ by a preset step size. Repeat 2 and 3 until $b_{\lambda} = \mathbf{0}$. \\
    5. & Select $b^*$ to be the $b_{\lambda}$ that minimizes the prediction error on $S$. \\
\bottomrule
\end{tabular}
\end{table}

A range of consistency properties have been established for the IC and penalized regression. \cite{shao97} proves that various IC and cross-validation are consistent in model selection. \cite{breiman95, chickering04} show that the IC have drawbacks: they tend to select more variables than necessary and are sensitive to small changes in the data. \cite{zhanghuang08, knightfu00, meinshausenbuhlmann06, zhaoyu06} show that $\mathcal{L}_1$-penalized regression is consistent in different settings. \citet{ huangall08, hoerlkennard70} show the consistency of penalized regression with $\gamma > 1$. \cite{zou06, caner09, friedman10} propose variants of penalized regression in different scenarios and \cite{fu98} compares different penalized regressions using a simulation study. Alternative approaches to model selection, such as combinatorial search algorithms may be computationally challenging to implement, especially with high-dimensional data.\footnote{\cite{chickering04} point out that the best subset selection method is unable to deal with a large number of variables, heuristically 30 at most.}

\subsection{Major results and contribution}

A central idea in this paper is that model evaluation and model selection may be re-framed from the perspective of GA. If the objective is to improve the GA of a model, model selection is necessary. Conversely, GA provides a new and elegant angle to understand model selection. By introducing generalization errors as the measure of GA, we connect GEM to the traditional bias-variance trade-off, the trade-off between in-sample and out-of-sample fit and the properties of validation and cross-validation. By the same token, the concept of GA may be used to derive additional theoretical properties of model selection. Specifically for the case of regression analysis, we use GEM to unify the properties for the class of penalized regressions with $\gamma \geqslant 0$, and show that the finite-sample and asymptotic properties of penalized regression are closely related to GA.

The \textbf{first contribution} of this paper is to quantify the GA of a model, \emph{ex ante}, by deriving an upper bound for the GE. The upper bounds depend on the sample size, an index of the complexity of models, a loss function, and the distribution of the underlying population. The upper bound also characterizes the trade-off between in-sample fit and out-of-sample fit. As shown in \citet{vc71,vc71b,shalizi2011,smale2009,hu2009}, the inequalities underlying conventional analysis of GA focus on the relation between the population error and the empirical in-sample error. Conventional methods to improve GA involve computing discrete measures of model complexity, such as the VC dimension, Radamacher dimension or Gaussian complexity, which typically are hard to compute. In contrast, for any out-of-sample data, we quantify bounds for the prediction error of the extremum estimate learned from in-sample data. Furthermore, we show that empirical GA analysis is straightforward to implement via validation or cross-validation and possesses desirable finite-sample and asymptotic properties for model selection.

A \textbf{second contribution} of the paper is to propose GEM as a general framework for model selection and estimation. By re-framing the bias-variance trade-off from the perspective of in-sample and out-of-sample fit, GA is connected to the traditional bias-variance trade-off and model selection. To be specific: model selection is necessary to improve the GA of an estimated model while GA naturally serves as an elegant way to understand the properties of model selection. Thus, a model estimated by GEM will not only achieve the highest GA---and thus some degree of external validity---but also possess a number of nice theoretical properties. GEM is also naturally connected to the properties of cross-validation. As argued by \citet{varian14}, cross-validation could be adopted more often in empirical analysis in economics, especially as big data sets are becoming increasingly available in many fields.

A \textbf{third contribution} of the paper is to use GA analysis to unify a class of penalized regression estimators and derive their finite-sample and asymptotic properties (including $\mathcal{L}_{2}$-consistency) under the same set of assumptions. Various properties of penalized regression estimators have previously been established, such as probabilistic consistency or the oracle property \citep{knightfu00,zhaoyu06,candestao07,meinshausenyu09,bickeletal09}. GA analysis reveals that similar properties can be established more generally and for a wider class of penalized regression estimators. We also show that the $\mathcal{L}_{2}$-difference between the OLS estimate and any penalized regression estimate depends on their respective GAs.

Lastly, a \textbf{fourth contribution} of the paper is to show that GA analysis may be used to tune the hyper-parameter for validation (i.e., the ratio of training sample size to test sample size) or cross-validation (i.e., the number of folds $K$). Existing research has studied cross-validation for the estimation of specific parametric and nonparametric models \citep{hall1991,hall2011,stone74,stone77}. In contrast, by adapting the classical error bound inequalities that follow from our analysis of GA, we derive the optimal tuning parameters for validation and cross-validation in a model-free setting. We also show how $K$ affects the bias-variance trade-off for cross-validation: a higher $K$ increases the variance and lowers the bias.

The paper is organized as follows. In Section~2, we propose empirical generalization errors as the measure of GA and the criterion to evaluate the external performance of the estimated model. Also, by deriving the upper bounds for the empirical generalization errors in different scenarios, we reveal important features of the extremum estimator such as the trade-off between in-sample and out-of-sample fit. Our approach also offers a framework to study the properties of validation and cross-validation. In Section~3, we apply the GEM framework as a model selection criterion for penalized regression. We also demonstrate a number of new properties for penalized regression that flow from the GEM framework. We prove the $\mathcal{L}_{2}$-consistency of penalized regression estimators for both $p\leqslant n$ and $p>n$ cases. Further, we establish the finite-sample upper bound for the $\mathcal{L}_{2}$-difference between penalized and unpenalized estimators based on their respective GAs. In Section~4, we use simulations to demonstrate the ability of penalized regression to control for overfitting. We also propose a measure of overfitting based on the empirical generalization errors, called the generalized $R^2$ or $GR^{2}$. Section~5 concludes with a brief discussion of our results. Proofs (showing detailed steps) are contained in Appendix~A and summary plots of the simulations are in Appendix~B.

%%%%%%%%%%%%%%%%%%%%%%%%%%%%%%%%%%%%%%%%%%%%%%%%%%%%%%%%%%%%%%%%%%%%%%%%%%%%%%%%%%%%%%%%%%%%%%%%%%%%%%%%%%%
%%%%%%%%%%%%%
%%%%%%%%%%%%%            SECTION 2
%%%%%%%%%%%%%
%%%%%%%%%%%%%%%%%%%%%%%%%%%%%%%%%%%%%%%%%%%%%%%%%%%%%%%%%%%%%%%%%%%%%%%%%%%%%%%%%%%%%%%%%%%%%%%%%%%%%%%%%%%

\section{Generalization ability and the upper bound for the finite-sample generalization error}

In this section, we propose the eGE, the measurement of GA for a model, as a new angle to evaluate the `goodness' of a model. By deriving the upper bounds of eGE of the extremum estimator, we show that the upper bounds of eGE directly quantifies how much the model overfits or underfits the finite samples, also the upper bounds of eGE may be used to study the properties of eGE for a model. Furthermore, these upper bounds can be used to study the properties of cross validation and validation on finite samples. All those result implies that eGE is a convenient and useful criterion for model evaluation.

\subsection{Generalization ability, generalization error and overfitting}

In econometrics, choosing the best approximation to data often involves measuring a loss function, $Q(b\vert y_i,\mathbf{x}_i)$, defined as a functional that depends on some estimate $b$ and the sample points $(y_i,\mathbf{x}_i)$. The population error (or risk) functional is defined as
\[
	\mathcal{R}(b\vert Y,X)=\int Q(b\vert y,\mathbf{x})\mathrm{d}F(y,\mathbf{x})
\]
where $F(y,\mathbf{x})$ is the joint distribution of $y$ and $\mathbf{x}$. Without knowing the distribution $F(y,\mathbf{x})$ a priori, given a random sample $(Y,X)$, we define the empirical error functional as follows
\[
	\mathcal{R}_{n}(b\vert Y,X) = \frac{1}{n}\;\sum_{i=1}^n\;Q(b\vert y_i,\mathbf{x}_i).
\]
For example, in the regression case, $b$ is the estimated parameter vector and $\mathcal{R}_{n}(b\vert Y,X)=\frac{1}{n}\sum_{i=1}^n(y_i - \mathbf{x}_i^T b)^{2}$.

When estimation involves minimizing the in-sample empirical error, we have the extremum estimator \citep{amemiya1985advanced}. In many settings, however, minimizing the in-sample empirical error does not guarantee a reliable model. In regression, for example, often the $R^2$ is used to measure goodness-of-fit for the in-sample data. However, an estimate with a high in-sample $R^{2}$ may fit out-of-sample data poorly, a feature commonly referred to as \textbf{overfitting}: the in-sample estimate is too tailored for the sample data, compromising its out-of-sample performance.\footnote{Likewise, if the model fits in-sample data poorly and hence compromise its out-of-sample performance, we say that the model \textbf{underfits} the data.}  As a result, in-sample fit (internal validity) of the model may not be a reliable indicator of the general applicability (external validity) of the model.

Thus, \citet{vc71} refer to the \textbf{generalization ability} (GA) of a model: a measure of how an extremum estimator performs on out-of-sample data. GA can be measured several different ways. In the case where $Y$ and $X$ are directly observed, GA is a function of the difference between the actual and predicted $Y$ for out-of-sample data. In this paper, GA is measured by the out-of-sample empirical error functional.
\begin{defn}
[Training set, test set, empirical training error and empirical generalization error]
\label{defn2.1}
\end{defn}

	\begin{enumerate}

    \item   Let $(y_i, \mathbf{x}_i)$ denote a sample point from $F(y, \mathbf{x})$, the joint distribution of $(y,\mathbf{x})$. Let $\Lambda$ denote the space of all models. The \textbf{loss function} for $b\in\Lambda$ is $Q(b\vert y_i,\mathbf{x}_i),\,i=1,\ldots,n$. The \textbf{population error functional} for $b\in\Lambda$ is $\mathcal{R}(b\vert Y,X)=\int Q(b\vert y,\mathbf{x})\mathrm{d}F(y,\mathbf{x})$. The \textbf{empirical error functional} is $\mathcal{R}_{n}(b\vert Y,X) =\frac{1}{n}\;\sum_{i=1}^n\;Q(b\vert y_i,\mathbf{x}_i)$.

    \item   Given a sample $(Y,X)$, the \textbf{training set} $(Y_{t},\,X_{t})\in\mathbb{R}^{n_t \times p}$ refers to data used to estimate $b$ (i.e., the in-sample data) and the \textbf{test set} $(Y_{s},\,X_{s})\in\mathbb{R}^{n_s \times p}$ refers to data \emph{not} used to estimate $b$ (i.e., the out-of-sample data). Let $\widetilde{n}=\min\{n_s,n_t\}$. The \textbf{effective sample size} for the training set, test set and the total sample, respectively, is $n_t/p$, $n_s/p$ and $n/p$.

    \item   Let $b_{train}\in\Lambda$ denote an extremum estimator, where $b_{train}$ minimizes $\mathcal{R}_{n_t}(b|Y_{t},X_{t})$. Under \textbf{validation}, the \textbf{empirical training error (eTE)} for $b_{train}$ is $\mathcal{R}_{n_t}(b_{train}|Y_{t},X_{t})$, the \textbf{empirical generalization error (eGE)} is $\mathcal{R}_{n_s}(b_{train}|Y_{s},X_{s})$ and the \textbf{population error} is $\mathcal{R}(b_{train}|Y,X)$.
		
    \item   For $K$\textbf{-fold cross-validation}, denote the training set and test set in the $q$th round, respectively, as $(Y_{t}^q,X_{t}^q)$ and $(Y_s^q,X_s^q)$. In each round, the sample size for the training set is $n_t = n(K-1)/K$ and the sample size for the test set is $n_s = n/K$. Thus, $\mathcal{R}_{n_s}(b_{train}|Y_{s}^q,X_{s}^q)$ is the eGE and $\mathcal{R}_{n_t}(b_{train}|Y_{t}^q,X_{t}^q)$ is the eTE, respectively, in the $q$th round of cross-validation.

	\end{enumerate}

Two methods are typically used to compute the eGE of an estimate: validation and cross-validation. Under the validation approach, the sample is randomly divided into a training set and a test set. Following estimation on the training set, the fitted model is applied to the test set to compute the validated eGE. $K$-fold cross-validation may be thought of as `multiple-round validation'. Under the cross-validation approach, the sample is randomly divided into $K$ subsamples or folds.\footnote{In practice, researchers arbitrarily choose $K = 5$, 10, 20, 40 or $n$.} One fold is chosen to be the test set and the remaining $K-1$ folds comprise the training set. Following estimation on the training set, the fitted model is applied to the test set to compute the eGE. The process is repeated $K$ times, with each of the $K$ folds in turn taking the role of the test set while the remaining $K-1$ folds are used as the training set. In this way, $K$ different estimates of the eGE for the fitted model are obtained. The average of the $K$ eGEs yields the cross-validated eGE.

Cross-validation uses each data point in both the training and test sets. The method also reduces resampling error by running the validation $K$ times over different training and test sets. Intuitively this suggests that cross-validation is more robust to resampling error and on average will perform at least as well as validation. In Section~3, we study the generalization ability of penalized extremum estimators in both the validation and cross-validation cases.

To study the properties of eGE for a model, three assumptions are required for the analysis in this section of the paper, stated as follows.

\bigskip
\noindent
\textbf{Assumptions}

\begin{enumerate}

    \item[\textbf{A1.}]   In the probability space $(\Omega,\mathcal{F},P)$, we assume $\mathcal{F}$-measurability of the loss function $Q(b\vert y, \mathbf{x})$, the population error $\mathcal{R}(b|Y,X)$ and the empirical error $\mathcal{R}_{n}(b\vert Y,X)$, for any $b\in\Lambda$ and any sample point $(y_i, \mathbf{x}_i)$. Distributions of loss functions have closed-form, first-order moments.

    \item[\textbf{A2.}]   The sample $\{(y_i, \mathbf{x}_i)\}_{i=1}^n$ is randomly drawn from the population. In cases with multiple random samples, both the training set and the test set are randomly sampled from the population. In cases with a single random sample, both the training set and the test set are randomly partitioned from the sample.

    \item[\textbf{A3.}]   For any sample, the extremum estimator $b_{train}\in\Lambda$ exists. The in-sample error for $b_{train}$ converges in probability to the minimal population error as $n\rightarrow\infty$.

\end{enumerate}

A few comments are in order for assumptions A1--A3. The loss distribution assumption A1 is made to simplify the analysis. The existence and convergence assumption A3 is standard (see, for example, \citet{neweymcfadden94}).

The independence assumption A2 is not essential; GA analysis is valid for both i.i.d.\ and non-i.i.d.\ data. While the original research on GA in \citet{vc74, vc74b} imposes the i.i.d.\ restriction, subsequent work has generalized their results to cases where the data are dependent or not identically distributed.\footnote{See, for example, \citet{yu1994rates, cesa2004generalization, smale2009, mohri2009rademacher, kakade2009generalization, shalizi2011}.} Others have shown that if heterogeneity is due to an observed random variable,\footnote{See \citet{yashin1986, skrondal2004generalized, wang2005learning, yu2009learning, pearl2015detecting}.} the variable may be added to the model to control for the heterogeneity, while if the heterogeneity is related to a latent variable, various approaches---such as the hidden Markov model, mixture modelling or factor modelling---are available for heterogeneity control. In this paper, due to the different measure-theory setting for dependent data, we focus on the independent case as a first step.

Given A1--A3, both the eTE and eGE converge to the population error:
\[
\mathrm{lim}_{\widetilde{n}\rightarrow\infty}\mathcal{R}_{n_t}(b_{train}|Y_{t},X_{t}) = \mathrm{lim}_{\widetilde{n}\rightarrow\infty}\mathcal{R}_{n_s}(b_{train}|Y_{s},X_{s}) = \mathcal{R}(b_{train}|Y,X).
\]

\subsection{The upper bound of the empirical generalization error}

The problem of GA and overfitting is discovered many years ago. Originally, to improve the generalization ability of a model, \citet{vc71,vc71b} propose minimizing the upper bound of the population error of the estimator $\mathcal{R}(b|Y,X)$ as opposed to minimizing the eTE. The balance between in-sample fit and out-of-sample fit is formulated by \citet{vc74b} using the Glivenko-Cantelli theorem and Donsker's theorem for empirical processes. Specifically, the so-called \textbf{VC inequality} \citep{vc74b} summarizes the relationship between $\mathcal{R}(b|Y,X)$ and $\mathcal{R}_{n_t}(b|Y,X)$.
%
%%%%%%%%%%%%%%%%%%%%%%%%%%%%%%%%%%%%%%%%%%%%%%%%%%%%%%%%%%%%%%%%%%%%%%%%%%%%%%%%%%%%%%%%%%%%
% Lemma 2.1
%%%%%%%%%%%%%%%%%%%%%%%%%%%%%%%%%%%%%%%%%%%%%%%%%%%%%%%%%%%%%%%%%%%%%%%%%%%%%%%%%%%%%%%%%%%%
%
\begin{lem}[VC inequality: the upper bound for the population error]
    Under A1--A3, the following inequality holds with probability at least $1-\eta$, $\forall b_{train} \in\Lambda$, and $\forall n\in\mathbb{N}^{+}$,
	\begin{equation}
		\mathcal{R}(b_{train}|Y,X) \leqslant\mathcal{R}_{n_t}(b_{train}|Y_{t},X_{t})
		+\frac{\sqrt{\epsilon}}{1-\sqrt{\epsilon}}\;\mathcal{R}_{n_t}(b_{train}|Y_{t},X_{t})
		\label{eq:lem2.1}
	\end{equation}
    where $\mathcal{R}(b_{train}|Y,X)$ is the population error, $\mathcal{R}_{n_t}(b_{train}|Y_t,X_t)$ is the empirical training error, $\epsilon=(1/n_t)[h\ln(n_t/h)+h-\ln\left(\eta\right)]$,
    and $h$ is the \textbf{VC dimension}.
\label{lem2.1}
\end{lem}

The VC dimension $h$ is a more general measure of the geometric complexity of a model than the number of parameters, $p$, which does not readily extend as a measure of complexity in nonlinear or non-nested models. While $h$ reduces to $p$ directly for generalized linear models, $h$ can also be used to order the complexity of nonlinear or non-nested models.\footnote{In empirical processes, several other geometric complexity measures are connected to or derived from the VC dimension, such as the minimum description length (MDL) score, the Rademacher dimension (or complexity), Pollard's pseudo-dimension and the Natarajan dimension. Most of these measures, like the VC dimension, are derived from the Glivenko-Cantelli class of empirical processes.} Thus, eq.~(\ref{eq:lem2.1}) can be used as a tool for linear, nonlinear and non-nested model selection.

While Lemma~\ref{lem2.1} is established under A2, eq.~(\ref{eq:lem2.1}) can be generalized to non-i.i.d.\ cases. \citet{shalizi2011} generalizes the VC inequality for $\alpha$- and $\beta$-mixing stationery time series while \citet{smale2009} generalizes the VC inequality for panel data. Moreover, a number of papers \citep{yashin1986, skrondal2004generalized, wang2005learning, yu2009learning, pearl2015detecting} show that heterogeneity can be controlled in the context of the VC inequality by implementing the latent variable model or by adding the variable causing heterogeneity into the model.

\begin{figure}
	\centering
    {\includegraphics[width=0.5\paperwidth]{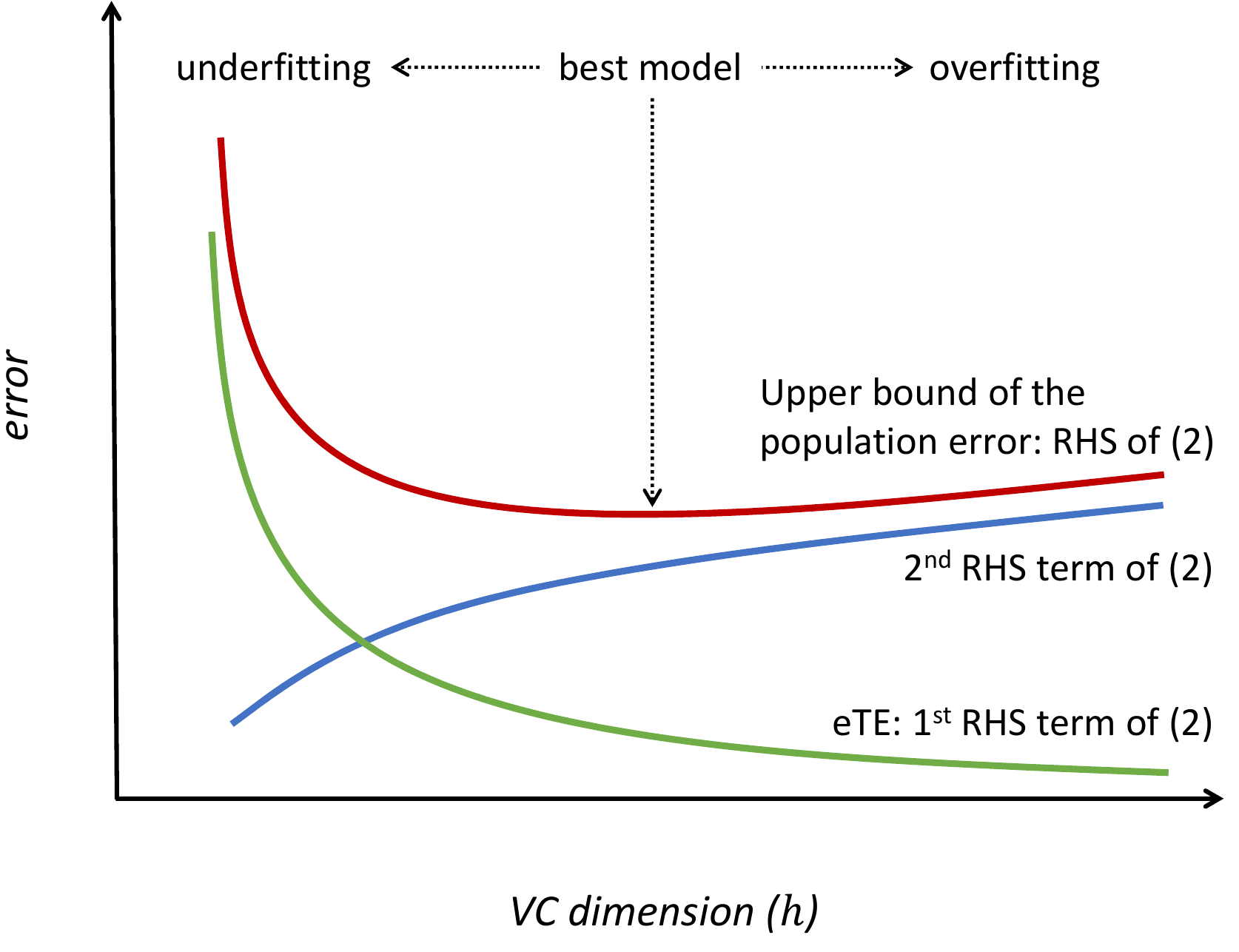}}

	\caption{\label{fig:VCineq}The VC inequality and the upper bound of the population error}

\end{figure}

The VC inequality provides an upper bound for the population error based on the eTE and the VC dimension of the model. As shown in Figure~\ref{fig:VCineq}, when the VC dimension is low (i.e., the model is simple), the effective sample size ($n_t/h$) of the training set is large, $\epsilon$ is small, the second term on the RHS of (\ref{eq:lem2.1}) is small, and the eTE is close to the population error. In this case the extremum estimator has a good GA. However, when the VC dimension is high (i.e., the model is very complicated), the effective sample size $n_t/h$ is small, the second term on the RHS of (\ref{eq:lem2.1}) becomes larger. In such situations, a small eTE does not guarantee a good GA, and overfitting is more likely.

\citet{vc71} show that minimizing the RHS of eq.\ (\ref{eq:lem2.1}) reduces overfitting and improves the GA of the extremum estimator. However, this can be hard to implement because the VC dimension is hard to calculate for anything other than linear models.\footnote{ For example, VC dimension is complicated and not handy to implement if we try compare the GA between two different distribution in kernel density estimation.} Also, VC inequality is for population. But in practice population is often not accessible and the data we collect is always finite, moreover we are often interesting in how a estimated model would perform on finite out-of-sample data, such as "how well the effect of a policy, estimated from a sample from suburb A, would explain the variation in suburb B, given relevant heterogeneity controlled. " Hence, as a handy and intuitive measurement for GA or overfitting in empirical study, eGE has been proposed and widely used in applications.

As a reasonable criterion for model evaluation, we would expect that eGE should demonstrate a number of nice properties that characterize its pattern or regularity as an empirical process. For example, intuitively if we get an estimated model from the training set and apply it to many test sets with same size, we would expect that value of different eGEs could be as stable as possible despite of the sampling error; also, if we calculate the eGE of a model on test sets with different sizes, we would expect that value of eGEs should be influenced by the size of training set, the size of test set and the tail behavior of the error distribution.  However, due to the procedure of sampling from population and subsampling by validation/cross validation, the randomness of eGE as a empirical process increase the difficulty to derive those properties. To reduce and absorb the influence of sampling and subsampling, one approach is to derive an upper bound for eGE,\footnote{All error functional is non-negative, hence $0$ could be seen as a natural lower bound. Moreover, typically the consequence of a large eGE is more severe than a small eGE. As a result, we focus on the upper bound here.} which is somehow similar to what eq.~(\ref{eq:lem2.1}) does to population error, Thus, we now describe the finite-sample relation between the eGE and the eTE  in validation, by adapting eq.~(\ref{eq:lem2.1}).
%
%%%%%%%%%%%%%%%%%%%%%%%%%%%%%%%%%%%%%%%%%%%%%%%%%%%%%%%%%%%%%%%%%%%%%%%%%%%%%%%%%%%%%%%%%%%%
% Theorem 2.1
%%%%%%%%%%%%%%%%%%%%%%%%%%%%%%%%%%%%%%%%%%%%%%%%%%%%%%%%%%%%%%%%%%%%%%%%%%%%%%%%%%%%%%%%%%%%
%
\begin{thm}[The upper bound for the eGE of the extremum estimator under validation]
    Under A1--A3 and given an extremum estimator $b_{train} \in \Lambda$, the following upper bound for the eGE holds with probability at least $\varpi(1-1/n_t)$, $\forall\varpi\in\left(0,1\right)$.
	\begin{equation}
		\mathcal{R}_{n_s}(b_{train}|Y_{s},X_{s}) \leqslant
		\frac{\mathcal{R}_{n_t}(b_{train}|Y_{t},X_{t})}{(1-\sqrt{\epsilon})} + \varsigma,
		\label{eq:thm2.1}
	\end{equation}
    where $\mathcal{R}_{n_s}(b_{train}|Y_{s},X_{s})$ is the eGE and $\mathcal{R}_{n_t}(b_{train}|Y_t,X_t)$ the eTE, $\epsilon$ is defined in Lemma~\ref{lem2.1}, $\varsigma = B\:\ln\sqrt{2/(1-\varpi)}/n_s$ if $Q(\cdot) \in(0,B]$ and $B$ is bounded, and otherwise
	\[
		\varsigma =
			\left\{
				\begin{array}{ll}
				   \mathrm{var}[Q(b_{train}\vert y, \mathbf{x})] / (n_s (1 - \varpi))
					& \mbox{if } \nu\in(2,\infty) \\[12pt]
                 \sqrt[\nu]{2} \tau
                \left(\mathbb{E}\left[Q(b_{train}|Y_{s},X_{s})\right]\right)/
                (n_s^{1-1/{\nu}}\sqrt[\nu]{1-\varpi})
					& \mbox{if } \nu\in(1,2]
               \end{array}
            \right.
	\]
    where
	\[
		\tau \geqslant \sup
        \frac{[\int\left(Q(b|y,\mathbf{x})\right)^{\nu} \mathrm{d}F(y,\mathbf{x})]^{1/{\nu}}}
		     {\int Q(b|y,\mathbf{x}) \mathrm{d}F(y,\mathbf{x})}.
	\]

\label{thm2.1}
\end{thm}

\bigskip
Eq.~(\ref{eq:thm2.1}) provides an upper bound for the eGE of the test set as it depends on the eTE of the model with a given $h$ estimated on the training set. Thus, eq.~(\ref{eq:thm2.1}) shows how the eTE from extremum estimation on the training set may be used to compute the eGE of the model on the test set. In other words, eq.~(\ref{eq:thm2.1}) measures the GA of a model of a given complexity under the validation approach.

\begin{figure}
    \centering
	{\includegraphics[width=0.4\paperwidth]{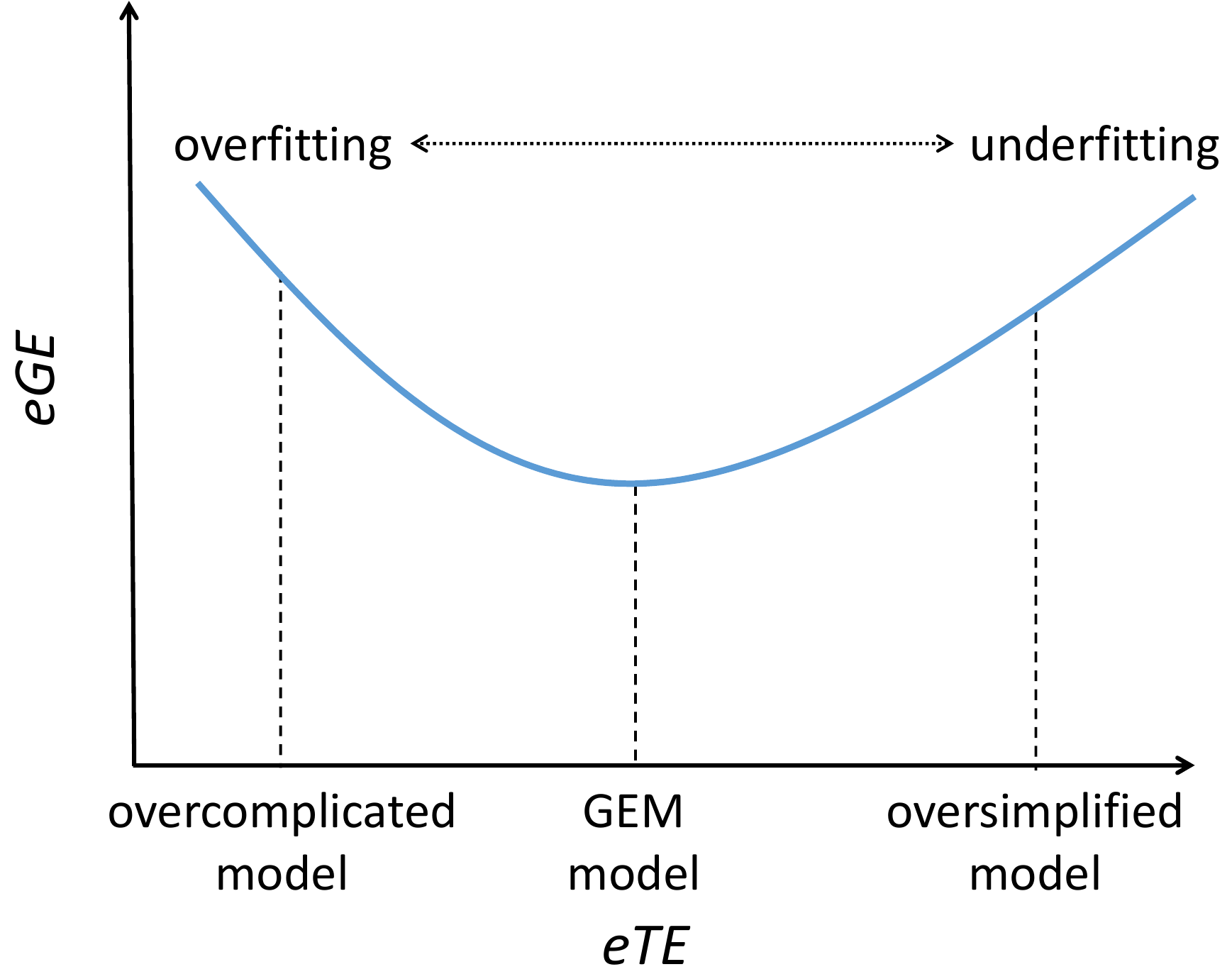}}

	\caption{\label{fig:TEGE}Representation of the trade-off between eGE and eTE}

\end{figure}

Theorem~\ref{thm2.1} has several important implications.
\begin{enumerate}

    \item   \emph{Upper bound of the eGE.} Eq.~(\ref{eq:thm2.1}) establishes the upper bound of the eGE for any out-of-sample data of size $n_s$ based on the eTE from any in-sample data of size $n_t$. Thus, eq.~(\ref{eq:thm2.1}) quantifies the upper bound of the eGE, as opposed to Lemma~\ref{lem2.1}, which quantifies the upper bound of the population error. Previously, measuring the eGE of a model with new data required the use of validation or cross-validation. Given Theorem~\ref{thm2.1}, the eGE may be quantified directly using the RHS of eq.~(\ref{eq:thm2.1}), avoiding the need for validation.

    \item   \emph{The trade-off between accuracy and looseness of the upper bound.} Theorem~(\ref{thm2.1}) also shows that $\varpi \in \left[ 0,1 \right]$ influences the trade-off between the accuracy and efficiency of the upper bound. The higher $\varpi$, the more likely the upper bound holds and the larger $\varsigma$. Thus, while the probability the bound holds increases, it comes at the cost of a looser upper bound. In contrast, a lower $\varpi$ reduces the upper bound, offering a empirically efficient upper bound at the cost of reducing the probability that it holds.

    \item   \emph{The eGE-eTE trade-off in model selection.} Eq.~(\ref{eq:thm2.1}) also characterizes the trade-off between eGE and eTE for model selection in both the finite-sample and asymptotic cases. The population GE and the population TE converge to the population error. Hence, minimizing eTE can lead directly to the true DGP in the population. By contrast, for the finite-sample case illustrated in Figure~\ref{fig:TEGE}, while an overcomplicated model (a low $n_t/h$) will have a small eTE, eq.~(\ref{eq:thm2.1}) shows it may result in a large eGE on new data. Thus, an overcomplicated model will tend to \textit{overfit} the in-sample data and have poor GA. On the other hand, an oversimplified model (a high $n_t/h$) will be unlikely to recover the true DGP leading to a large upper bound for the eGE. Thus, an oversimplified model will tend to \textit{underfit}, fitting both the in-sample and out-of-sample data poorly. Thus, the complexity of a model is related to a trade-off between the eTE and eGE.

    \item   \emph{GA and tails of the loss function distribution.} Eq.~(\ref{eq:thm2.1}) also shows how the tail of the loss function distribution affects the upper bound of the eGE through $\varsigma$, the second term on the RHS of eq.~(\ref{eq:thm2.1}). If $Q(\cdot)$ is bounded or light-tailed, $\varsigma$ is mathematically simple and converges to zero at the rate $1/n_s$. If the loss function is heavy-tailed and $\mathcal{F}$-measurable, $\nu$, the highest order of the population moment that is closed-form for the loss distribution,\footnote{It is closed-form owing to A1, which guarantees closed-form, first-order moments for all loss distributions.} can be used to measure the heaviness of the loss distribution tail, a smaller $\nu$ implying a heavier tail. In the case of a heavy tail, $\varsigma$ is mathematically complicated and its convergence rate decreases to $1/n_{s}^{1-1/\nu}$. Hence, the heavier the tail of the loss distribution, the higher the upper bound of the eGE and the harder it is to control GA in finite samples. In the extreme case with $\nu = 1$, there is no way to adapt eq.~(\ref{eq:thm2.1}).

\end{enumerate}

\bigskip
Our next step is to establish a similar bound to eq.~(\ref{eq:thm2.1}) for $K$-fold cross-validation. Given that $K$-fold cross-validation is simply the multiple-round validation, a similar bound to eq.~(\ref{eq:thm2.1}) can be established for cross-validation by convolution. For convenience, define the empirical process
\[
T_q = \mathcal{R}_{n_s}(b_{train}|Y_{s}^q,X_{s}^q) - \frac{\mathcal{R}_{n_t}(b_{train}|Y_{t}^q,X_{t}^q)}
                       {1-\sqrt{\epsilon}},\quad \forall q \in \left[1,K\right]
\]
as the \textbf{eGE gap} in the $q$th round of cross-validation.

%
%%%%%%%%%%%%%%%%%%%%%%%%%%%%%%%%%%%%%%%%%%%%%%%%%%%%%%%%%%%%%%%%%%%%%%%%%%%%%%%%%%%%%%%%%%%%
% Theorem 2.2
%%%%%%%%%%%%%%%%%%%%%%%%%%%%%%%%%%%%%%%%%%%%%%%%%%%%%%%%%%%%%%%%%%%%%%%%%%%%%%%%%%%%%%%%%%%%
%
\begin{thm}[The upper bound for the eGE of the extremum estimator for cross-validation]
Under A1--A3, given an extremum estimator $b_{train} \in \Lambda$ and given $T_q$, then

        \begin{description}

        \item[(i)] if $\: \mathbb{E}(\vert T_q \vert^m) \leqslant m!B^{m-2}\mathrm{var}(T_q)/2,\;\forall m \geqslant 2$, the following upper bound for the eGE gap holds with probability at least $\alpha$

               \begin{eqnarray}
                       \frac{1}{K}\sum_{q=1}^{K}
                       \mathcal{R}_{n_s}(b_{train}|Y_{s}^q,X_{s}^q)
                       \leqslant
               \frac{\frac{1}{K}\sum_{q=1}^{K}\mathcal{R}_{n_t}(b_{train}|Y_{t}^q,X_{t}^q)}
                       {1-\sqrt{\epsilon}} + \varsigma,
                       \label{eq1:thm2.2}
               \end{eqnarray}
               where
               \begin{eqnarray}
                       \alpha & = & 1 - 2\exp\left\{-\frac{1}{2}
                       \frac{(\varsigma - \mathbb{E}[T_q])^2}
                       {\mathrm{var}(T_q)/K + B(\varsigma - \mathbb{E}[T_q])/(3K)}\right\},
                       \notag \\
                       \varsigma & = & \frac{\mathrm{var}[Q(b_{train}\vert y,\mathbf{x})]}
                       {(1-\varpi)n/K},  \notag
               \end{eqnarray}

        \item[(ii)] if $T_q$ is heavy-tailed and sub-exponential, the following upper bound for the eGE holds with probability at least $\alpha$

               \begin{eqnarray}
                       \frac{1}{K}\sum_{q=1}^{K}
                       \mathcal{R}_{n_s}(b_{train}|Y_{s}^q,X_{s}^q)
                       \leqslant
               \frac{\frac{1}{K}\sum_{q=1}^{K}\mathcal{R}_{n_t}(b_{train}|Y_{t}^q,X_{t}^q)}
                       {1-\sqrt{\epsilon}} + \varsigma,
                       \label{eq2:thm2.2}
               \end{eqnarray}
               where
               \begin{eqnarray}
                       \varsigma & = & \sqrt[\nu]{2} \tau
                        \frac{\mathbb{E}\left[Q(b_{train}|Y_{s},X_{s})\right]}
               {\sqrt[\nu]{1-\varpi}\; (n/K)^{1-1/{\nu}}},
              \notag \\
             \alpha & = & \left(1 - 2\tau^\nu\cdot
                       \frac{\left(\mathbb{E}\left[Q(b_{train}\vert y,\mathbf{x})\right]\right)^\nu}
                       {\varsigma^\nu\cdot n^{\nu}}-K/n_t \right)^+,
                       \notag
               \end{eqnarray}
        $\mathcal{R}_{n_s}(b_{train}|Y_{s}^q,X_{s}^q)$ is the eGE and $\mathcal{R}_{n_t}(b_{train}|Y_{t}^q,X_{t}^q)$ is the eTE, respectively, in the qth round of cross-validation.
        \end{description}

\label{thm2.2}
\end{thm}

Theorem~\ref{thm2.2} provides an upper bound for the average eGE from $K$ rounds of cross-validation. Generally speaking, the errors generated from cross-validation are affected both by sampling randomness (from the population) and by sub-sampling randomness that arises from partitioning the sample into folds. Thus, the errors from cross-validation are potentially more volatile than the usual errors from estimation. Eqs.~(\ref{eq1:thm2.2}) and (\ref{eq2:thm2.2}) offers a way to characterize the property of eGE despite of the effect of sub-sampling randomness.

\begin{figure}

    \centering
	\subfloat[\label{fig:Ktradeoff1}eGE in each round of cross-validation]
	{\includegraphics[width=0.35\paperwidth]{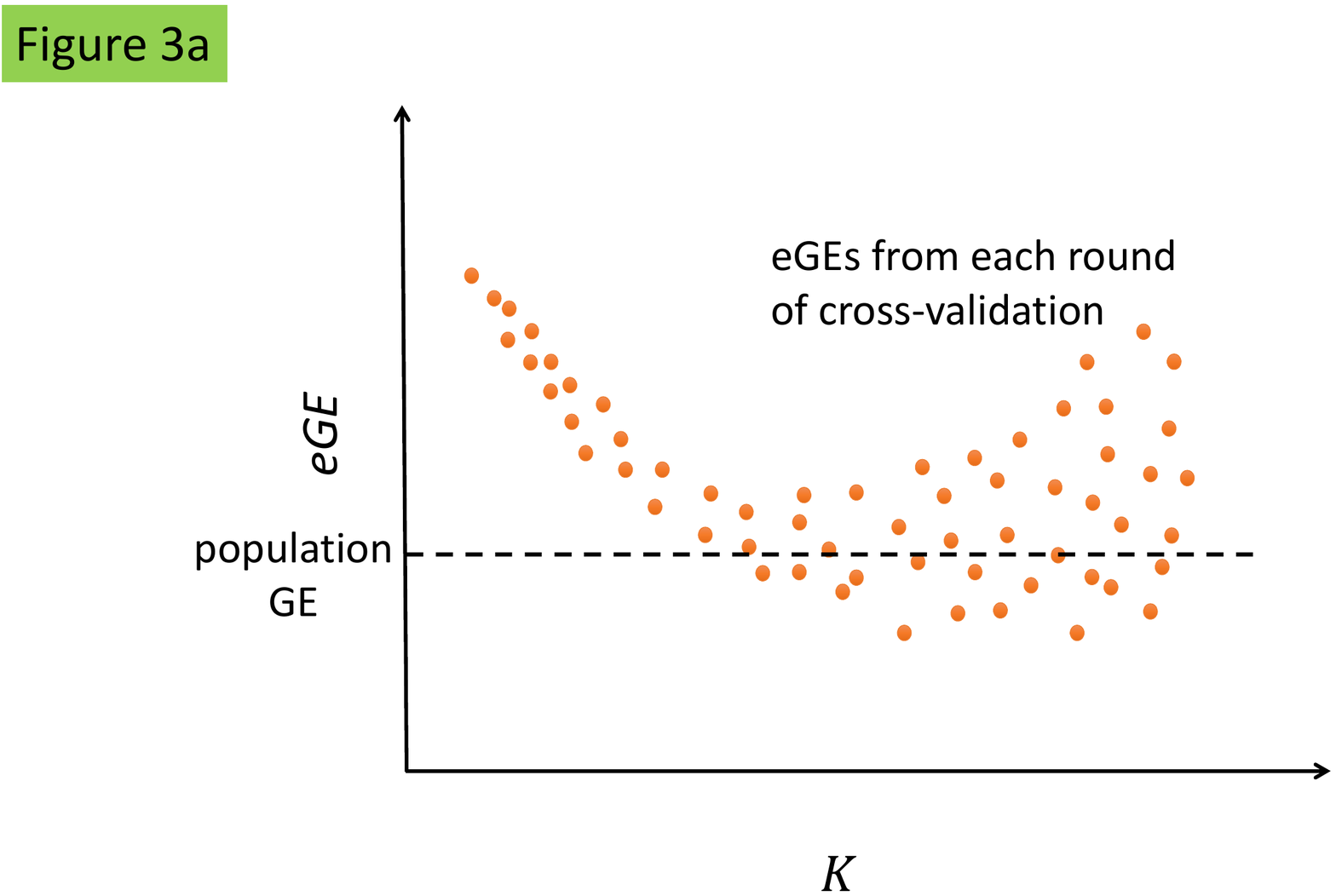}}
	\centering	
	\subfloat[\label{fig:Ktradeoff2}average eGEs from $K$ rounds of cross-validation]
	{\includegraphics[width=0.35\paperwidth]{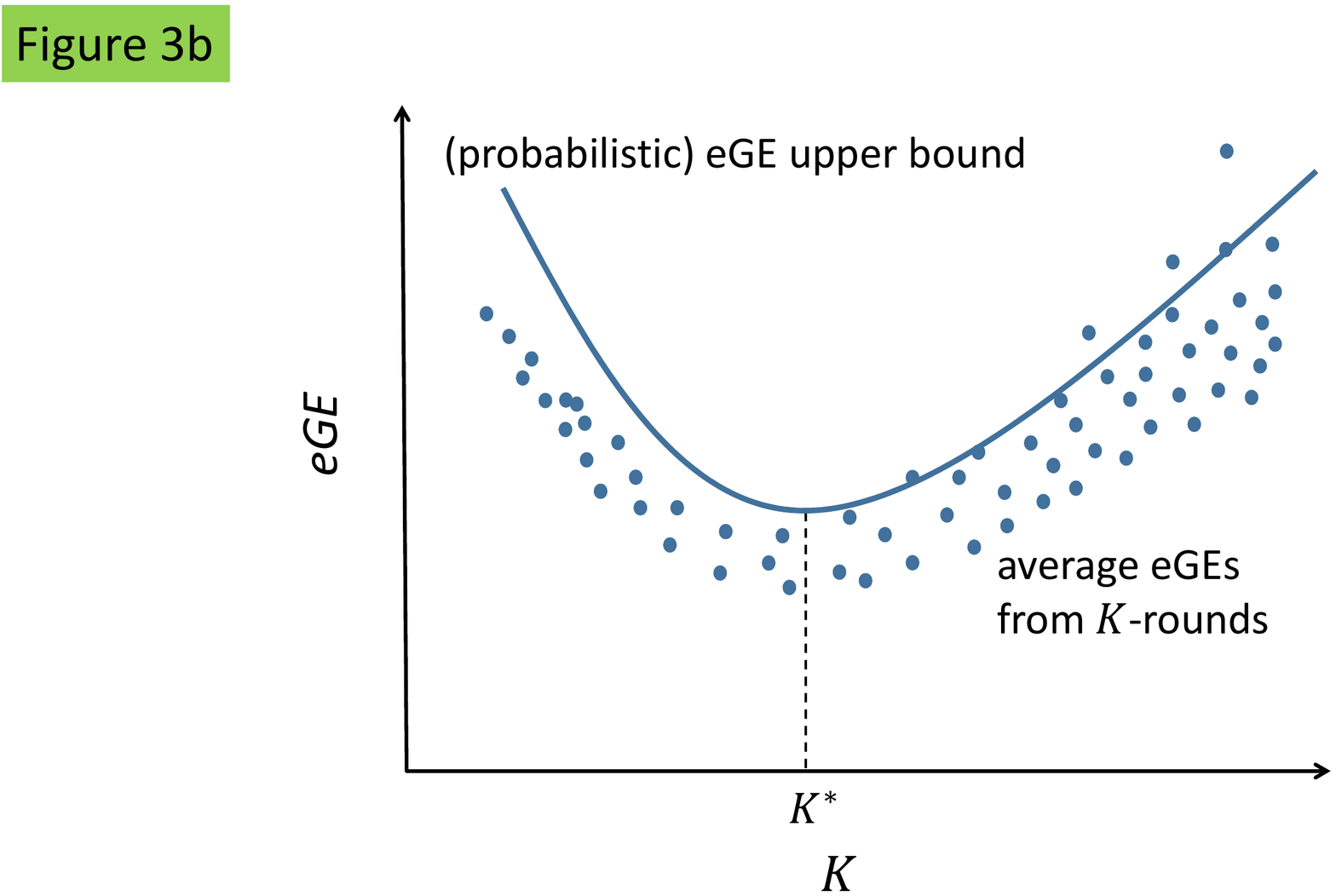}}
	
    \caption{\label{fig:Ktradeoff}The bias-variance trade-off for cross-validation eGE}

\end{figure}

The implications of eqs.~(\ref{eq1:thm2.2}) and (\ref{eq2:thm2.2}) are as follows.
\begin{enumerate}

    \item  \textit{Upper bound of the eGE.} Similar to eq.~(\ref{eq:thm2.1}), eq.~(\ref{eq1:thm2.2}) and (\ref{eq2:thm2.2}) serve as the upper bound of the cross-validated eGE. Both equations reveal the trade-off between eTE and eGE and the influence of the tails of the loss distribution. In convolution, the tails dramatically affect the upper bound of the eGE. If $T_q$ is light-tailed, $\alpha$ converges to $1$ exponentially when $n \rightarrow \infty$. When $T_q$ is sub-exponential, the convolution is more complicated and it is hard to approximate the probability. However, with sufficiently large $n$, Theorem~\ref{thm2.2} shows that we can approximate the convoluted probability.

    \item  \textit{The trade-off between accuracy and looseness of the upper bound.} Similar to eq.~(\ref{eq:thm2.1}), in both eq.~(\ref{eq1:thm2.2}) and (\ref{eq2:thm2.2}) we need to consider the trade-off between between $\alpha$ and $\varsigma$ (or $\varpi$ as in Theorem~\ref{thm2.1}), i.e., the trade-off between efficiency and accuracy. Similar to Theorem~\ref{thm2.1}, ceteris paribus, a larger $\varsigma$ in each round of cross-validation increases $\varpi$ and $\alpha$. Thus, in each round, the upper bound moves upwards and the probability the upper bound holds increases, implying that both increase overall under cross-validation. While the probability the bound holds increases, it comes at the cost of a looser bound.

    \item  \textit{Cross-validation hyperparameter $K$.} Eqs.~(\ref{eq1:thm2.2}) and (\ref{eq2:thm2.2}) characterize how $K$ affects the average eGE from cross-validation (also called the cross-validation error in the literature). With a given sample and fixed $K$, sub-sampling randomness will produce a different average eGE each time cross-validation is performed. From Definition~\ref{defn2.1}, $n_t = n(K-1)/K$ and $n_s = n/K$, so the sizes of the training and test sets change with $K$. As $K$ increases the \emph{test sets} become smaller, increasing the influence of sub-sampling randomness on the eGE. On the other hand, as $K$ decreases the \emph{training sets} become smaller, increasing the influence of sub-sampling randomness on eTE. The two effects complicate analysis of the effect of $K$ on trade-off between eTE and eGE under cross-validation. Thus, to characterize the influence of sub-sampling randomness, we establish the trade-off for cross-validation by a bound, after running cross-validation many times.\footnote{By contrast, for extremum estimators like OLS, the bias-variance trade-off is much more straightforward to analyze for different $p$ because the sample is fixed.} Figure~\ref{fig:Ktradeoff} illustrates the effect of $K$.
	\begin{itemize}

        \item   \textit{Small $K$.} For low values of $K$, $n_t$ is low in each round of in-sample estimation and the eTE in each round $q$, $\mathcal{R}_{n_t}(b_{train}|Y_{t}^q,X_{t}^q)/(1-\sqrt{\epsilon})$, is more biased away from the population error, as shown in Figure~\ref{fig:Ktradeoff1}. Also for small $K$, the $K$-round average eTE (the first term on the RHS of eqs.~(\ref{eq1:thm2.2}) and (\ref{eq2:thm2.2})), is more biased away from the true population error, as shown in Figure~\ref{fig:Ktradeoff2}. As a result, the RHS of eqs.~(\ref{eq1:thm2.2}) and (\ref{eq2:thm2.2}) suffer more from finite-sample bias for low values of $K$. However, since a small $K$ implies $n_s$ is relatively large, more data is used for eGE calculation in each round, in each round the eGE on the test set should be less volatile. Thus, the $K$-round averaged eGE for cross-validation is relatively less volatile, reflecting the fact that $\varsigma$ is not very large in eqs.~(\ref{eq1:thm2.2}) and (\ref{eq2:thm2.2}).

        \item   \textit{Large $K$.} For high values of $K$, $n_s$ is low. Given a small test set size, the eGE in each round may be hard to bound from above, the averaged eGE from $K$ rounds will be more volatile and $\varsigma$ will increase. However, with a high $K$, the first term on the RHS of eqs.~(\ref{eq1:thm2.2}) and (\ref{eq2:thm2.2}) tends to be closer to the true population error and the averaged eGE suffers less from bias.
	\end{itemize}	
	
    In summary, Figure~\ref{fig:Ktradeoff2} shows that as the value of $K$ increases, the averaged eGE from cross-validation follows a typical bias-variance trade-off. For low values of $K$, the average eGE is less volatile but more biased away from the population error. As $K$ increases, the averaged eGE becomes more volatile but less biased away from the population error.

\end{enumerate}

\bigskip
Theorem~\ref{thm2.2} confirms the exhaustive simulation study results from \citet{kohavi1995study}. As a side benefit, Theorem~\ref{thm2.2} also suggests that an optimal number of folds $K$ may exist for each sample when the cross-validation approach is used with extremum estimation, as in the $K^*$ shown in Figure~\ref{fig:Ktradeoff2}. More specifically, the $K$ that minimizes the upper bound (\ref{eq1:thm2.2}) and (\ref{eq2:thm2.2}) also maximizes the GA from cross-validation. We leave to the next section discussion on the optimal $K$ for regression.

Theorems~\ref{thm2.1} and~\ref{thm2.2} establish upper bounds for the eGE of the extremum estimator given any size random sample, which reveals a method to analyze the prorperty of eGE for the extremum estimators. Potentially, if we take eGE as the criterion for model selection, Theorems~\ref{thm2.1} and~\ref{thm2.2} may be used to evaluate the performance of model selection under validation and cross-validation. Hence it would be natural to select the model with minimal eGE in the space of alternative models

\subsection{Generalization error minimization}

By establishing upper bounds for eGE under validation or cross-validation in section 2.1, we show that, as a criterion for model evaluation, a number of properties for eGE could be shown in finite samples and the asymptotic case, which suggest that it may serve as a good angle to understand model selection. Hence, by considering eGE as a criterion for model selection, we propose selecting the model based on minimizing the eGE, which we refer to as \textbf{generalization error minimization} or \textbf{GEM}.

Generally speaking,  GEM can be implemented alongside with many conventional techniques of model selection, such as penalized regressions, the information criteria and maximum a posteriori (MAP). However, in next section, we show that GEM works especially well for penalized regression. As shown in Algorithm~1, penalized regression estimation returns a $b_{\lambda}$ for each $\lambda$. Each value of $\lambda$ generates a different model and a different eGE. As a result, Theorems~\ref{thm2.1} and~\ref{thm2.2} guarantee that the model with the minimum eGE among $\left\{ b_{\lambda}\right\}$ has the best empirical generalization ability. By applying GEM in conjunction with validation/cross-validation and various penalty methods, the theoretical properties of the penalized regressions, such as its robustness, mode of consistency and convergence rate could be analyzed by directly applying the upper bounds we derive previously.

%%%%%%%%%%%%%%%%%%%%%%%%%%%%%%%%%%%%%%%%%%%%%%%%%%%%%%%%%%%%%%%%%%%%%%%%%%%%%%%%%%%%%%%%%%%%%%%%%%%%%%%%%%%
%%%%%%%%%%%%%
%%%%%%%%%%%%%            SECTION 3
%%%%%%%%%%%%%
%%%%%%%%%%%%%%%%%%%%%%%%%%%%%%%%%%%%%%%%%%%%%%%%%%%%%%%%%%%%%%%%%%%%%%%%%%%%%%%%%%%%%%%%%%%%%%%%%%%%%%%%%%%

\section{Finite-sample and asymptotic properties for penalized regression under GEM}

In Section~2,  to analyze eGE from the perceptive of model-free, we establish a class of upper bounds for eGE of the extremum estimators, which is potentially connected to the eTE-eGE trade-off, bias-variance trade-off and cross validation; we also propose the GEM as the general framework of model selection. In this section, we implement the GEM onto regression analysis. As shown in the following section, we apply eGE to regression analysis and show that GEM serves as a new and clear angle to understand the procedure of penalized regression. By applying the eGE as the criterion of model selection, model selection directly serves as an effective method to improve GA; the other way around, the properties of model selection can directly be explained and re-framed by the eGE of the model. Moreover, other than the properties in section 2, additional properties can be established for penalized regression under GEM framework.  Specifically, we establish: (1) specific error bounds for any penalized regression, (2) $\mathcal{L}_2$-consistency for all penalized regression estimators, (3) that the upper bound for the $\mathcal{L}_2$-difference between the penalized regression estimator and the OLS estimator is a function of the eGE, the tail behavior of distribution for the loss function and the exogeneity of the sample.

\subsection{Penalized regression}

Firstly, we formally define penalized regression and its two most popular variants:  lasso ($\mathcal{L}_1$-penalized regression) and ridge ($\mathcal{L}_2$-penalized regression). It is important to stress that each variable in $(Y,X)$ must be standardized before implementing penalized regression. As shown by \citet{tibshirani96}, without standardization the penalized regression estimates may be influenced by the magnitude (units) of the variables. After standardization, of course, $X$ and $Y$ are unit- and scale-free.

\begin{defn}
[Penalized regression, ridge regression, lasso regression, $\mathcal{L}_2$-eGE and $\mathcal{L}_2$-eTE]
\end{defn}

		\begin{enumerate}
            \item   The general form of the objective function for penalized regression is
                \begin{equation}
			        \min_{b_\lambda}
                    \frac{1}{n}\left(\left\Vert Y-Xb_{\lambda}\right\Vert_{2}\right)^{2}
			      	+\lambda\;\mathrm{Penalty}(\Vert b_{\lambda}\Vert_{\gamma}^\gamma).
			    \label{lasso-type}
			    \end{equation}
                where the penalty term $\mathrm{Penalty}(\Vert\cdot\Vert_{\gamma})$ is a function of the $\mathcal{L}_{\gamma}$-norm of $b_{\lambda}$.

            \item   Let $b_{\lambda}$ denote the solution to the constrained minimization eq.~(\ref{lasso-type}) for a given value of the penalty parameter $\lambda$. Let $b^*$ denote the estimator with the minimum eGE among all the alternative $\{b_{\lambda}\}$ (as in Algorithm~1 in Section~1). Let  $b_{OLS}$ denote the OLS estimator on the training set.

            \item   The objective function for the lasso ($\mathcal{L}_1$-norm penalty) is
			    \begin{equation}
			      	\min_{b_\lambda}
			      	\frac{1}{n}
			      	\left(\left\Vert Y-Xb_{\lambda}\right\Vert _{2}\right)^{2}
			      	+\lambda\Vert b_{\lambda}\Vert_{1},
			    \label{lasso-reg}
			    \end{equation}
                and the objective function for ridge regression ($\mathcal{L}_2$-norm penalty) is
			    \begin{equation}
			      	\min_{b_\lambda}
			      	\frac{1}{n}
			      	\left(\left\Vert Y-Xb_{\lambda}\right\Vert _{2}\right)^{2}
			      	+\lambda\Vert b_{\lambda}\Vert_{2}^2.
			      	\label{ridge-reg}
			    \end{equation}

                \item   The $\mathcal{L}_2$-norm eTE and eGE for any $b$ are defined, respectively,
			        \begin{align*}
			      	    \mathcal{R}_{n_t}(b|Y_{t},X_{t})
                        & =\frac{1}{n_t}\,(\Vert Y_{t}-X_{t}b\Vert_{2})^{2} \\
			      	    \mathcal{R}_{n_s}(b|Y_{s},X_{s})
                        & =\frac{1}{n_s}\,(\Vert Y_{s}-X_{s}b\Vert_{2})^{2}
			        \end{align*}

		\end{enumerate}

\begin{figure}[ht]
    \centering
    \subfloat[\label{figpenreg:fig3}$\mathcal{L}_{0.5}$, $\mathcal{L}_1$ and $\mathcal{L}_2$ constraints]
	{\includegraphics[width=0.235\paperwidth]{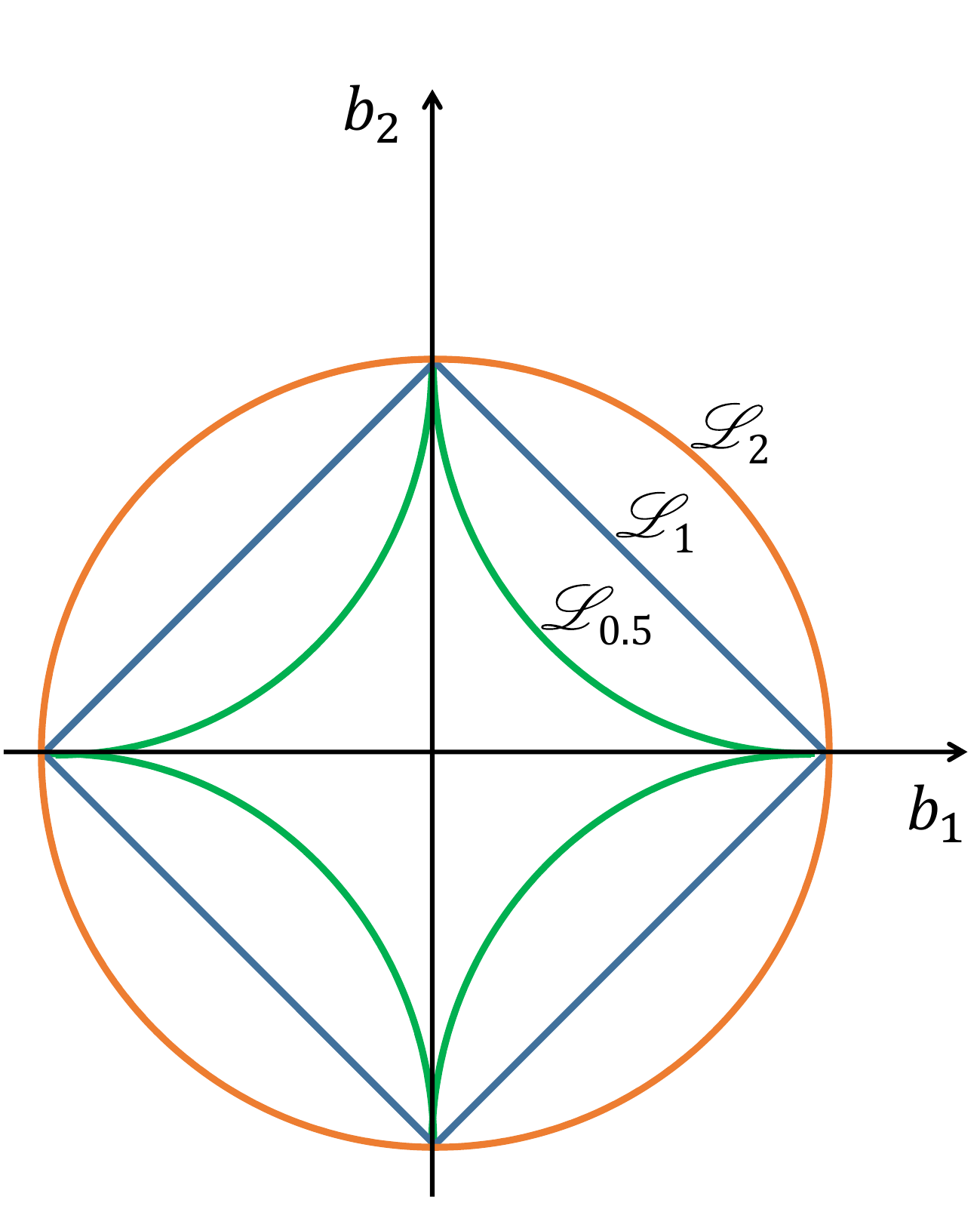}}
	\subfloat[\label{figpenreg:fig1}$\mathcal{L}_1$ penalty (lasso)]
	{\includegraphics[width=0.235\paperwidth]{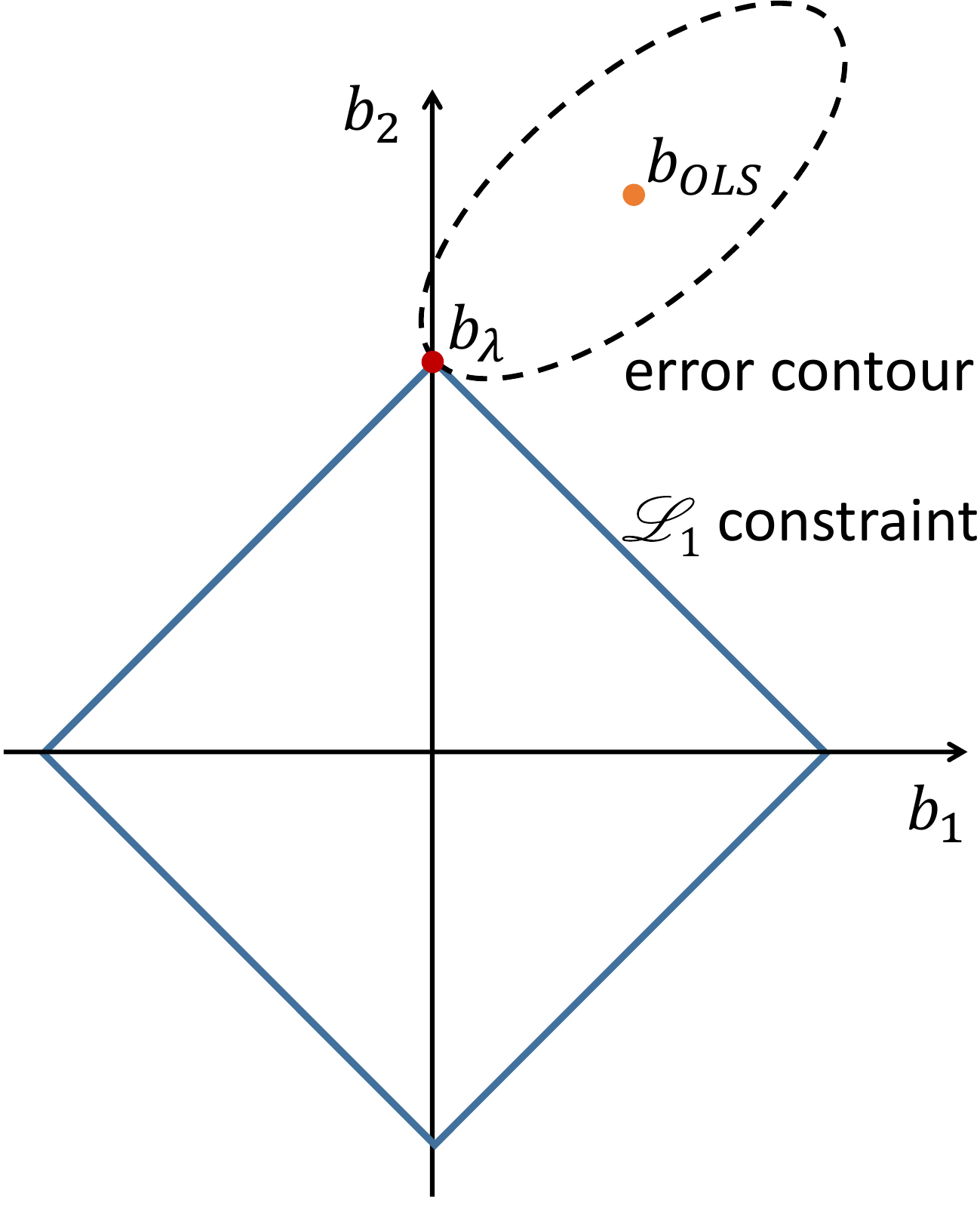}}
	\subfloat[\label{figpenreg:fig2}$\mathcal{L}_2$ penalty (ridge)]
	{\includegraphics[width=0.235\paperwidth]{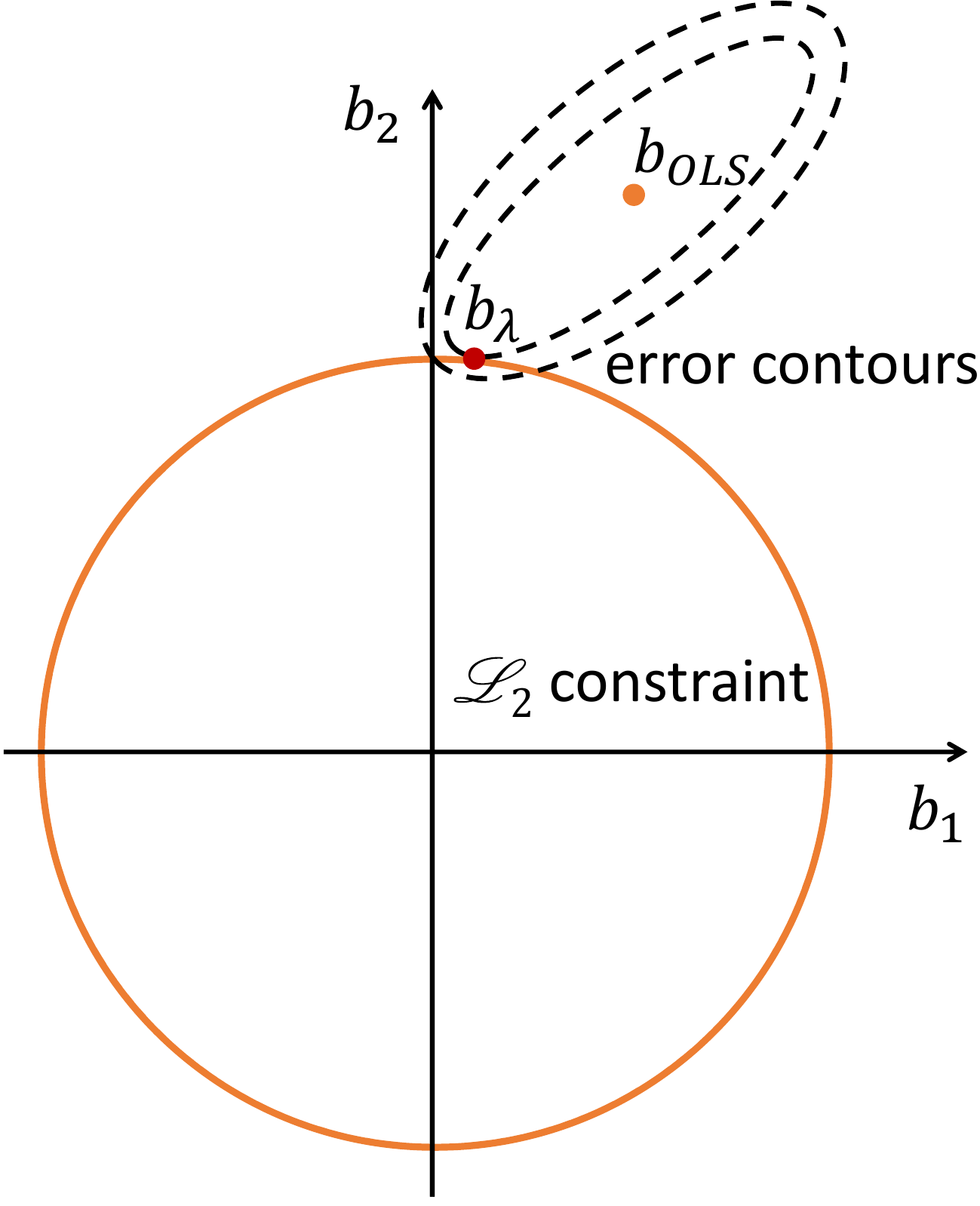}}

	\caption{\label{figpenreg}Comparison of estimates from OLS and various penalized regressions}
\end{figure}

The idea behind penalized regression is illustrated in Figure~\ref{figpenreg}. As shown in Figure~\ref{figpenreg:fig3}, different $\mathcal{L}_{\gamma}$ penalties correspond to different boundaries for the estimation feasible set. For the $\mathcal{L}_1$ penalized regression (lasso), the feasible set is a diamond. The feasible set expands to a circle under an $\mathcal{L}_{2}$ penalty. As illustrated in Figures~\ref{figpenreg:fig1} and~\ref{figpenreg:fig2}, for given $\lambda$, the smaller $\gamma$, the more likely $b_{\lambda}$ is a corner solution. It follows that under the $\mathcal{L}_{1}$ penalty, variables are more likely to be dropped compared with the $\mathcal{L}_{2}$ penalty.\footnote{For $0<\gamma<1$, the penalized regression may be a non-convex programming problem. While general algorithms have not been found for non-convex optimization, \citet{strongin2013global}, \citet{yan2001global} and \citet{noor2008differentiable} have developed working algorithms. For $\gamma=0$, the penalized regression becomes a discrete programming problem, which can be solved by Dantzig-type methods; see \citet{candestao07}.} In the special case when $\gamma=0$ and $\lambda=2$ ($\lambda=\ln n_t$), the $\mathcal{L}_0$ penalized regression is identical to the Akaike (Bayesian) information criterion.

Penalized regression primarily focuses on overfitting. By contrast, OLS minimizes the eTE without any penalty, often resulting in a large eGE (e.g., the `overfitting' in Figure~\ref{fig:VCineq}). It is also possible that OLS fits the training data poorly, causing both the eTE and eGE to be large (e.g., the `underfitting' in Figure~\ref{fig:VCineq}). Generally speaking, it is easier to deal with overfitting than underfitting.\footnote{See eq.~(\ref{eq1:thm3.1}) and (\ref{eq2:thm3.1}).} Overfitting in OLS is usually caused by including too many variables, which can be resolved by reducing $p$. Underfitting, however, is likely due to a lack of data (variables) and the only remedy is to collect more data.

\subsection{GEM for penalized regression}

\begin{figure}
	\centering
	\includegraphics[width=0.5\paperwidth]{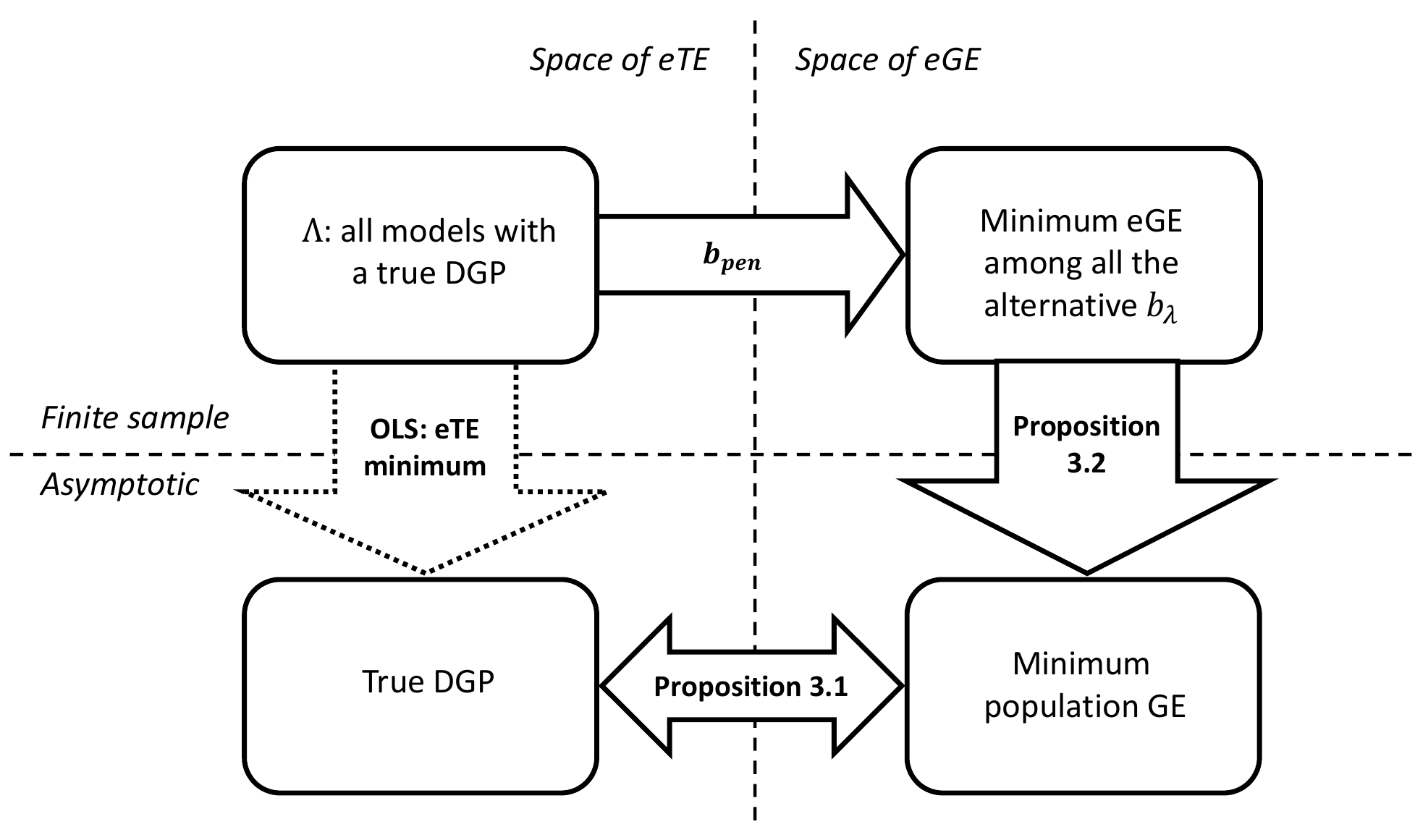}

	\caption{\label{fig:outline}Outline of proof strategy}
\end{figure}

The conventional route to establish finite-sample and asymptotic properties for regression is by analyzing the properties of the estimator in the space of the eTE. By contrast, to study how penalized regression improves GA, we reformulate the analysis in the space of the eGE. Figure~\ref{fig:outline} outlines our proof strategy. We show that a number of finite-sample properties for penalized regression can be established under the GEM framework.

In asymptotic analysis, consistency is typically considered to be one of the most fundamental properties. To demonstrate that GEM is a viable estimation framework, we prove that the penalized regression model selected by eGE minimization converges to the true DGP as $n\rightarrow\infty$. Essentially, we show that penalized regression bijectively maps $b^*$ to the minimal eGE among $\{b_{\lambda}\}$ on the test set. To link the finite-sample and asymptotic results we need to show that, if the true DGP $\beta$ is bijectively assigned to the global minimum eGE in the population, and if
\[ \min_{b\in b_{\lambda}}\;
	       \frac{1}{n_s}\sum_{i=1}^{n_s}\Vert Y_{s}-X_{s}b\Vert_{2}^{2}
	       \quad\longrightarrow\quad
	       \min_b \int\Vert y-\mathbf{x}^{T}b\Vert_{2}^{2}~\mathrm{d}F(y, \mathbf{x}),
\]
then $b^*$ is consistent in probability or $\mathcal{L}_2$, or
\[
       b^* =
       \argmin_{b_{\lambda}}\{\mbox{eGEs}\}
       \quad\xrightarrow{\mathbf{P}~\mathrm{or}~\mathcal{L}_2}\quad
       \argmin_b\int\Vert y_{s}-\mathbf{x}_{s}^{T}b\Vert_{2}^{2}
       \mathrm{d}F(y,\mathbf{x}) = \beta.
\]

To establish consistency for penalized regression, we make the following three additional assumptions.

\bigskip
\noindent
\textbf{Further assumptions}

\begin{enumerate}
\setlength\itemsep{1em}

    \item[\textbf{A4.}]    The true DGP is $Y = X\beta + u$.

	\item[\textbf{A5.}]    $\mathbb{E} \left( u^T X \right) = \mathbf{0}$.

	\item[\textbf{A6.}]    No perfect collinearity in $X$.

\end{enumerate}
Assumptions A4--A6 restrict the true DGP indexed by $\beta$ to be identifiable. Otherwise, there may exist an alternative that is not statistically different from the true DGP. The assumptions are standard for linear regression.

Under assumptions A1--A6, we first show that the true DGP has the lowest generalization error.
%
%%%%%%%%%%%%%%%%%%%%%%%%%%%%%%%%%%%%%%%%%%%%%%%%%%%%%%%%%%%%%%%%%%%%%%%%%%%%%%%%%%%%%%%%%%%%
% Proposition 3.1
%%%%%%%%%%%%%%%%%%%%%%%%%%%%%%%%%%%%%%%%%%%%%%%%%%%%%%%%%%%%%%%%%%%%%%%%%%%%%%%%%%%%%%%%%%%%
%
\begin{prop}[Identification of $\beta$ in the space of eGE]
Under A1--A6, the true DGP, $Y = X\beta + u$, is the one and only one offering the minimal eGE as $\widetilde{n} \rightarrow \infty$.

\label{prop3.1}
\end{prop}
Proposition~\ref{prop3.1} states that there is a bijective mapping between $\beta$ and the global minimum eGE in the population. If A5 or A6 were violated, variables may exist in the sample that render the true DGP not to have the minimum eGE in the population.

As shown in Algorithm~1, penalized regression chooses $b^*$ to be the model with the minimum eGE in $\{b_\lambda\}$. Thus, we need to establish that when the sample size is large enough, the true DGP is included among $\{b_{\lambda}\}$, the list of models selected by validation or cross-validation.
%
%%%%%%%%%%%%%%%%%%%%%%%%%%%%%%%%%%%%%%%%%%%%%%%%%%%%%%%%%%%%%%%%%%%%%%%%%%%%%%%%%%%%%%%%%%%%
% Proposition 3.2
%%%%%%%%%%%%%%%%%%%%%%%%%%%%%%%%%%%%%%%%%%%%%%%%%%%%%%%%%%%%%%%%%%%%%%%%%%%%%%%%%%%%%%%%%%%%
%
\begin{prop}[Existence of $\mathcal{L}_2$-consistency]
Under A1--A6 and Proposition~\ref{prop3.1}, there exists at least one $\widetilde{\lambda}$ such that $\lim_{\widetilde{n} \rightarrow \infty} \Vert  b_{\widetilde{\lambda}} - \beta \Vert_2 = 0$.

\label{prop3.2}
\end{prop}

\begin{figure}[ht]
	\centering
	\subfloat[\label{figlassoshrink:fig1}low $\lambda$ (under-shrinkage)]
	{\includegraphics[width=0.235\paperwidth]{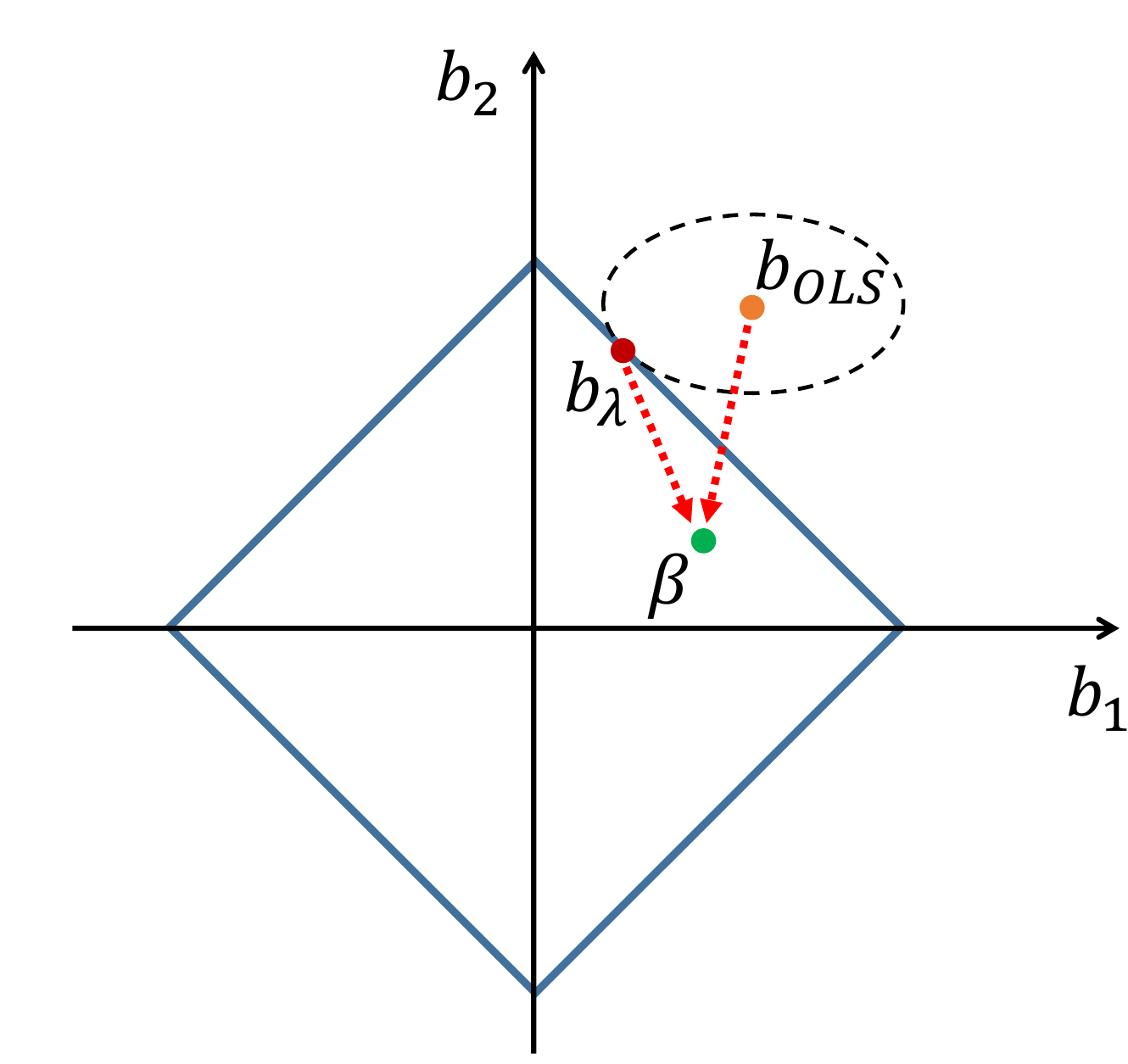}}
	\subfloat[\label{figlassoshrink:fig2}oracle $\lambda$ (perfect shrinkage)]
	{\includegraphics[width=0.235\paperwidth]{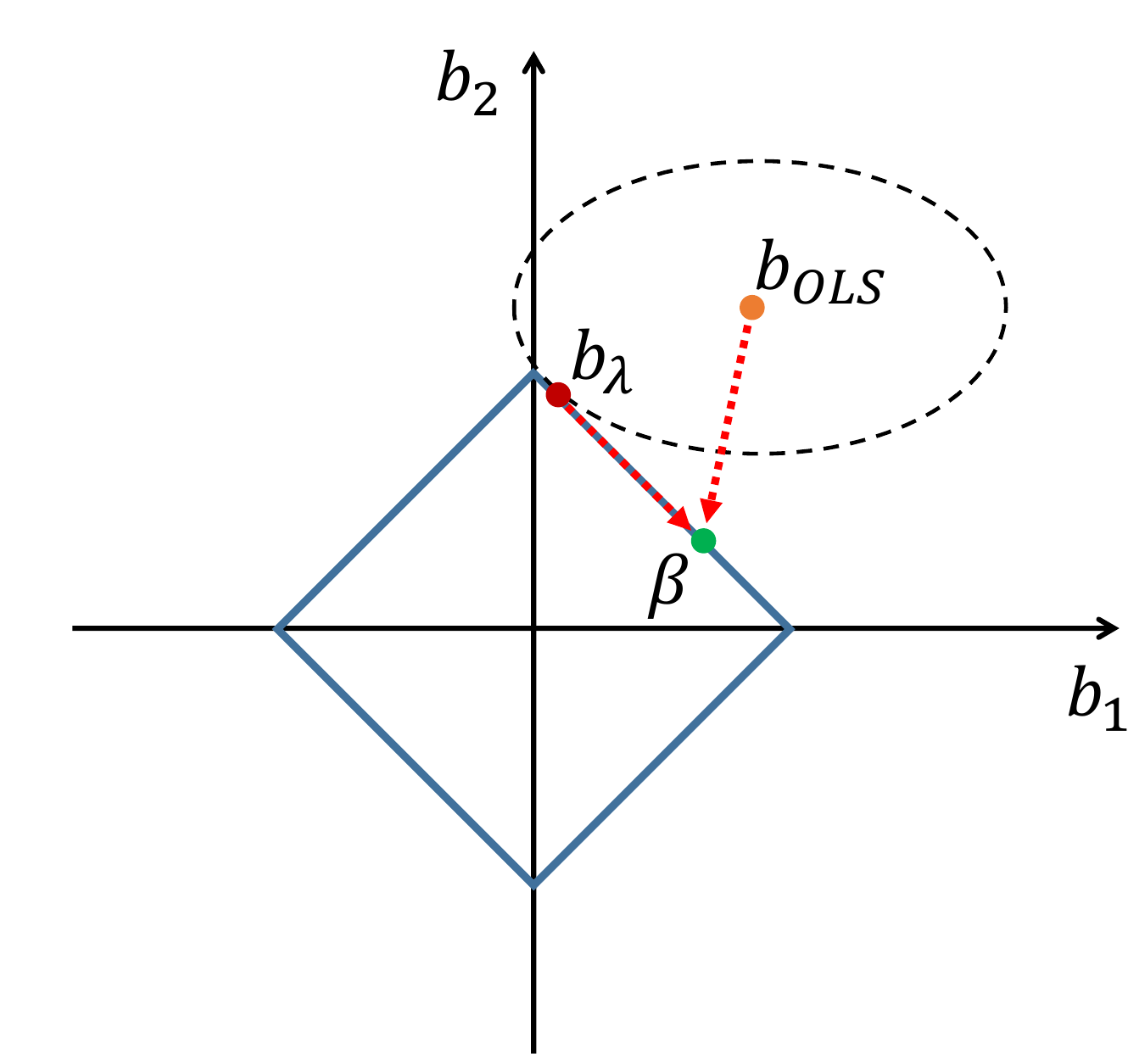}}
	\subfloat[\label{figlassoshrink:fig3}high $\lambda$ (over-shrinkage)]
	{\includegraphics[width=0.235\paperwidth]{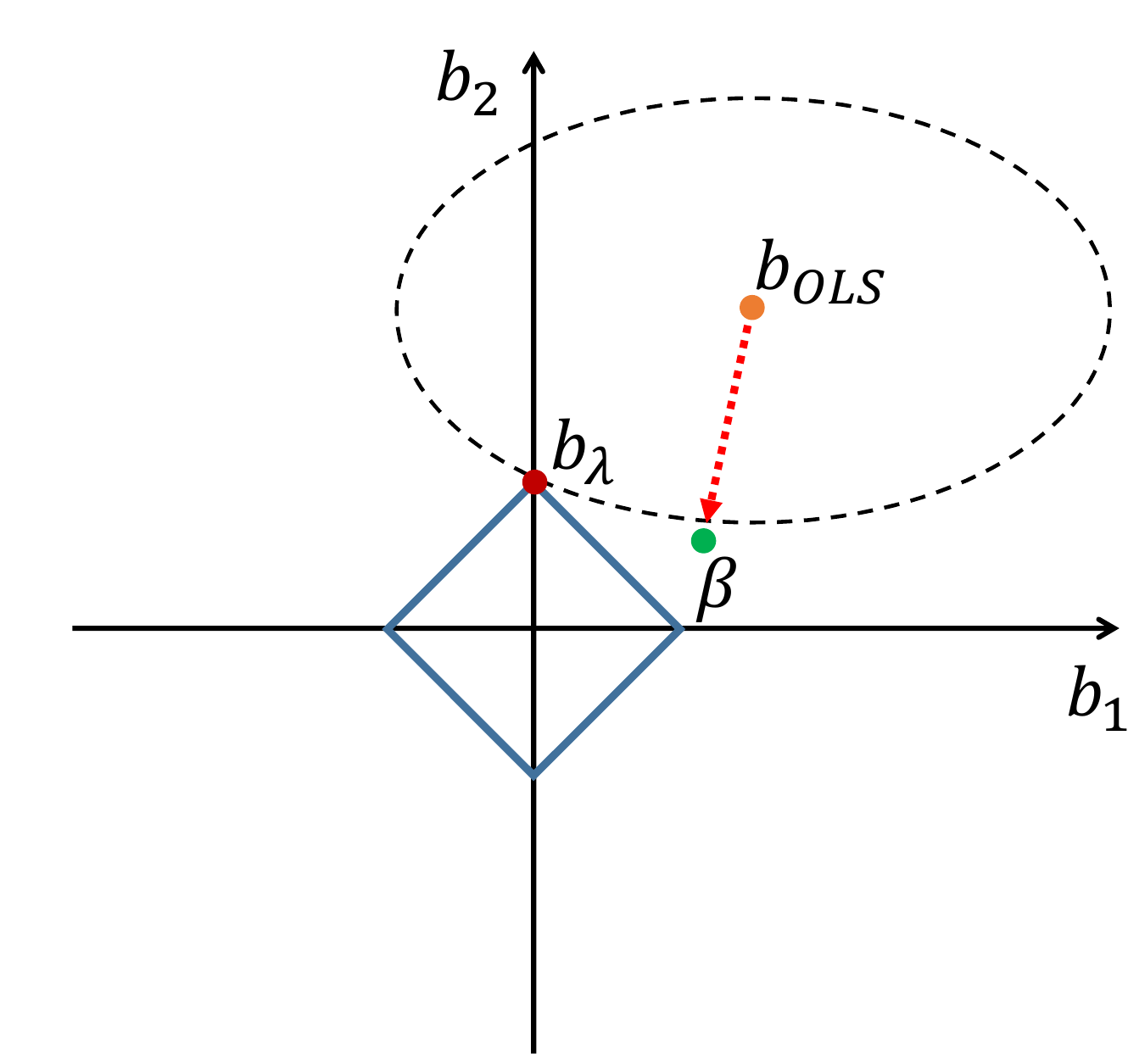}}

    \caption{\label{figlassotypeshrink:fig}Shrinking for various values of $\lambda$ under the $\mathcal{L}_1$ penalty}

\end{figure}

Using lasso as the example of penalized regression, Figure~\ref{figlassotypeshrink:fig} illustrates Propositions~\ref{prop3.1} and~\ref{prop3.2}. In Figure~\ref{figlassotypeshrink:fig}, $\beta$ refers to the true DGP, $b_{\lambda}$ refers to the solution of eq.~(\ref{lasso-reg}), and the diamond-shaped feasible sets are due to the $\mathcal{L}_1$ penalty. Different values of $\lambda$ imply different areas for the feasible sets, which get smaller as the value of $\lambda$ increases. There are three possible cases: (i) under-shrinkage---for a low value of $\lambda$ (Figure~\ref{figlassoshrink:fig1}), $\beta$ lies within the feasible set and has the minimum eTE in the population; (ii) perfect-shrinkage---for the oracle $\lambda$ (Figure~\ref{figlassoshrink:fig2}), $\beta$ is located precisely on the boundary of the feasible set and still has the minimum eTE in the population; (iii) over-shrinkage---for a high value of $\lambda$ (Figure~\ref{figlassoshrink:fig3}), $\beta$ lies outside the feasible set. In cases (i) and (ii), the constraints become inactive as $\widetilde{n}\rightarrow\infty$, so $\lim_{\widetilde{n}\rightarrow\infty}b_{\lambda} = \lim_{\widetilde{n}\rightarrow\infty}b_{OLS} = \beta$. However, in case (iii), $\lim_{\widetilde{n}\rightarrow\infty}b_{\lambda}\neq\beta$. An important implication is that tuning the penalty parameter $\lambda$ is critical for the theoretical properties of the penalized regression estimators.

\subsection{Main results for penalized regression under GEM}

Intuitively, the penalized regression estimator will be consistent in some norm or measure as long as, for a specific $\lambda$, $\beta$ lies in the feasible set and offers the minimum eTE. In practice, however, we may not know a priori whether $\lambda$ causes over-shrinking or not, especially when the number of variables, $p$, is not fixed. As a result, we need to use an approach like validation or cross-validation to tune the value of $\lambda$. Applying the results in Section~2, we show that GEM guarantees that the model selected by penalized regression with the appropriate $\lambda$, $b^*$, asymptotically converges in $\mathcal{L}_{2}$ to the true DGP.

In this section we analyze the finite-sample and asymptotic properties of the GEM estimator in two settings: $n\geqslant p$ and $n < p$. In the case where $n\geqslant p$, OLS is feasible, and is the unpenalized regression estimator. In the case where $n < p$, OLS is not feasible, and forward stagewise regression (FSR) is the unpenalized regression estimator.

\subsubsection{GEM for penalized regression with $n\geqslant p$}

Firstly, by adapting eq.~(\ref{eq:thm2.1}) and (\ref{eq1:thm2.2}) and (\ref{eq2:thm2.2}) for regression, we establish the upper bound of the eGE.
%
%%%%%%%%%%%%%%%%%%%%%%%%%%%%%%%%%%%%%%%%%%%%%%%%%%%%%%%%%%%%%%%%%%%%%%%%%%%%%%%%%%%%%%%%%%%%
% Lemma 3.1
%%%%%%%%%%%%%%%%%%%%%%%%%%%%%%%%%%%%%%%%%%%%%%%%%%%%%%%%%%%%%%%%%%%%%%%%%%%%%%%%%%%%%%%%%%%%
%
\begin{lem}[Upper bound for the eGE of the OLS estimator]
Under A1--A6, if we assume $u \sim N(0,\sigma^2)$,

    \begin{enumerate}

    \item \textbf{Validation case.} The following bound for the eGE for $b_{OLS}$ holds with probability at least $\varpi(1-1/n_t)$, $\forall\varpi\in\left(0,1\right)$.
		\begin{equation}
			\frac{1}{n_s}(\Vert e_s \Vert_2)^2
			\leqslant \frac{(\Vert e_t \Vert_2)^2}{n_t(1-\sqrt{\epsilon})}
			+ \frac{2 \sigma^4}{n_s\sqrt{1-\varpi}},
			\label{eq1:lem3.1}
		\end{equation}
        where $(\Vert e_s \Vert_2)^2$ is the eGE and $(\Vert e_t \Vert_2)^2$ is the eTE of the OLS estimator, and $\epsilon$ is defined in Lemma~\ref{lem2.1}.

    \item \textbf{$K$-fold cross-validation case.} The following bound for the eGE for $b_{OLS}$ holds with probability at least $\varpi(1-1/n_t)$, $\forall\varpi\in\left(0,1\right)$.
		\begin{equation}
			\frac{1}{K}\sum^{K}_{q=1}\frac{(\Vert e_s^q \Vert_2)^2}{n/K}
			\leqslant
			\frac{\sum_{q=1}^K (\Vert e_t^q \Vert_2)^2}
            {n(K-1)(1-\sqrt{\epsilon})} + \frac{2\sigma^4}{(n/K)\sqrt{1-\varpi}},
			\label{eq2:lem3.1}
		\end{equation}
        where $(\Vert e_s^q \Vert_2)^2$ is the eGE and $(\Vert e_t^q \Vert_2)^2$ is the eTE of the OLS estimator in the $q$th round of cross-validation, and $\epsilon$ is defined in Lemma~\ref{lem2.1}.

   \end{enumerate}    	

\label{lem3.1}	
\end{lem}

In a similar fashion to eqs.~(\ref{eq:thm2.1}) and~(\ref{eq1:thm2.2}) and (\ref{eq2:thm2.2}), eqs.~(\ref{eq1:lem3.1}) and (\ref{eq2:lem3.1}) measure the upper bound of the eGE for the OLS estimator under validation and cross-validation, respectively. In standard practice, of course, neither validation nor cross-validation are implemented as part of OLS estimation and the eGE of the OLS estimator is not computed. Nevertheless, eqs.~(\ref{eq1:lem3.1}) and (\ref{eq2:lem3.1}) show that it is possible to compute the eGE of the OLS estimator without having to carry out validation or cross-validation. Eqs.~(\ref{eq1:lem3.1}) and (\ref{eq2:lem3.1}) also show that the higher the variance of $u$ in the true DGP, the higher the upper bound of the eGE under validation and cross-validation.

As a bonus of the GEM approach, eq.~(\ref{eq2:lem3.1}) also shows that we can find the $K$ that maximizes the GA from cross-validation by tuning $K$ to the lowest upper bound of the cross-validated eGE, determined by minimizing the expectation of the RHS of eq.~(\ref{eq2:lem3.1}).
%
%%%%%%%%%%%%%%%%%%%%%%%%%%%%%%%%%%%%%%%%%%%%%%%%%%%%%%%%%%%%%%%%%%%%%%%%%%%%%%%%%%%%%%%%%%%%
% Corollary 3.1
%%%%%%%%%%%%%%%%%%%%%%%%%%%%%%%%%%%%%%%%%%%%%%%%%%%%%%%%%%%%%%%%%%%%%%%%%%%%%%%%%%%%%%%%%%%%
%
\begin{cor}[The optimal $K$ for penalized regression]
Under A1--A6, and based on eq.~(\ref{eq2:lem3.1}) from Lemma~\ref{lem3.1}, if we also assume the error term $u \sim N(0, \sigma^2)$, the optimal $K$ for cross-validation in penalized regression (the minimum expected upper bound of eGE) is defined:
	\[
        K^* = \argmin_{K} \frac{\sigma^2}{1 - \sqrt{\epsilon}} + \frac{2\sigma^4}{(n/K)\sqrt{1 - \varpi}}
	\]

\label{cor3.1}
\end{cor}

The penalty parameter $\lambda$ can be tuned by validation or $K$-fold cross-validation. For $K\geqslant 2$, we have $K$ different test sets for tuning $\lambda$ and $K$ different training sets for estimation. Using eqs.~(\ref{eq1:lem3.1}) and (\ref{eq2:lem3.1}), we now establish, in two steps, an upper bound for the $\mathcal{L}_2$-norm difference between the unpenalized estimator $b_{OLS}$ and the corresponding penalized estimator $b^*$ under validation and cross-validation.
%
%%%%%%%%%%%%%%%%%%%%%%%%%%%%%%%%%%%%%%%%%%%%%%%%%%%%%%%%%%%%%%%%%%%%%%%%%%%%%%%%%%%%%%%%%%%%
% Proposition 3.3
%%%%%%%%%%%%%%%%%%%%%%%%%%%%%%%%%%%%%%%%%%%%%%%%%%%%%%%%%%%%%%%%%%%%%%%%%%%%%%%%%%%%%%%%%%%%
%
\begin{prop}[$\mathcal{L}_2$-difference between the penalized and unpenalized predicted values]
Under A1--A6 and based on Lemma~\ref{lem3.1}, Propositions~\ref{prop3.1}, and~\ref{prop3.2},

	\begin{enumerate}

        \item \textbf{Validation case.} The following bound for the difference between the predicted values from $b_{OLS}$ and the validated $b^*$ holds with probability at least $\varpi(1-1/n_{t})$
			\begin{equation}
				\frac{1}{n_{s}}(\Vert X_{s}b_{OLS}-Xb^*\Vert _{2})^{2}
				\leqslant
				\left(\frac{1}{n_{t}}\frac{\Vert e_{t}\Vert_{2}^{2}}{1-\sqrt{\epsilon}}
				-\frac{1}{n_{s}}\Vert e_{s}\Vert _{2}^{2}\right)
				+\frac{4}{n_{s}}\Vert e_{s}^{T}X_{s}\Vert_{\infty}\Vert b_{OLS}\Vert _{1}+\varsigma
				\label{eq1:prop3.3}
			\end{equation}
            where $\varsigma$ is defined in Theorem~\ref{thm2.1}.

        \item \textbf{$K$-fold cross-validation case.} The following bound for the difference between the predicted values from the $K$-fold cross-validated $b_{OLS}$ and $b^*$ holds with probability at least ${\varpi}(1-1/n_{t})$
		      \begin{eqnarray}
                \frac{1}{K}\sum_{q=1}^K
                \frac{1}{n_s}(\Vert X_s^q b_{OLS}^q -X_s^q b^{*q} \Vert_{2})^{2}
		      	& \leqslant &
                \left\vert\frac{1}{n_t}\frac{\frac{1}{K}\sum_{q=1}^K
                \left\Vert e_t^q \right\Vert_{2}^{2}}{1-\sqrt{ \epsilon}} -
		      	\frac{1}{K}\sum_{q=1}^K\frac{1}{n_s}
                \left\Vert e_s^q \right\Vert_{2}^{2}\right\vert \\ \notag
		      	& + &
                \frac{1}{K}\sum_{q=1}^K
                \frac{4}{n_s}\left\Vert\left( e_s^q \right)^TX_s^q\right\Vert_{\infty}		      	\left\Vert b_{OLS}^q\right\Vert_{1} +
                \varsigma.
		      	\label{eq2:prop3.3}
		      \end{eqnarray}
            where $b_{OLS}^q$ is the OLS estimator and $b^{*q}$ is the penalized estimator in the $q$th round of cross-validation, and $\varsigma$ is defined in Theorem~\ref{thm2.2}.

	\end{enumerate}

\label{prop3.3}
\end{prop}

Following Markov's classical proof of consistency for OLS, Proposition~\ref{prop3.3} establishes the $\mathcal{L}_2$-norm convergence of the fitted values to the true values. Based on A1-A6, the identification condition is satisfied, and the convergence of the fitted values implies the $\mathcal{L}_2$-norm consistency of the penalized regression estimator.

We now establish the upper bound of the $\mathcal{L}_2$-norm difference between $b_{OLS}$ and $b^*$, or  $\Vert b_{OLS} - b^*\Vert_2$, under validation and cross-validation.
%
%%%%%%%%%%%%%%%%%%%%%%%%%%%%%%%%%%%%%%%%%%%%%%%%%%%%%%%%%%%%%%%%%%%%%%%%%%%%%%%%%%%%%%%%%%%%
% Theorem 3.1
%%%%%%%%%%%%%%%%%%%%%%%%%%%%%%%%%%%%%%%%%%%%%%%%%%%%%%%%%%%%%%%%%%%%%%%%%%%%%%%%%%%%%%%%%%%%
%
\begin{thm}[$\mathcal{L}_2$-difference between the penalized and unpenalized regression estimators]
Under A1--A6 and based on Propositions~\ref{prop3.1}, \ref{prop3.2}, and \ref{prop3.3},

	\begin{enumerate}

        \item   \textbf{Validation case.} The following bound for the $\mathcal{L}_2$-difference between $b_{OLS}$ and the validated $b^*$ holds with probability at least $\varpi(1-1/n_{t})$
		      \begin{equation}
		      	\Vert b_{OLS}-b^*\Vert _{2} \leqslant
		      	\sqrt{\left\vert\frac{1}{\rho n_{t}}\frac{\Vert e_{t}\Vert_{2}^{2}}{(1-\sqrt{\epsilon})}
		      		-\frac{1}{\rho n_{s}}\Vert e_{s}\Vert_{2}^{2}\right\vert}
		      	+\sqrt{\frac{4}{\rho n_{s}}\Vert e_{s}^{T}X_{s}\Vert_{\infty}\Vert b_{OLS}\Vert_{1}}
		      	+ \left(\frac{\varsigma}{\rho}\right)^{\frac{1}{2}}
		      	\label{eq1:thm3.1}
		      \end{equation}
              where $\rho$ is the minimum eigenvalue of $X^{T}X$ and $\varsigma$ is defined in Lemma~\ref{lem3.1}.

        \item   \textbf{$K$-fold cross-validation case.} The following bound for the $\mathcal{L}_2$- difference between the $K$-fold cross-validated $b_{OLS}$ and $b^*$ holds with probability at least $\varpi(1-1/n_{t})$
		      \begin{eqnarray}
		      	\frac{1}{K}\sum_{q=1}^K
                (\Vert b_{OLS}^q - b^{*q} \Vert _{2})^{2}
                & \leqslant	&     	
		      	\left\vert
                \frac{1}{K}\sum_{q=1}^K \frac{1}{n_t{\rho^*}}
	      	    \frac{\left\Vert e_t^q \right\Vert_{2}^{2}}{1-\sqrt{\epsilon}}
		      	- \frac{1}{K}\sum_{q=1}^K \frac{1}{n_s{\rho^*} }
		      	\left\Vert e_s^q \right\Vert_{2}^{2}
                \right\vert
		      	\notag \\
		      	& + &
                \frac{1}{K}\sum_{q=1}^K \frac{4}{ n_s {\rho^*} }
		      	\left\Vert (e_s^q)^TX_s^q \right\Vert_{\infty}
		      	\left\Vert b_{OLS}^q \right\Vert_{1}
		      	+\frac{\varsigma}{{\rho^*}}
		      	\label{eq2:thm3.1}
		      \end{eqnarray}
              where ${\rho^*}$ is defined $\min_q\{\rho_q\;\vert\;\rho_q \mbox{ is the minimum eigenvalue of} \left(X^q_s\right)^{T}X^q_s, \mbox{ given } q\}$.

    \end{enumerate}

\label{thm3.1}
\end{thm}

Some important remarks apply to Theorem~\ref{thm3.1}. The LHS of eq.~(\ref{eq1:thm3.1}) measures the  $\mathcal{L}_2$-norm difference between the penalized regression estimator and the OLS estimator under validation. The RHS of eq.~(\ref{eq1:thm3.1}) essentially captures the \emph{maximum} $\mathcal{L}_2$-norm difference between $b_{OLS}$ and $b^*$. As shown in eq.~(\ref{eq1:thm3.1}), the maximum difference depends on the GE of the true DGP and the GE of the OLS model.

\begin{itemize}

    \item   The first term on the RHS of eq.~(\ref{eq1:thm3.1}) (ignoring $1/\rho$) is the difference between the eGE from OLS and the upper bound of the population error, or, equivalently, \emph{the difference between the GA of the OLS estimator and its maximum}. The better the GA of $b_{OLS}$, the less overfitting OLS generates, the closer the eGE of $b_{OLS}$ is to the upper bound of the population error, and the smaller the first term on the RHS of eq.~(\ref{eq1:thm3.1}).

    \item   The second term on the RHS of eq.~(\ref{eq1:thm3.1}) (ignoring $4/\rho$) measures the empirical endogeneity of the OLS estimator on the test set. On the training set $e_t^T X_s = 0$, but on the test set, in general, $e_s^T X_s \neq 0$. Hence, $\frac{1}{n_{s}}\Vert e_{s}^{T}X_{s}\Vert_{\infty}\Vert b_{OLS}\Vert_{1}$ measures \emph{the GA for the empirical moment condition of the OLS estimator on out-of-sample data}.\footnote{Because we standardize the test and training data, the moment condition $\mathbb{E}(e_s)=0$ holds directly.} The more generalizable the OLS estimate, the closer $e_s^T X_s$ is to zero on out-of-sample data, and the smaller the second term on the RHS of eq.~(\ref{eq1:thm3.1}).

    \item   The third term on the RHS of eq.~(\ref{eq1:thm3.1}) is affected by $\varsigma$, which measures \emph{the heaviness of the tail in the distribution of the loss function of the OLS estimator}. Similar to the comments on Theorem~\ref{thm2.1}, the OLS loss distribution affects the GA of the OLS estimator. The heavier the loss function distribution tail, the more volatile the eGE on out-of-sample data, and the more difficult to bound the eGE for OLS.

    \item   All three RHS terms in eq.~(\ref{eq1:thm3.1}) are affected by $\rho$, the minimum eigenvalue of $X^TX$, which can be thought of as a measure \emph{of the curvature of the objective function~(\ref{lasso-type}) for penalized regression}. The larger the minimum eigenvalue, the more convex the objective function. Put another way, it is easier to identify the true DGP from the alternatives as $n$ get larger.

\end{itemize}

The interpretation of eq.~(\ref{eq2:thm3.1}) is similar to eq.~(\ref{eq1:thm3.1}) adjusting for cross-validation. Hence, the first term on the RHS of eq.~(\ref{eq2:thm3.1}) (ignoring $1/{\rho^*}$) stands for how far away the \textit{average} GA of OLS estimator is from its maximum in $K$ rounds of validation. The second term on the RHS of eq.~(\ref{eq2:thm3.1}) (ignoring $4/{\rho^*}$) indicates \textit{on average} how generalizable the empirical moment condition of the OLS estimator is with out-of-sample data in $K$ rounds of validation. Similarly, ${\varsigma}$ indicates \textit{on average} the heaviness in the tail of the loss distribution in $K$ rounds of validation. As a direct result of Theorem~\ref{thm3.1}, the $\mathcal{L}_2$-consistency for the penalized regression estimate is established as follows.
%
%%%%%%%%%%%%%%%%%%%%%%%%%%%%%%%%%%%%%%%%%%%%%%%%%%%%%%%%%%%%%%%%%%%%%%%%%%%%%%%%%%%%%%%%%%%%
% Corollary 3.2
%%%%%%%%%%%%%%%%%%%%%%%%%%%%%%%%%%%%%%%%%%%%%%%%%%%%%%%%%%%%%%%%%%%%%%%%%%%%%%%%%%%%%%%%%%%%
%
\begin{cor}[$\mathcal{L}_2$-consistency of the penalized regression estimator when $n\geqslant p$]
Under A1--A6 and based on Propositions~\ref{prop3.1}, \ref{prop3.2} and \ref{prop3.3}, $b^*$ converges in the $\mathcal{L}_{2}$-norm to the true DGP if \textup{$\lim_{n\rightarrow\infty}p/\widetilde{n}=0$}.

\label{cor3.2}
\end{cor}

\begin{figure}
	\centering
	\subfloat[\label{figillustrate:fig1}$b_{OLS}$ and $b^*$ under an $\mathcal{L}_{1}$ penalty]
	{\includegraphics[width=0.33\paperwidth]{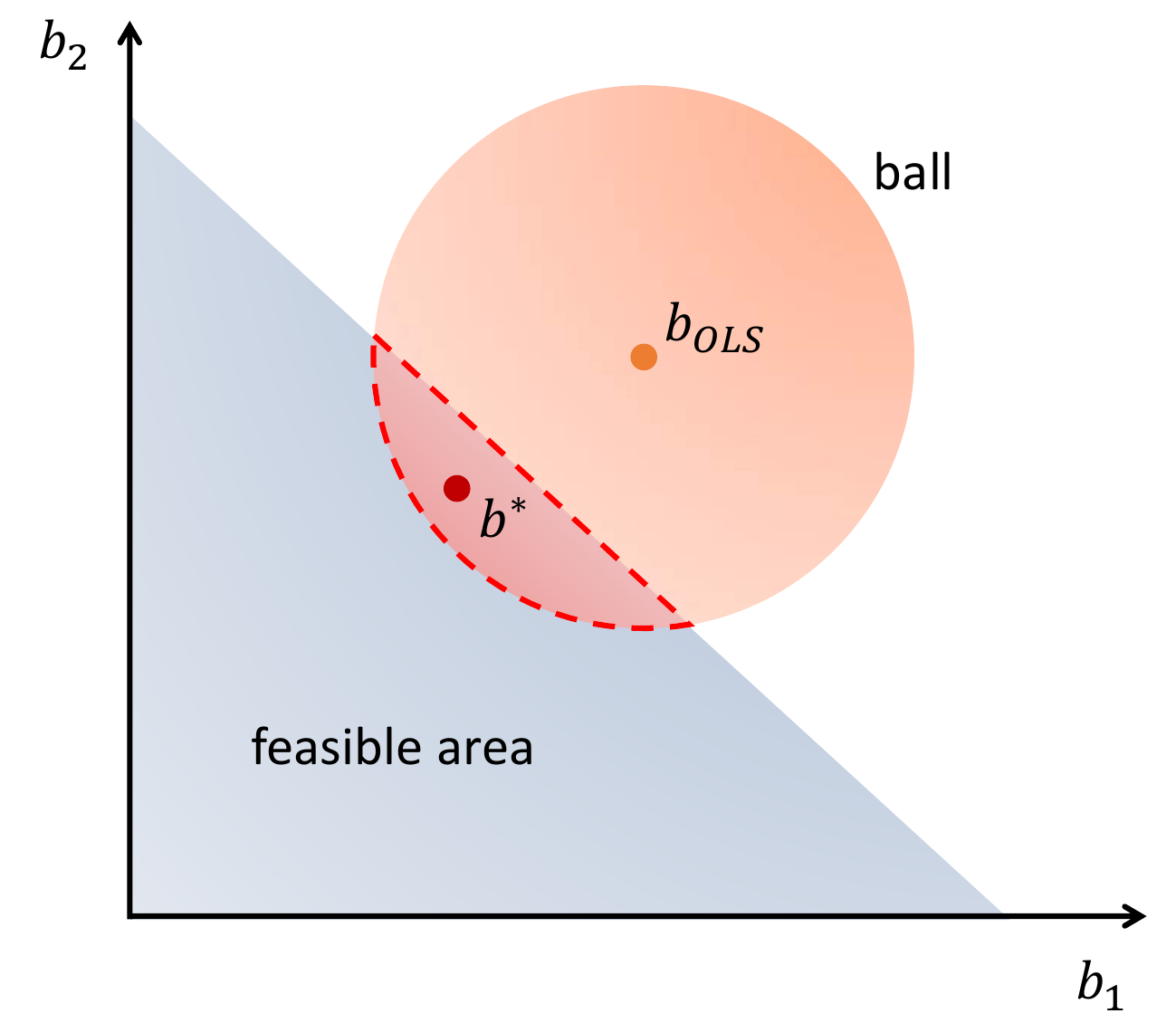}}
	\subfloat[\label{figillustrate:fig2}$b_{OLS}$ and $b^*$ convergence as $n$ increases]
	{\includegraphics[width=0.33\paperwidth]{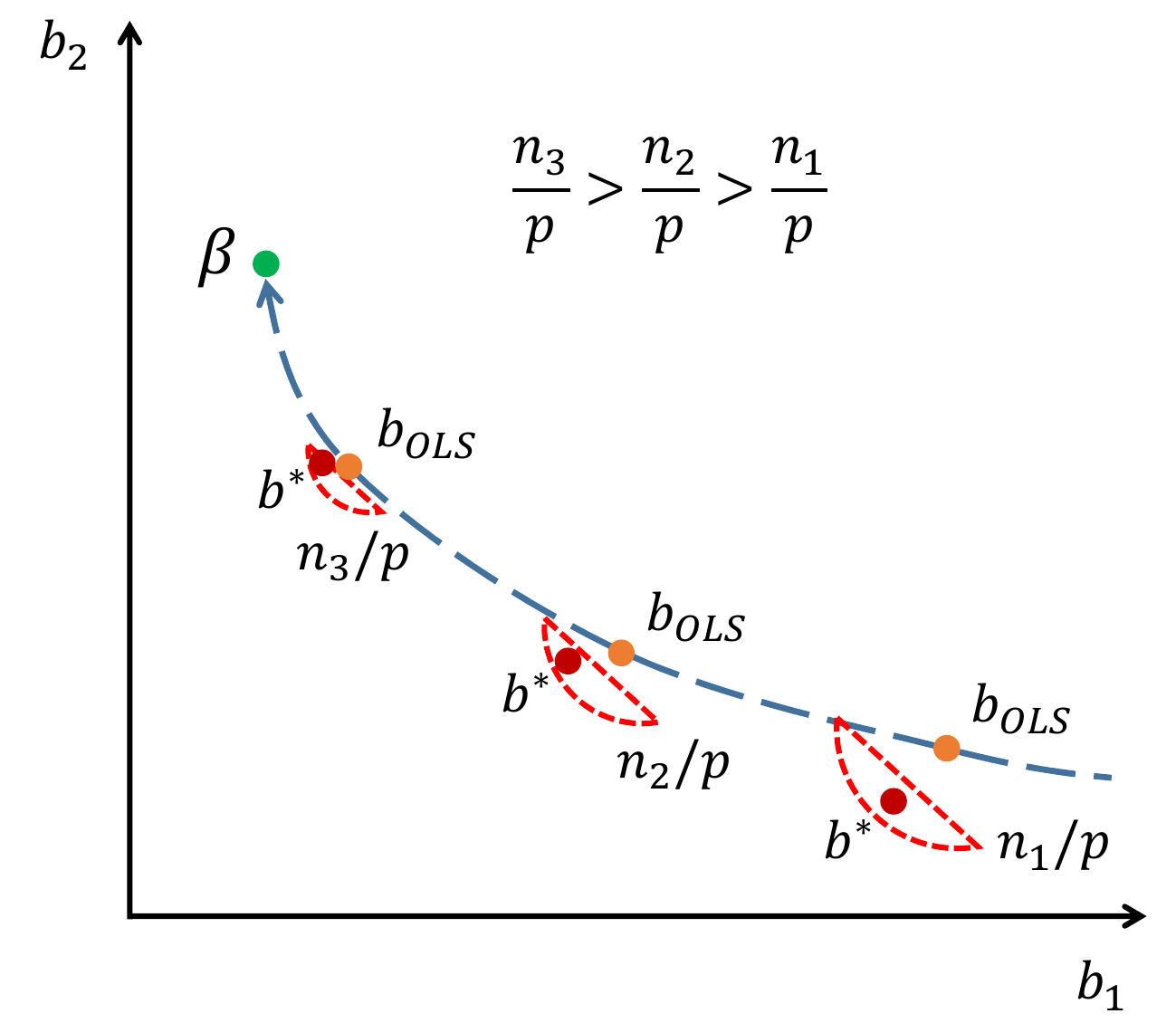}}

    \caption{\label{figillustrate}The relation between $b_{OLS}$ and $b^*$ under an $\mathcal{L}_{1}$ penalty}
\end{figure}

Figure~\ref{figillustrate} illustrates Theorem~\ref{thm3.1} and Corollary~\ref{cor3.2} for the lasso. Due to the poor GA of the OLS estimator, the penalized regression estimator $b^*$ will not usually lie on the same convergence path as the OLS estimator. However, Theorem~\ref{thm3.1} shows that the deviation of $b^*$ from the OLS convergence path is bounded: $b^*$ typically lies within a ball centered around $b_{OLS}$ whose radius is a function of the eGEs of the OLS estimator and the true DGP. Also, as shown in Figures~\ref{figlassoshrink:fig1} and \ref{figlassoshrink:fig2}, $b^*$ always lies within the feasible set parameterized by $\lambda$. Hence, as shown in Figures~\ref{figillustrate:fig1} and \ref{figillustrate:fig2}, $b^*$ typically is located in the small area at the intersection of the $\mathcal{L}_1$ feasible area and the ball. Unless the optimal $\lambda$ from validation or cross-validation is $0$, the OLS estimate will never be in the feasible area for penalized regression, which is why the intersection region is always below $b_{OLS}$. As $n/p$ increases, the ball becomes smaller, the penalized regression estimate gets closer to the OLS estimator, and both converge to $\beta$.

\subsubsection{GEM for penalized regression with $n < p$}

Typically, to ensure $b_{OLS}$ can identify the true DGP $\beta$, we require that $(\Vert e\Vert_2)^2$ is strongly convex in $b$, or that the minimal eigenvalue of $X^T X$, $\rho$, is strictly larger than 0. However, if $p>n$, $\rho=0$ and the space of $(\Vert e\Vert_2)^2$ is flat in some direction. As a result, the $\Vert b_{OLS}\Vert_1$ is not of closed-form, the true DGP cannot be identified and eqs.~(\ref{eq1:thm3.1}) and (\ref{eq2:thm3.1}) are trivial.

To establish results that are non-trivial, we need to ensure $\beta$ is identifiable when $p>n$. Put another way, we need to ensure that the strong convexity of the space $(\Vert e\Vert_2)^2$ is maintained for the $p>n$ case. This is guaranteed by the \textbf{restricted eigenvalue condition} \citep{bickeletal09, meinshausenyu09, zhang10}---see the proof of Proposition~\ref{prop3.4} (below) in Appendix~A for the details.

Regression can at most estimate $n$ coefficients. When $p > n$, penalized regression has to drop some variables to make it estimable, implying that a penalty of $\gamma>1$ does not apply to the $p > n$ case. Hence, for the $p>n$ case, we focus only on $\mathcal{L}_{1}$ penalized regression, i.e., lasso. As shown by \citet{efronall04} and \citet{zhang10}, lasso may be thought of as a forward stagewise regression (FSR) with an $\mathcal{L}_1$-norm constraint.\footnote{The method of solving lasso by forward selection is the least angle regression (LARS). For details of LARS and its consistency, see \citet{efronall04} and \citet{zhang10}.} Hence, lasso regression can be viewed as a way to control the eGE of FSR when $p>n$. As shown in \citet{zhang10}, while FSR may result in overfitting in finite samples, it is $\mathcal{L}_2$-consistent under the restricted eigenvalue condition.

Thus, for $p>n$, we use FSR, $b_{FSR}$, as the unpenalized regression estimator, and the lasso, $b^*$, as the penalized regression estimator. In Proposition~\ref{prop3.4} and Corollary~\ref{cor3.3}, we show that lasso preserves the properties and interpretations of the $n\geqslant p$ case by reducing the overfitting inherent in FSR.
%
%%%%%%%%%%%%%%%%%%%%%%%%%%%%%%%%%%%%%%%%%%%%%%%%%%%%%%%%%%%%%%%%%%%%%%%%%%%%%%%%%%%%%%%%%%%%
% Proposition 3.4
%%%%%%%%%%%%%%%%%%%%%%%%%%%%%%%%%%%%%%%%%%%%%%%%%%%%%%%%%%%%%%%%%%%%%%%%%%%%%%%%%%%%%
%
\begin{prop}[$\mathcal{L}_2$-difference between the $\mathcal{L}_1$-penalized and unpenalized FSR estimators]
Under A1--A6 and the restricted eigenvalue condition, and based on Lemma~\ref{lem3.1}, Propositions~\ref{prop3.1}, and~\ref{prop3.2},

	\begin{enumerate}

        \item \textbf{Validation case.} The following bound holds with probability at least $\varpi(1-1/n_{t})$
		    \begin{eqnarray}
		      	\Vert b_{FSR}-b^*\Vert_{2}
                & \leqslant &
		      	\sqrt{\left|\frac{1}{\rho_{re}n_{t}}
                \frac{\Vert e_{t}\Vert_{2}^{2}}{(1-\sqrt{\epsilon})} -
                \frac{1}{\rho_{re}n_{s}}\Vert e_{s}\Vert_{2}^{2}\right|} \nonumber \\
		      	& + &
                \sqrt{\frac{4}{\rho_{re}n_{s}}
                \Vert e_{s}^{T}X_{s}\Vert_{\infty}\Vert b_{FSR}\Vert_{1}} + \left(\frac{\varsigma}{\rho_{re}}\right)^{\frac{1}{2}}
		      	\label{eq1:prop3.4}
		    \end{eqnarray}
            where $\rho_{re}$ is the minimum restricted eigenvalue of $X^{T}X$ and $b_{FSR}$ is the FSR estimator.

        \item \textbf{$K$-fold cross-validation case.} The following bound holds with probability at least $\varpi(1-1/n_{t})$
		    \begin{eqnarray}
		    	\frac{1}{K}\sum_{q=1}^K \left\Vert b_{FSR}^q - b^{*q}\right\Vert _{2}^{2}
                & \leqslant &
		      	\left\vert \frac{1}{K}\sum_{q=1}^K \frac{1}{n_t{\rho_{re}^*}}
		      	\frac{\left\Vert e_t^q \right\Vert_{2}^{2}}{1-\sqrt{\epsilon}}
		      	- \frac{1}{K}\sum_{q=1}^K \frac{1}{n_s{\rho_{re}^*}}
		      	\left\Vert e_s^q \right\Vert_{2}^{2}\right\vert \notag \\
		      	& + &
                \frac{1}{K}\sum_{q=1}^K\frac{4}{n_s{\rho_{re}^*}}
		      	\left\Vert \left( e_s^q\right)^TX_s^q\right\Vert_{\infty}
                \left\Vert b_{FSR}\right\Vert_{1}
		      	+ \frac{\varsigma}{{\rho_{re}^*}}
		      	\label{eq2:prop3.4}
		    \end{eqnarray}
            where ${\rho_{re}^*}$ is defined $\min_q\{\rho_{re}^q~\vert~\rho_{re}^q \mbox{ is the minimum restricted eigenvalue of }\left(X^q_s\right)^{T}X^q_s, \mbox{ given } q\}$ and  $b_{FSR}^q$ is the FSR estimator in the $q$th round of cross-validation.

	\end{enumerate}

\label{prop3.4}
\end{prop}

\begin{figure}[ht]
	\centering
	\includegraphics[width=0.4\paperwidth]{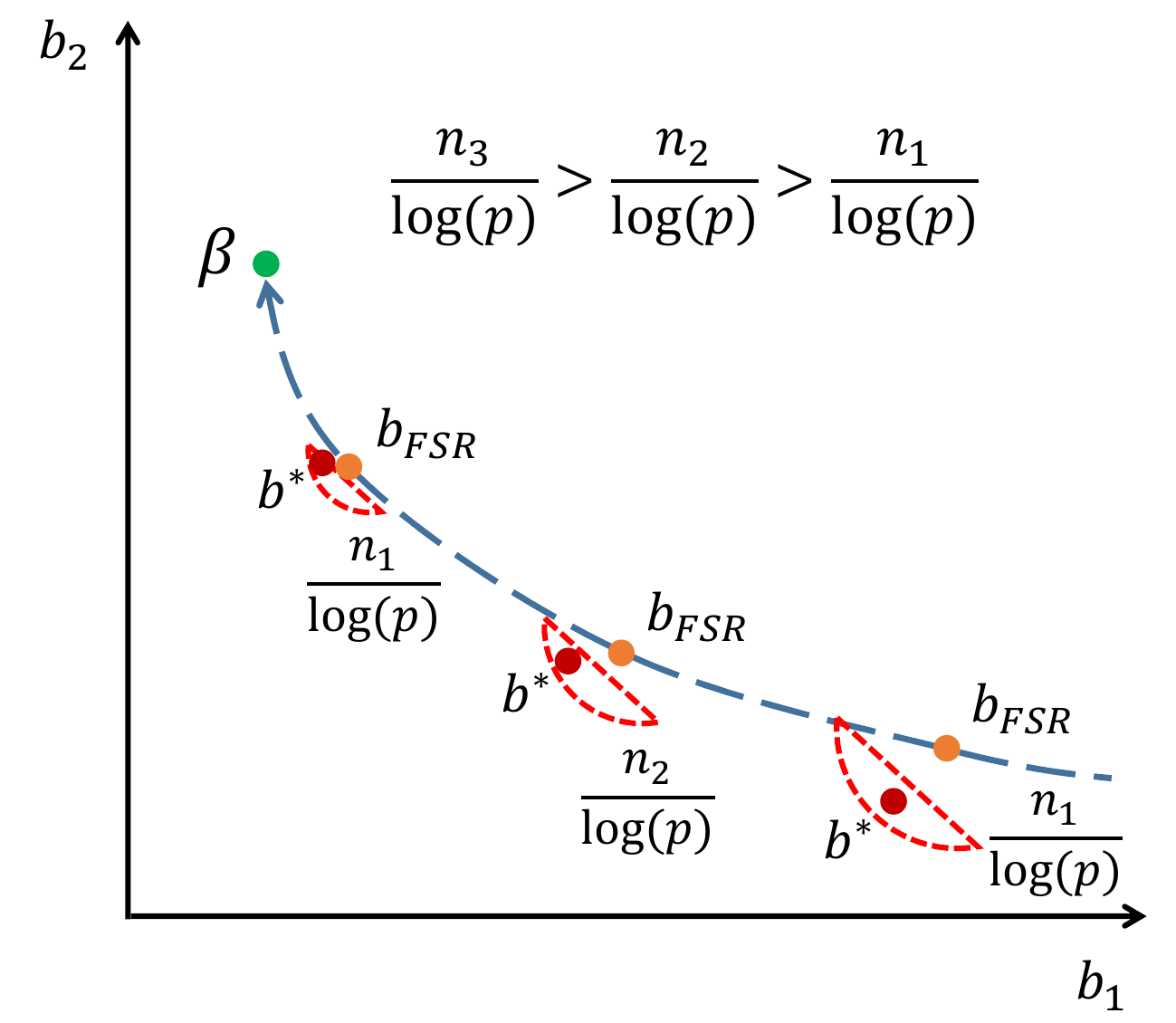}

    \caption{\label{figillustrateFSR}Convergence of $b_{FSR}$ and $b^*$ as $n$ increases}
\end{figure}

%
%%%%%%%%%%%%%%%%%%%%%%%%%%%%%%%%%%%%%%%%%%%%%%%%%%%%%%%%%%%%%%%%%%%%%%%%%%%%%%%%%%%%%%%%%%%%
% Corollary 3.3
%%%%%%%%%%%%%%%%%%%%%%%%%%%%%%%%%%%%%%%%%%%%%%%%%%%%%%%%%%%%%%%%%%%%%%%%%%%%%%%%%%%%%%%%%%%%
%
\begin{cor}[$\mathcal{L}_2$-consistency of the $\mathcal{L}_1$-penalized regression estimator]
Under A1--A6 and the restricted eigenvalue condition, and based on Propositions~\ref{prop3.1}, \ref{prop3.2}, \ref{prop3.4} $b^*$ converges in the $\mathcal{L}_{2}$-norm asymptotically to the true DGP if \textup{$\lim_{n\rightarrow\infty}\log(p)/\widetilde{n}=0$}.

\label{cor3.3}
\end{cor}

The interpretation of Proposition~\ref{prop3.4} and Corollary~\ref{cor3.3}, which specify the upper bound of $\mathcal{L}_2$-difference between $b_{FSR}$ and $b^*$ as a function of the GA of the FSR estimator and the population error, is very similar to that of eqs.~(\ref{eq1:thm3.1}) and (\ref{eq2:thm3.1}). Thus, Figure~\ref{figillustrateFSR} illustrates Proposition~\ref{prop3.4} and Corollary~\ref{cor3.3} in a similar fashion.

%%%%%%%%%%%%%%%%%%%%%%%%%%%%%%%%%%%%%%%%%%%%%%%%%%%%%%%%%%%%%%%%%%%%%%%%%%%%%%%%%%%%%%%%%%%%%%%%%%%%%%%%%%%
%%%%%%%%%%%%%
%%%%%%%%%%%%%            SECTION 4
%%%%%%%%%%%%%
%%%%%%%%%%%%%%%%%%%%%%%%%%%%%%%%%%%%%%%%%%%%%%%%%%%%%%%%%%%%%%%%%%%%%%%%%%%%%%%%%%%%%%%%%%%%%%%%%%%%%%%%%%%

\section{Simulations}

In Sections~2 and~3, we use eTE to measure in-sample fit and eGE to measure out-of-sample fit. However, to measure GA and the degree of overfitting, we need to compare the eTE and eGE to the total sums of squares for the training set and test set, respectively. Thus, to summarize the in-sample and out-of-sample goodness of fit, we propose the following generalized $R^2$ measure:
\begin{equation}
	GR^{2} =
	\left(1 - \frac{\mathcal{R}_{n_s}(b_{train}|Y_{s},X_{s})}{\mathrm{TSS}\left(Y_{s}\right)}\right)
	\times
	\left(1 - \frac{\mathcal{R}_{n_t}(b_{train}|Y_{t},X_{t})}{\mathrm{TSS}\left(Y_{t}\right)}\right) \\
	= R^2_s \times R^2_t
	\label{grsq}
\end{equation}
where $R^2_s$ is the $R^2$ for the test set and $R^2_t$ for the training set. Clearly $GR^2$ is a summary measure of in-sample and out-of-sample fit. Provided $b_{train}$ is consistent, both $\mathcal{R}_{n_s}(b_{train}|Y_{s},X_{s})$ and $\mathcal{R}_{n_t}(b_{train}|Y_{t},X_{t})$ converge to the same limit in probability as $\widetilde{n}\rightarrow\infty$.

\begin{table}[h!]
	\centering
	\caption{Four stylized scenarios for $GR^2$}
	\begin{tabular}{l@{\hskip 10mm}l@{\hskip 5mm}rcl}
        \toprule
		        &      &                       & $R^2_s$ \\
                &      & \emph{high}                  &         & \emph{low} \\[8pt]
		 & \emph{high} & high $GR^2$ (ideal model)    &         & relatively low $GR^2$ (overfitting) \\
        $R^2_t$ \\[-4pt]
		 & \emph{low}  & relatively low $GR^2$ (rare) &         & very low $GR^2$ (underfitting)      \\
        \bottomrule
	\end{tabular}
	\label{table1}
\end{table}

Table~\ref{table1} summarizes four basic scenarios for $GR^2$. A model that fits both the training set and the test set well will have high $R^2_t$ and $R^2_s$ values and hence a high $GR^2$. When overfitting occurs, the $R^2_t$ will be relatively high and the $R^2_s$ will be low, reducing the $GR^2$. When underfitting occurs, the $R^2_t$ and $R^2_s$ will be low, reducing the $GR^2$ even further. It is unlikely, but possible, that the model estimated on the training set fits the test set better (the $R^2_s$ is high while the $R^2_t$ low).

In the simulations, we illustrate the result that penalized regression, by constraining the $\mathcal{L}_\gamma$-norm of the estimate, produces a superior GA compared with OLS or FSR. As illustrated in Figures~\ref{figpenreg} and~\ref{figillustrate}, penalized regression is less efficient at model selection when the norm of penalty term $\gamma > 1$. Thus, we focus on the $\mathcal{L}_1$-penalized or lasso-type regression.

For the simulations, we assume the outcome $y$ is generated by the following DGP:
\[
	y = X\beta+u = X_{1}\beta_{1} + X_{2}\beta_{2} + u
\]
where $X=\left(x_{1},\cdots,x_{p}\right)\in\mathbb{R}^{p}$ is generated by a multivariate normal distribution with zero mean, $\mbox{var}(x_{i}) = 1$, $\mbox{corr}(x_{i},x_{j}) = 0.9, \forall i,\thinspace j$, $\beta_{1}=\left(2,4,6,8,10,12\right)^T$ and $\beta_{2}$ is a $(p - 6)$-dimensional zero vector. $u$ is generated by a normal distribution with zero mean and variance $\sigma^2$. Here $x_{i}$ does not cause $x_{j}$ and there is no causal relationship between $u$ and $x_{i}$. We set the sample size at 250, $p$ is set at 200 or 500 and $\sigma^2$ at 1 or 5. Hence, we have four different cases. In each case, we repeat the simulation 50 times. In each simulation, we apply Algorithm~1 to find the estimate of $\beta$ and calculate its distance to the true value, the eGE, as well as our goodness-of-fit measure $GR^{2}$. As a comparison, we also apply OLS in the $n\geqslant p$ cases and the FSR algorithm for the $n<p$ cases.

Boxplots (see Appendix~B) summarize the estimates of all the coefficients in $\beta_{1}$ (labeled $b_1$ to $b_6$) along with the four worst estimates among the coefficients in $\beta_{2}$ (labeled $b_7$ to $b_{10}$), where `worst' refers to the estimates with the largest bias. The lasso and OLS/FSR estimates and histograms of the $GR^{2}$ are reported for each case in Figures~9--12. Lastly, the distance between the estimates and the true values, the eGE, and the $GR^{2}$ (all averages across the 50 simulations) are reported in Table~\ref{table1} for all four cases.

\begin{figure}[b]
\section*{Appendix B: Simulation plots}
	\centering
	\subfloat[Lasso estimates]
	{\includegraphics[width=0.33\paperwidth]{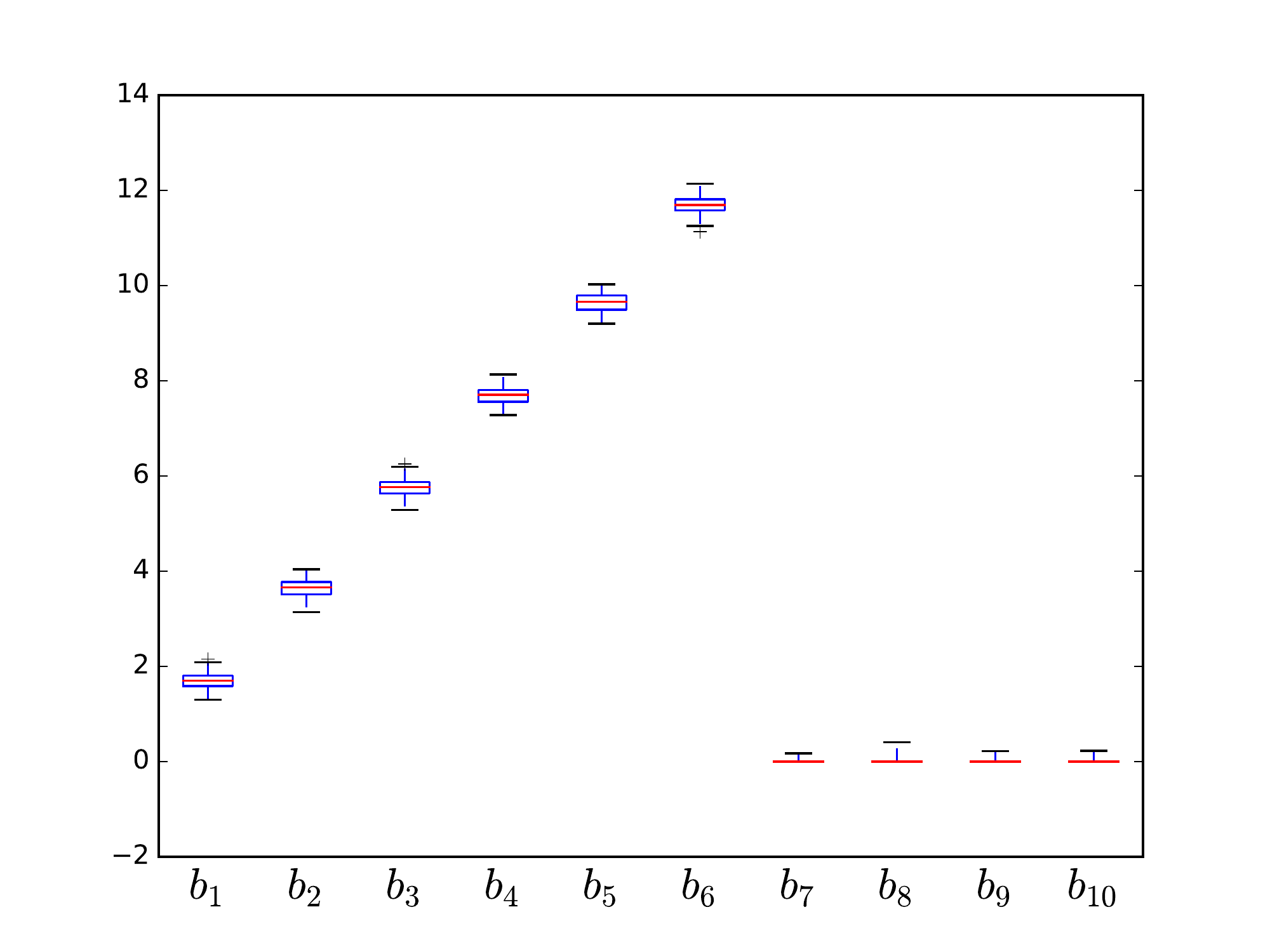}}
	\subfloat[OLS estimates]
	{\includegraphics[width=0.33\paperwidth]{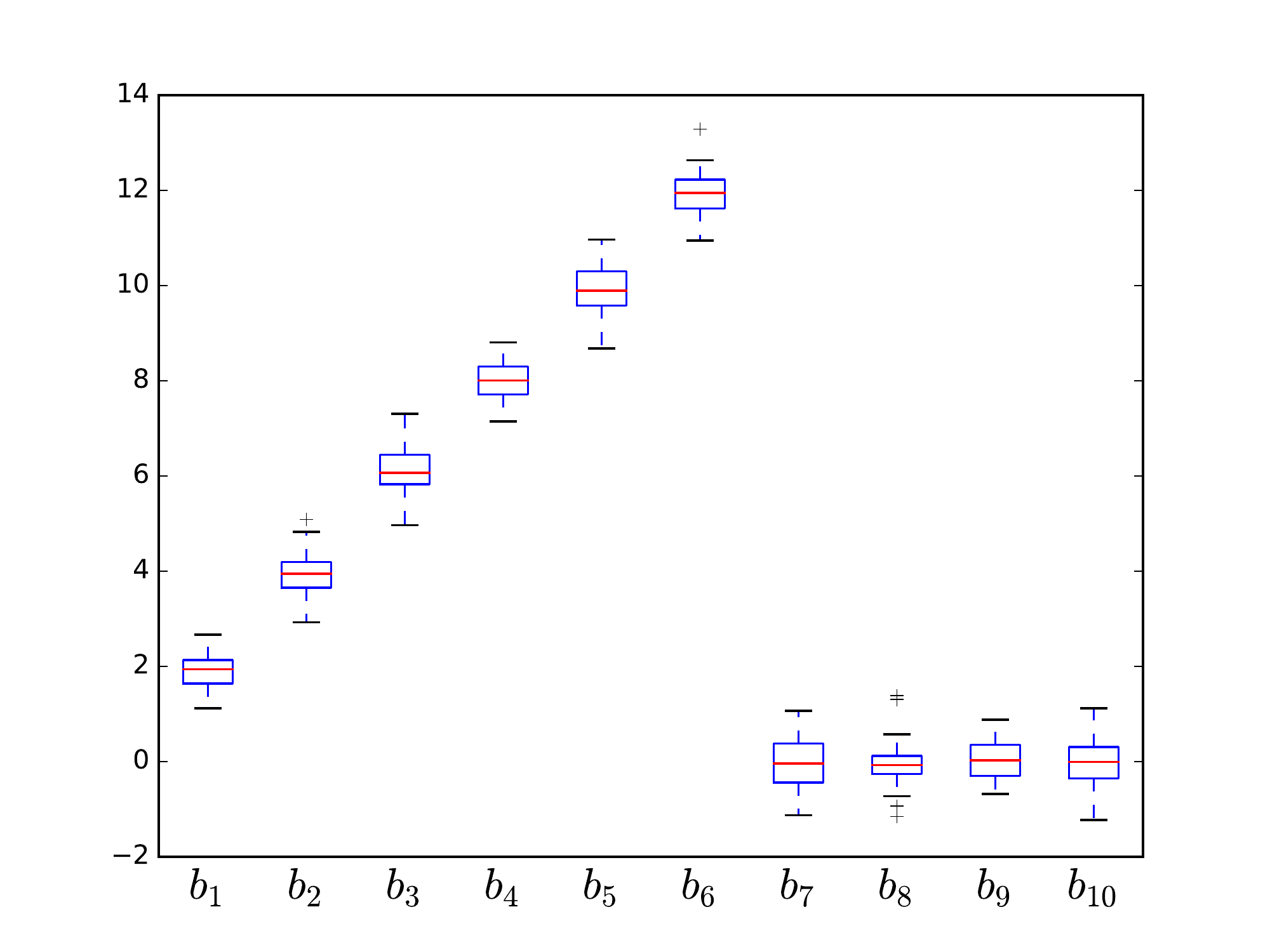}}

    \subfloat[Histogram of $GR^{2}$]
	{\includegraphics[width=0.6\paperwidth]{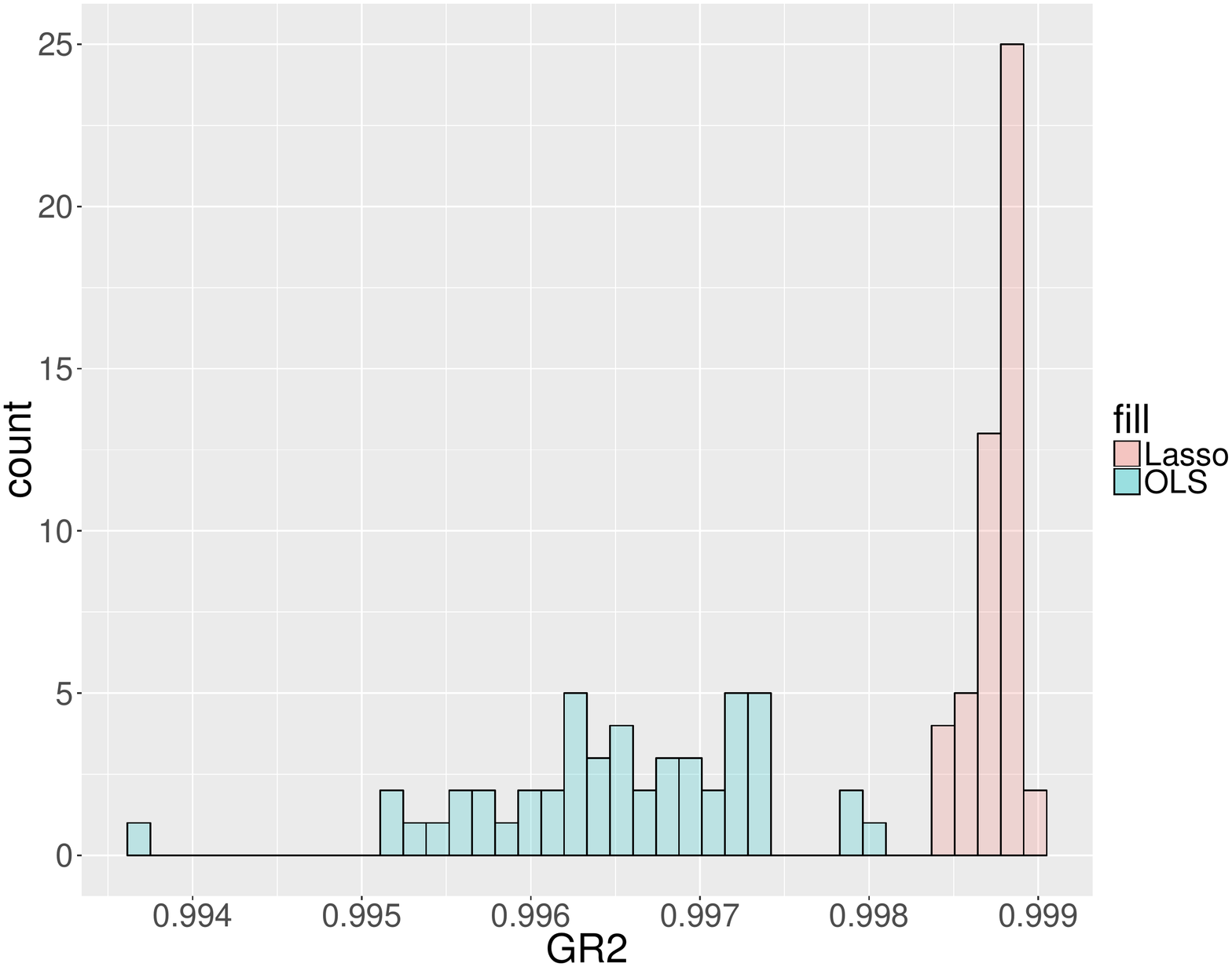}}

	\caption{\label{p=200v=1}DGP with $n = 250$, $p = 200$, and $\sigma^2 = 1$.}
\end{figure}

When $n>p=200$, as we can see from the boxplots in Figures~\ref{p=200v=1} and \ref{p=200v=5}, both lasso and OLS perform well at estimating $\beta_1$: all the coefficient estimates are centered around the true values and the deviations are relatively small. As expected, both perform better with $\sigma^2=1$ (Figure~\ref{p=200v=1}) compared with $\sigma^2=5$ (Figure~\ref{p=200v=5}). Lasso clearly outperforms OLS for the estimates of $\beta_{2}$ in terms of having much smaller deviations. Indeed, a joint significance test ($F$ test) fails to reject the null hypothesis that all coefficients in $\beta_{2}$ are zero for the OLS estimates. As shown in Figures~\ref{p=200v=1} and \ref{p=200v=5}, the $GR^{2}$ for the lasso is marginally larger than for OLS, but the differences are inconsequential.

\begin{figure}[ht]
	\centering
	\subfloat[Lasso estimates]
	{\includegraphics[width=0.33\paperwidth]{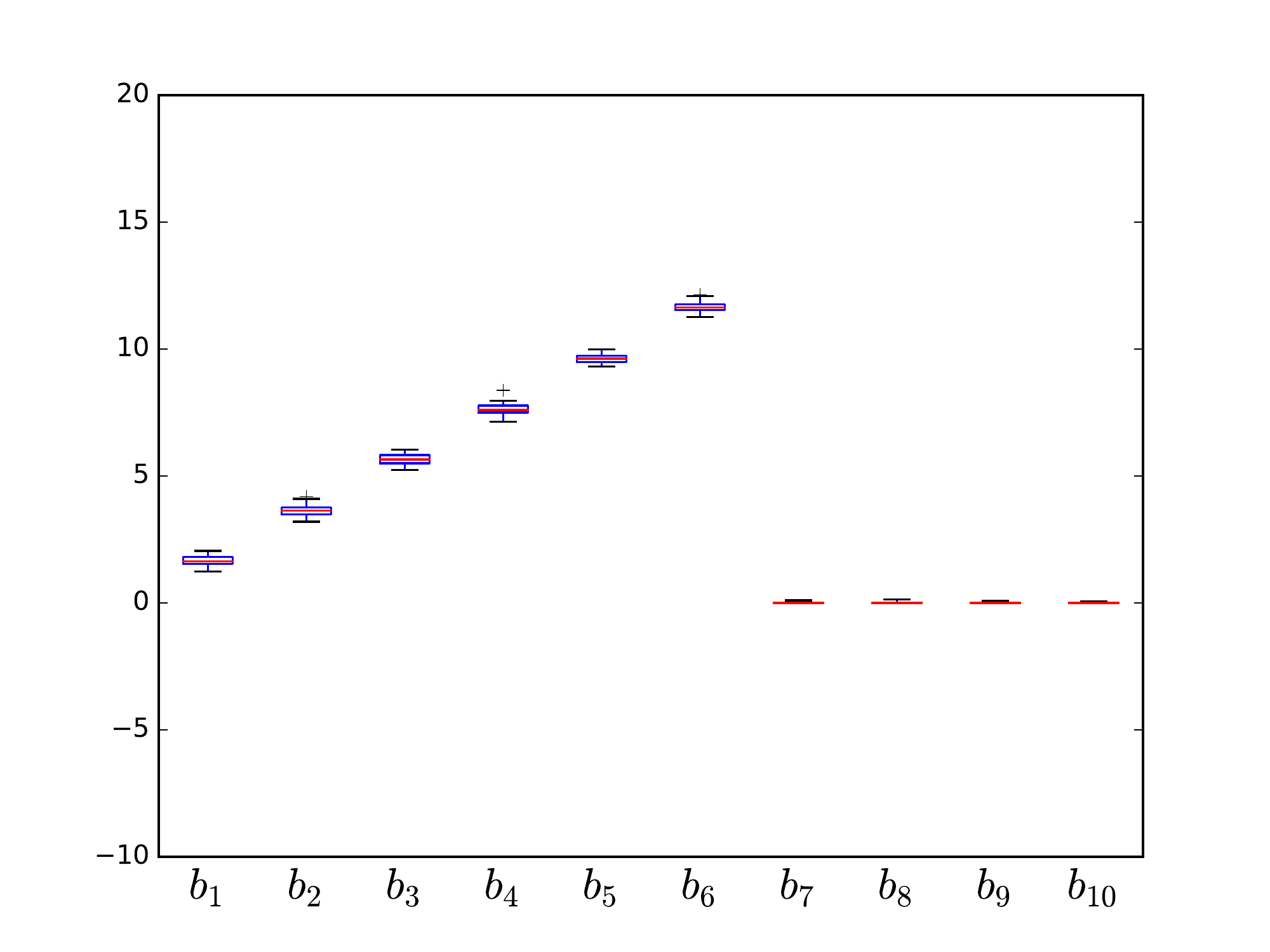}}
	\subfloat[FSR estimates]
	{\includegraphics[width=0.33\paperwidth]{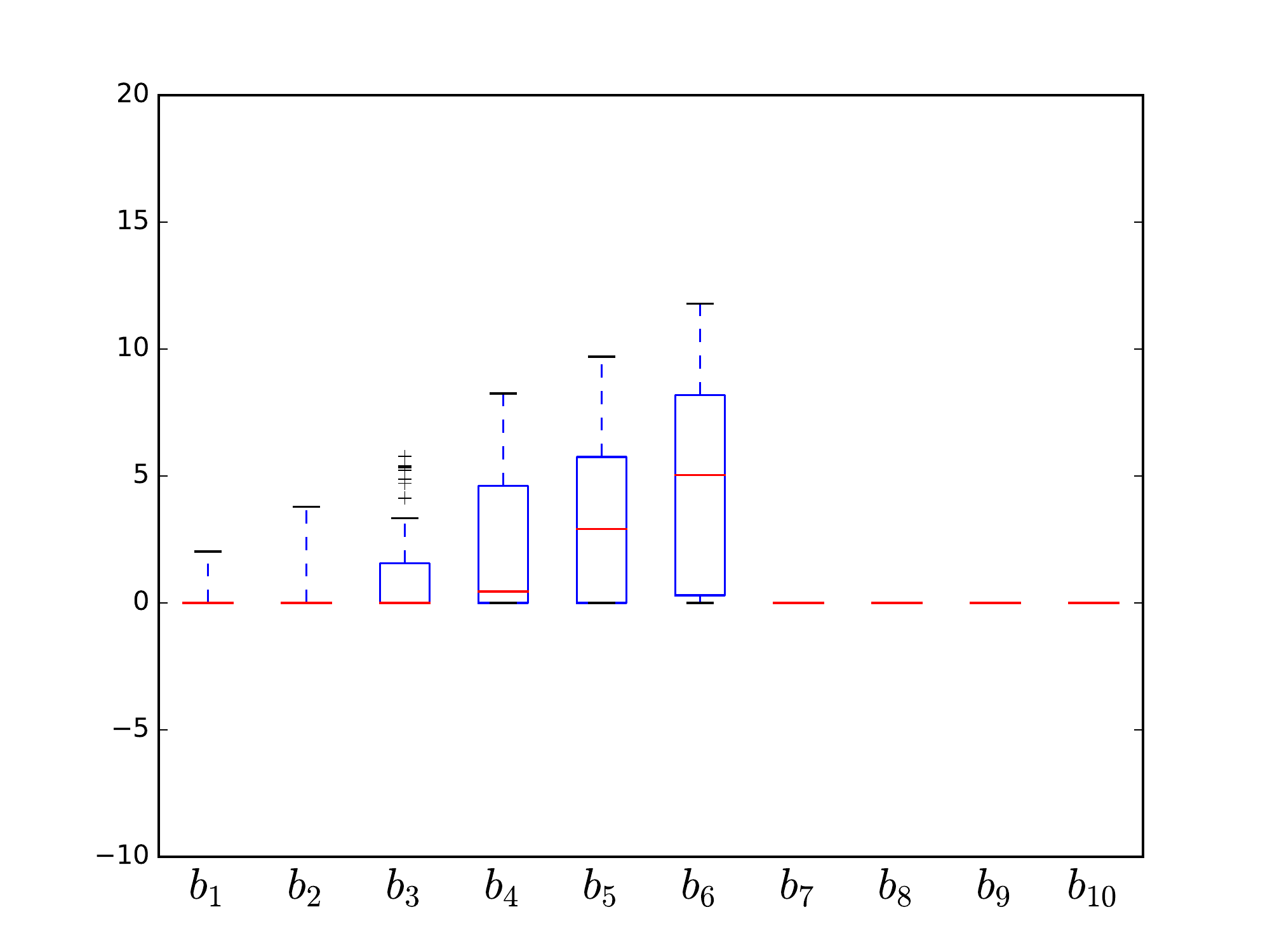}}

	\subfloat[Histogram of $GR^{2}$]
	{\includegraphics[width=0.6\paperwidth]{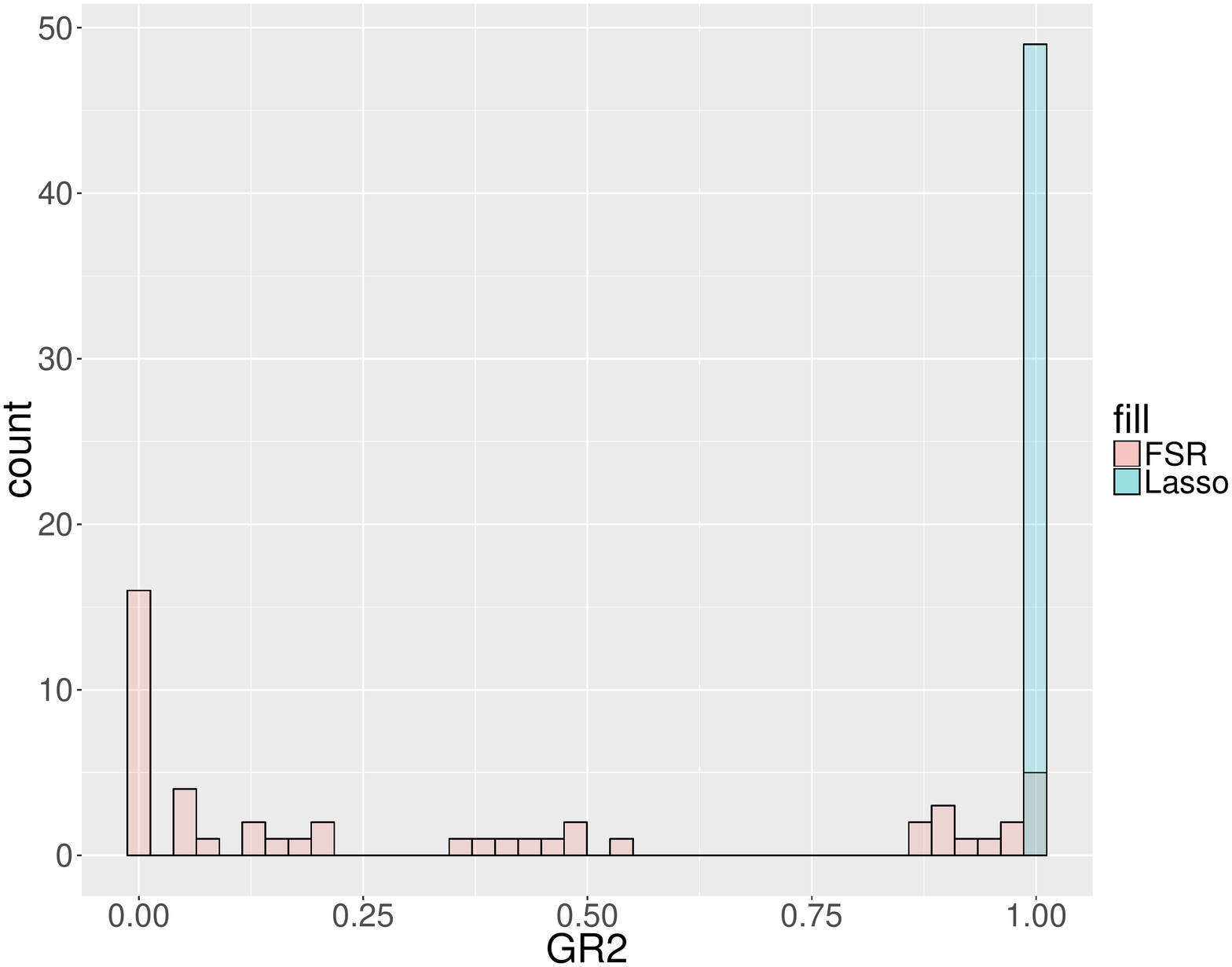}}

	\caption{\label{p=500v=1}DGP with $n = 250$, $p = 500$, and $\sigma^2 = 1$.}
\end{figure}

\begin{figure}[ht]
	\centering
	\subfloat[Lasso estimates]
	{\includegraphics[width=0.33\paperwidth]{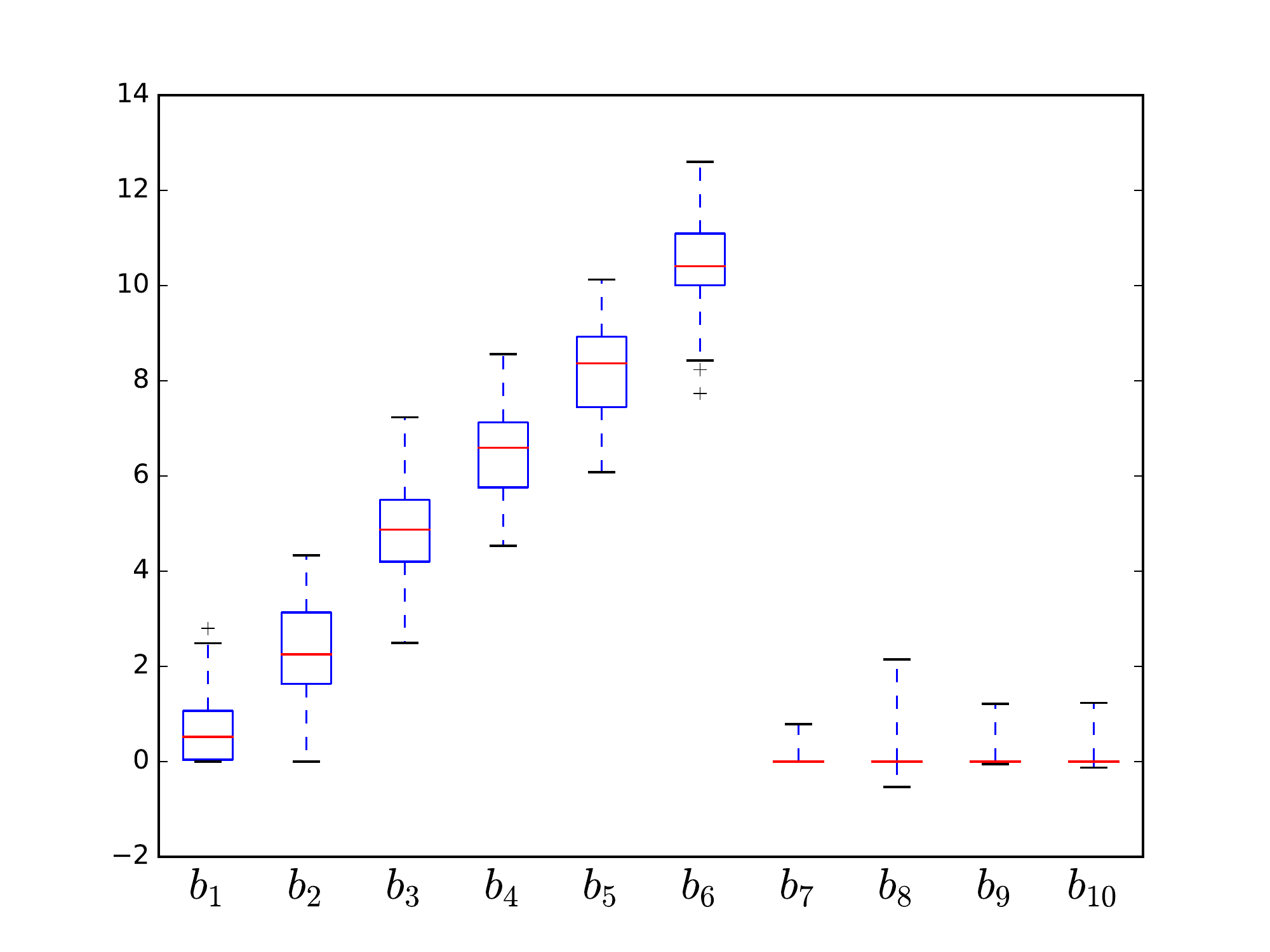}}
	\subfloat[OLS estimates]
	{\includegraphics[width=0.33\paperwidth]{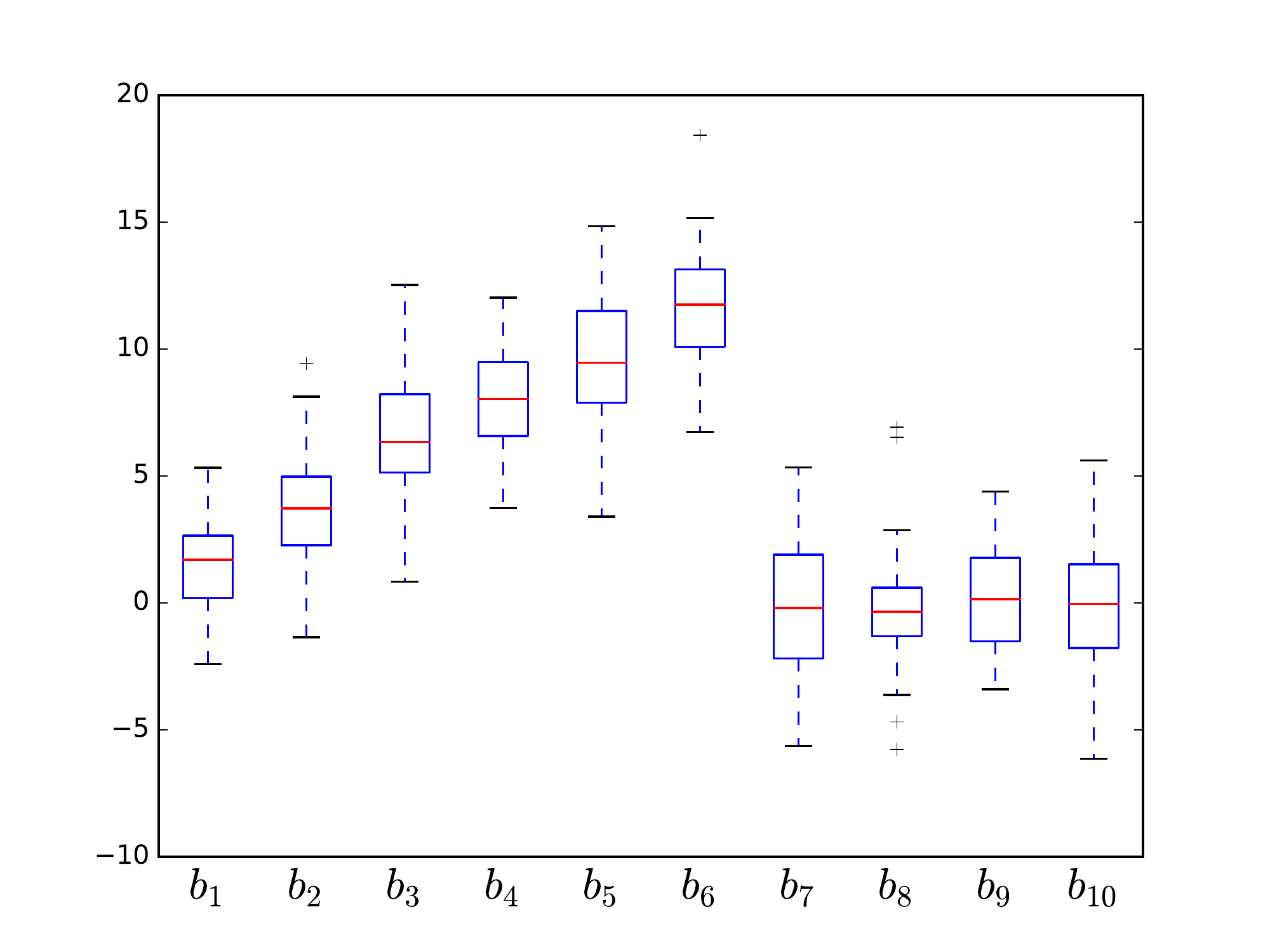}}

	\subfloat[Histogram of $GR^{2}$]
	{\includegraphics[width=0.6\paperwidth]{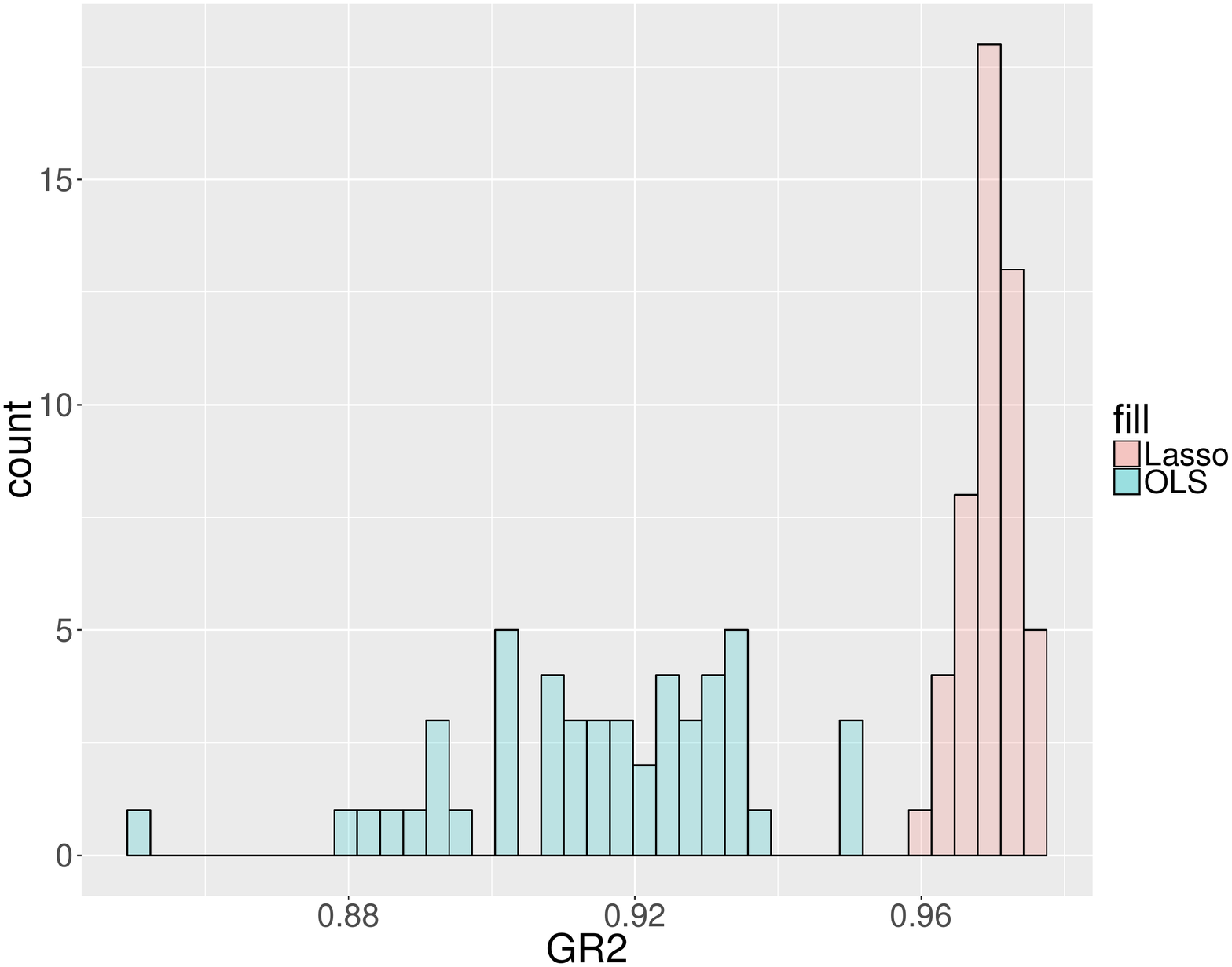}}

	\caption{\label{p=200v=5}DGP with $n = 250$, $p = 200$, and $\sigma^2 = 5$.}
\end{figure}

\begin{figure}[ht]
	\centering
	\subfloat[Lasso estimates]
	{\includegraphics[width=0.33\paperwidth]{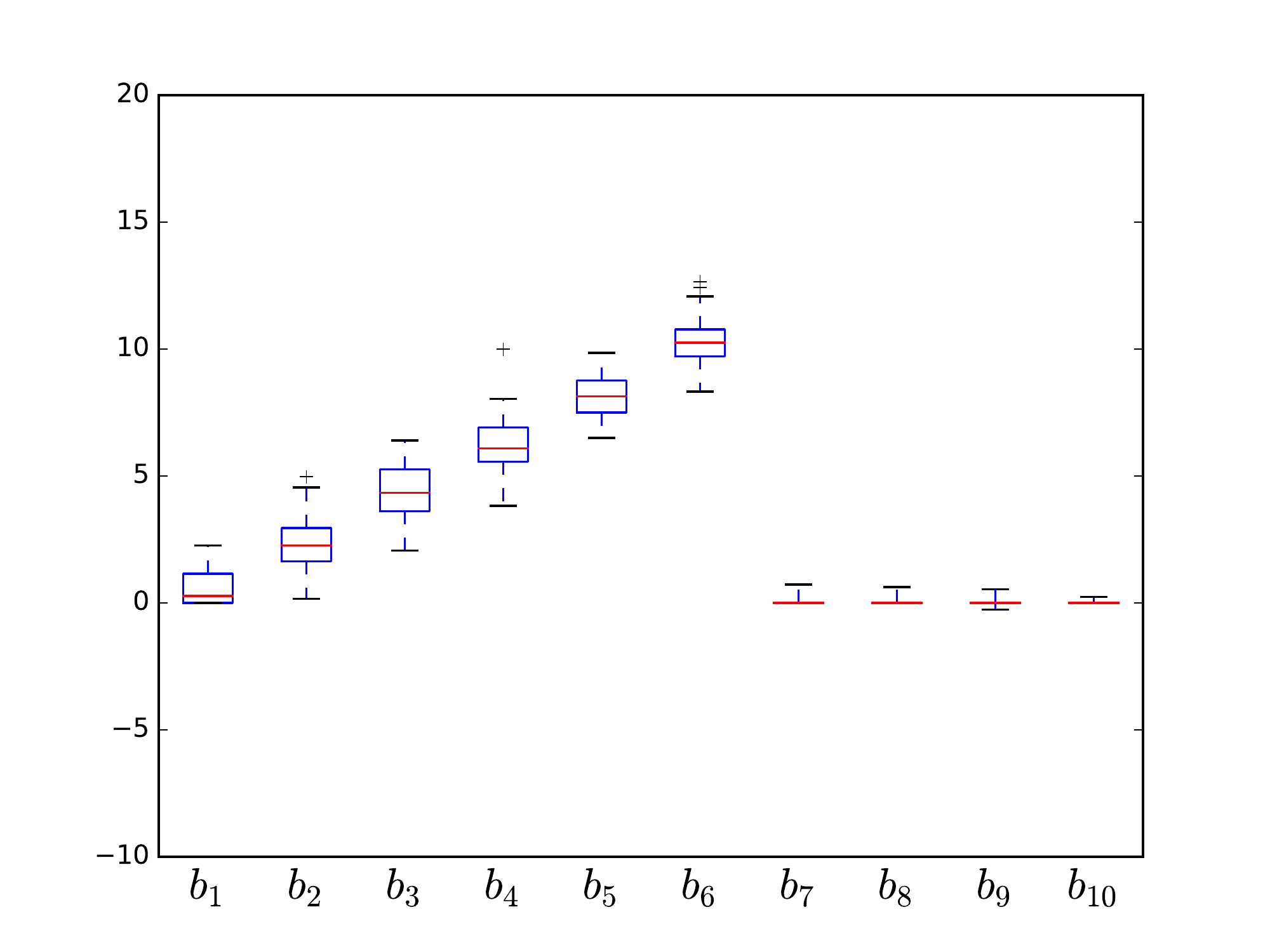}}
	\subfloat[FSR estimates]
	{\includegraphics[width=0.33\paperwidth]{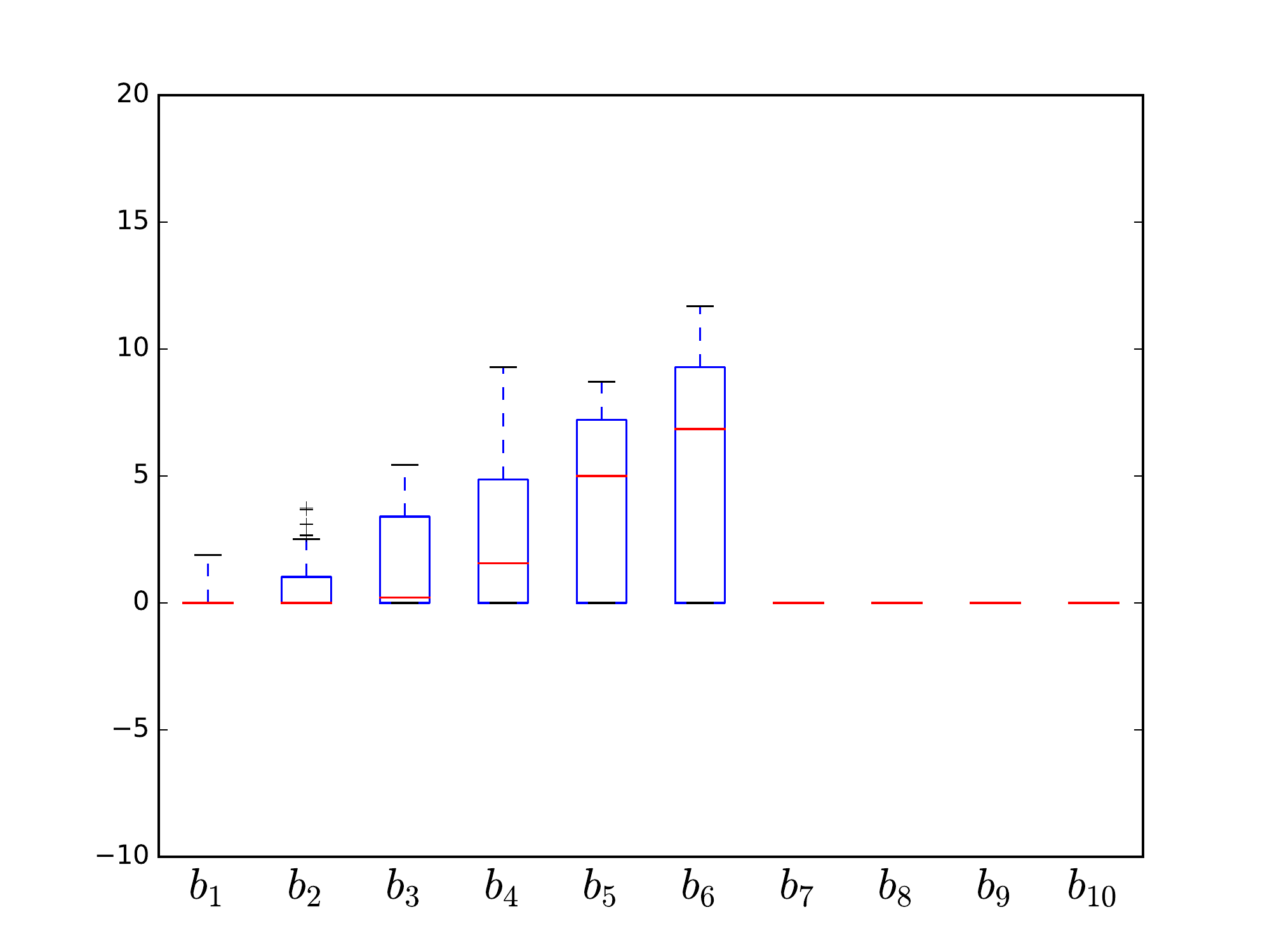}}

	\subfloat[Histogram of $GR^{2}$]
	{\includegraphics[width=0.6\paperwidth]{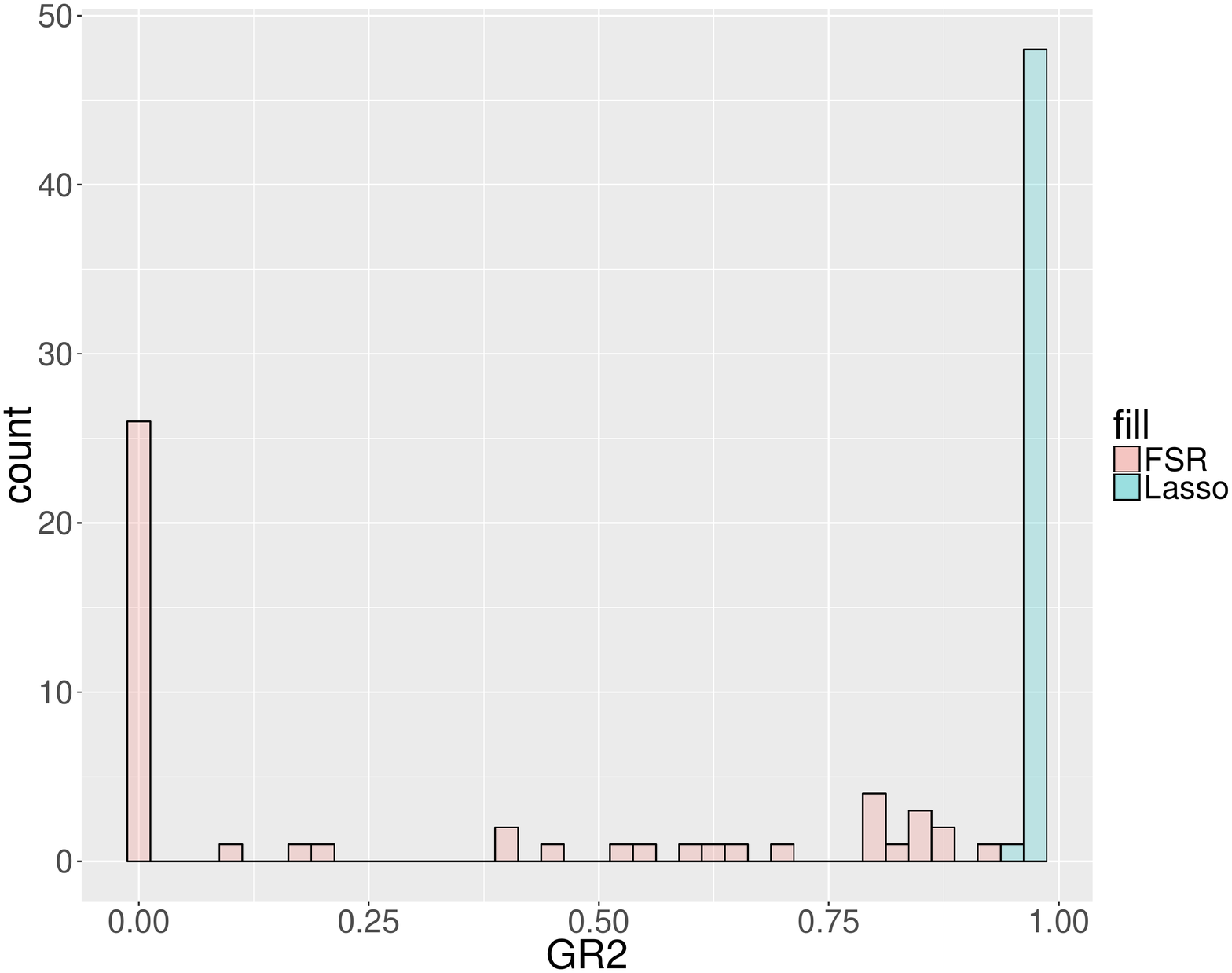}}

	\caption{\label{p=500v=5}DGP with $n = 250$, $p = 500$, and $\sigma^2 = 5$.}
\end{figure}

When $n<p=500$, the regression model is not identified, OLS is not feasible, and we apply FSR. As shown in Figures~\ref{p=500v=1} and \ref{p=500v=5}, lasso still performs well and correctly selects the variables with non-zero coefficients. In contrast, although FSR also correctly identifies the non-zero coefficients, the biases and deviations are much larger than for the lasso. The $GR^2$ in Figures~\ref{p=500v=1} and \ref{p=500v=5} clearly indicate that the FSR estimates are unreliable. Generally speaking, overfitting is controlled well by lasso (the $GR^2$ are close to 1) whereas the performance of FSR is poor. This suggests that, by imposing an $\mathcal{L}_1$-penalty on estimation, lasso mitigates the overfitting problem and, moreover, that the advantage of lasso is likely to be more pronounced as $p$ increases.

\begin{table}[ht]
	\caption{Bias, eTE, eGE, $R^2_t$, $R^2_s$, and $GR^{2}$ for lasso and OLS/FSR with $n=250$}
	\centering

	\begin{tabular}{l.....}
		\toprule
		Measure & \multicolumn{2}{c}{$\sigma^2 = 1$} & \multicolumn{2}{c}{$\sigma^2 = 5$} \\
        \cmidrule(r{0.5em}){2-3} \cmidrule(r{0.5em}){4-5}
        & \multicolumn{1}{c}{$p = 200$}
		& \multicolumn{1}{c}{$p = 500$}
		& \multicolumn{1}{c}{$p = 200$}
		& \multicolumn{1}{c}{$p = 500$} \\
		\midrule
		Bias\\
		\hspace*{4mm}$b_{Lasso}$   & 0.7923 & 0.8810   & 3.8048   & 4.1373    \\
		\hspace*{4mm}$b_{OLS/FSR}$ & 0.9559 & 11.7530  & 4.7797   & 13.7622   \\
		eTE\\
		\hspace*{4mm}Lasso         & 0.9167 & 0.8625   & 22.2476  & 21.1334   \\
		\hspace*{4mm}OLS/FSR       & 0.2164 & 832.9988 & 5.4097   & 1034.2636 \\
		eGE\\
		\hspace*{4mm}Lasso         & 1.1132 & 1.1478   & 27.8672  & 28.5125   \\
		\hspace*{4mm}OLS/FSR       & 5.2109 & 852.5822 & 134.8725 & 1070.6329 \\
		$R^2_t$\\
		\hspace*{4mm}Lasso         & 0.9994 & 0.9994   & 0.9866   & 0.9867    \\
		\hspace*{4mm}OLS/FSR       & 0.9999 & 0.4678   & 0.9967   & 0.3619    \\
		$R^2_s$\\
		\hspace*{4mm}Lasso         & 0.9993 & 0.9993   & 0.9830   & 0.9826    \\
		\hspace*{4mm}OLS/FSR       & 0.9967 & 0.4681   & 0.9181   & 0.3627    \\
		$GR^{2}$\\
		\hspace*{4mm}Lasso         & 0.9988 & 0.9987   & 0.9698   & 0.9695    \\
		\hspace*{4mm}OLS/FSR       & 0.9965 & 0.3659   & 0.9151   & 0.2935    \\
		\bottomrule
	\end{tabular}
    \label{table2}
\end{table}

Table~\ref{table2} reinforces impressions from the boxplots and histograms. When $n>p=200$, OLS performs extremely well in terms of training error although more poorly in terms of generalization error while its $GR^2$ is very close to the lasso value. For $n<p=500$ what is noteworthy is the stable performance of the lasso relative to that of FSR. The training errors, generalization errors, and $GR^2$ are particulary poor for FSR, again illustrating the advantage of the lasso in avoiding overfitting.

\section{Conclusion}

In this paper, we study the performance of penalized and unpenalized extremum estimators from the perspective of generalization ability (GA), the ability of a model to predict outcomes in new samples from the same population. We analyze the GA of penalized regression estimates for the $n\geqslant p$ and the $n<p$ cases. We propose inequalities for the extremum estimators, which bound empirical out-of-sample prediction errors as a function of in-sample errors, sample sizes, model complexity and the heaviness in the tail of the error distribution. The inequalities serve not only to quantify GA, but also to illustrate the trade-off between in-sample and out-of-sample fit, which in turn may be used for tuning estimation hyperparameters, such as the number of folds $K$ in cross-validation or $n_t/n_s$ in validation. We show that some finite-sample and asymptotic properties of the penalized estimators are explained directly by their GA. Furthermore, we use the bounds to quantify the $\mathcal{L}_2$-norm difference between the penalized and corresponding unpenalized regression estimates.

Our work sheds new light on penalized regression and on the applicability of GA for model selection, as well as further insight into the bias-variance trade-off. In this paper, we focus mainly on implementing penalized regression. However, other penalty methods, such as penalized MLE, functional regression, principle component analysis and decision trees, potentially fit the GA framework. Furthermore, the results we establish for penalized regression and GA may be implemented with other empirical methods, like the EM algorithm, clustering, mixture and factor modeling, Bayes networks, and so on.

In providing a general property for all penalized regressions, the generalization error bounds are necessarily conservative. Finer error bounds may well be derived by focusing on specific penalized regression methods. Lastly, as an early attempt to incorporate GA analysis into econometrics, we focus on the i.i.d.\ case. However, it is clear that the framework has the potential to be generalized to non-i.i.d.\ data in settings like $\alpha$- and $\beta$-mixing stationery time series data as well as dependent and non-identical panel data.

\newpage{}
\section*{References}

\bibliographystyle{elsarticle-harv}
\bibliography{LASSOrefs2}

\newpage{}
\renewcommand{\theequation}{A.\arabic{equation}}
\setcounter{equation}{0}
%%%%%%%%%%%%%%%%%%%%%%%%%%%%%%%%%%%%%%%%%%%%%%%%%%%%%%%%%%%%%%%%%%%%%%%%%%%%%%%%%%%%%%%%%%%%%%%%%%%%%%%%%%%
%%%%%%%%%%%%%
%%%%%%%%%%%%%            APPENDIX A
%%%%%%%%%%%%%
%%%%%%%%%%%%%%%%%%%%%%%%%%%%%%%%%%%%%%%%%%%%%%%%%%%%%%%%%%%%%%%%%%%%%%%%%%%%%%%%%%%%%%%%%%%%%%%%%%%%%%%%%%

\section*{Appendix A: Proofs}

%%%%%% Theorem 2.1 proof %%%%%%%%%%%%%%%%%%%%%%%%%%%%%%%%%%%%%%%%%%%%%%%%%%%%%%%%%%%%%%%%%%%%%%%%%%%%%%%%%
\begin{proof}
\textbf{Theorem~\ref{thm2.1}}
Since $b_{train}=\mathrm{argmin}_{b}\;\mathcal{R}_{n_t}\left(b\vert Y_t,X_t\right)$, eq.~{(\ref{eq:lem2.1})} forms an upper bound for the generalization error with probability $1-1/n_t$, $\forall b$,
\[
    \mathcal{R}(b_{train}|Y,X) \leqslant \mathcal{R}_{n_t}(b_{train}|Y_{t},X_{t})\left(1-\sqrt{\epsilon}\right)^{-1}
\]
where $\mathcal{R}_{n_t}(b_{train}|Y_{t},X_{t})$ is the training error on $(Y_{t},X_{t})$, $\mathcal{R}(b_{train}|Y,X)$ is the true population error of $b_{train}$ and $\epsilon=(1/n_t)\left\{h\ln\left[\left(n_t/h\right)\right]+h-\ln\left(1/n_t\right)\right\} $.

To use eq.~(\ref{eq:lem2.1}) to quantify the relation between eGE and eTE, we need to consider whether the loss function $Q(b_{train}\vert y,\mathbf{x})$ has no tail, a light tail, or a heavy tail.

\begin{description}

\item[\textbf{No tail.}] If the loss function $Q(\cdot)$ is bounded between $[0,B]$, where $B\in(0,\infty)$, then from Hoeffding's inequality \citep{hoeffding63} for the extremum estimator $b_{train}$, the empirical process satisfies, $\forall\varsigma\geqslant0$,
    \begin{equation}
     	\mathrm{P}\left\{\vert\mathcal{R}_{n_s}(b_{train}|X_{s},Y_{s})-\mathcal{R}(b_{train}|X,Y)\vert
	  	\leqslant\varsigma\right\}
		\geqslant
        1 - 2\exp\left(-\frac{2n_s\varsigma}{B}\right)
    \end{equation}	
	If we define $\varpi=1-2\exp(-2n^2\varsigma/\sum_{i=1}^{n}B_i)$, then	
	\begin{equation}
		\varsigma=\frac{B}{n_s}\ln\sqrt{\frac{2}{1-\varpi}}
	\end{equation}
	This implies, for any extremum estimator $b_{train}$	
	\begin{equation}
        \mathrm{P}\{\mathcal{R}_{n_s}(b_{train}|X_{s},Y_{s})\leqslant\mathcal{R}(b_{train}|X,Y)+\varsigma\} \geqslant
        \varpi.
	\end{equation}	
    Since both the training and test set are randomly sampled from the population, eq.~(\ref{eq:lem2.1}) can be modified as follows: $\forall\varsigma\geqslant0,\,\forall\tau_{1}\geqslant0$, there exists an $N_{1}\in\mathbb{R}^{+}$ subject to	
	\begin{equation}
		\mathrm{P}\left\{\mathcal{R}_{n_s}\left(b_{train}|X_{s},Y_{s}\right) \leqslant
        \frac{\mathcal{R}_{n_t}(b_{train}|X_{t},Y_{t})}{1-\sqrt{\epsilon}}+
		\varsigma\right\}
		\geqslant
        \varpi\left(1-\frac{1}{n_t}\right)
	\end{equation}

\item[\textbf{Light tail.}] Suppose the loss function $Q(\cdot)$ is unbounded, but still $\mathcal{F}$-measurable, and possesses a finite $\nu$th moment when $\nu>2$. Based on Chebyshev's inequality for the extremum estimator $b_{train}$, the empirical process satisfies, $\forall\varsigma\geqslant0$,
    \begin{eqnarray}
     	\mathrm{P}\left\{\vert\mathcal{R}_{n_s}(b_{train}|X_{s},Y_{s})-\mathcal{R}(b_{train}|X,Y)\vert
	  	\leqslant\varsigma\right\}
		& \geqslant &
        1 - \frac{\mathrm{var}(Q(b_{train}\vert y,\mathbf{x}))}{n_s\varsigma}
    \end{eqnarray}	
	If we define $\varpi = 1 - \mathrm{var}(Q(b_{train}\vert y,\mathbf{x}))/n_s\varsigma$, then	
	\begin{equation}
		\varsigma=\frac{\mathrm{var}(Q(b_{train}\vert y,\mathbf{x}))}
		{n_s(1-\varpi)}
	\end{equation}
	This implies, for any extremum estimator $b_{train}$	
	\begin{equation}
        \mathrm{P}\{\mathcal{R}_{n_s}(b_{train}|X_{s},Y_{s})\leqslant\mathcal{R}(b_{train}|X,Y)+\varsigma\} 		\geqslant
        \varpi.
	\end{equation}	
    Since both the training and test set are randomly sampled from the population, eq.~(\ref{eq:lem2.1}) can be modified as follows: $\forall\varsigma\geqslant0,\,\forall\tau_{1}\geqslant$, there exists an $N_{1}\in\mathbb{R}^{+}$ subject to	
	\begin{equation}
		\mathrm{P}\left\{\mathcal{R}_{n_s}\left(b_{train}|X_{s},Y_{s}\right) \leqslant\frac{\mathcal{R}
		_{n_t}(b_{train}|X_{t},Y_{t})}{1-\sqrt{\epsilon}}+
		\varsigma\right\}
		\geqslant \varpi\left(1-\frac{1}{n_t}\right)
	\end{equation}

\item[\textbf{Heavy tail.}] Suppose the loss function $Q(\cdot)$ is unbounded, but still $\mathcal{F}$-measurable, and has heavy tails with the property that, for $1<\nu\leqslant 2$, $\exists\tau$, such that
	\begin{equation}
		\sup
		\frac{\sqrt[\nu]{\int[Q(b_{train}\vert y,\mathbf{x})]^\nu dF\left(y,\mathbf{x}\right)}}
		{\int Q(b_{train}\vert y,\mathbf{x})dF\left(y,\mathbf{x}\right)}
		\leqslant
        \tau.
	\end{equation}	
    Based on the Bahr-Esseen inequality for the extremum estimator $b_{train}$, the empirical process satisfies, $\forall\varsigma\geqslant0$,	
	\begin{eqnarray}
        \mathrm{P}\{\vert\mathcal{R}(b_{train}|Y,X)-\mathcal{R}_{n_s}(b_{train}|Y_{s},X_{s})\vert
        \leqslant\varsigma\}
		& \geqslant & 1-2\frac{\mathbb{E}\left[Q(b_{train}\vert y,\mathbf{x})^\nu\right]}
		{\varsigma^\nu n_s^{\nu-1}} \notag \\
		& \geqslant & 1-2\tau^\nu\frac{\left(\mathbb{E}\left[Q(b_{train}\vert y,\mathbf{x})\right]
		\right)^\nu}{\varsigma^\nu n_s^{\nu-1}}
	\end{eqnarray}	
	If we define
	\begin{eqnarray}
        \varpi=1-2\tau^\nu\frac{\left(\mathbb{E}\left[Q(b_{train}\vert y,\mathbf{x})\right]\right)^\nu}
        {\varsigma^\nu n_s^{\nu-1}}
	\end{eqnarray}	
  	then  	
  	\begin{equation}
        \varsigma=\frac{\sqrt[\nu]{2}\tau\left(\mathbb{E}
        \left[Q(b_{train}\vert y,\mathbf{x})\right]\right)}
        {\sqrt[\nu]{1-\varpi} n_s^{1-1/{\nu}}}
	\end{equation}
	This implies, for any extremum estimator $b_{train}$	
	\begin{equation}
        \mathrm{P}\{\mathcal{R}_{n_s}(b_{train}|X_{s},Y_{s})
        \leqslant\mathcal{R}(b_{train}|X,Y)+\varsigma\}
        \geqslant
        \varpi.
	\end{equation}	
    Since both the training and test set are randomly sampled from the population, eq.~(\ref{eq:lem2.1}) could be modifed and relaxed as follows: $\forall\varsigma\geqslant0,\,\forall\tau_{1}\geqslant$, there exists an $N_{1}\in\mathbb{R}^{+}$ subject to	
	\begin{equation}
		\mathrm{P}\left\{\mathcal{R}_{n_s}\left(b_{train}|X_{s},Y_{s}\right) \leqslant\frac{\mathcal{R}
		_{n_t}(b_{train}|X_{t},Y_{t})}{1-\sqrt{\epsilon}}+
		\varsigma\right\}
		\geqslant \varpi\left(1-\frac{1}{n_t}\right)
	\end{equation}	
\end{description}
\end{proof}
%%%%%%%%%%%%%%%%%%%%%%%%%%%%%%%%%%%%%%%%%%%%%%%%%%%%%%%%%%%%%%%%%%%%%%%%%%%%%%%%%%%%%%%%%%%%%%%%%%%%%%%%%%

%%%%%% Theorem 2.2 proof %%%%%%%%%%%%%%%%%%%%%%%%%%%%%%%%%%%%%%%%%%%%%%%%%%%%%%%%%%%%%%%%%%%%%%%%%%%%%%%%%
\begin{proof}
\textbf{Theorem~\ref{thm2.2}}
The upper bound of the eGE under cross-validation can be established by adapting eq.~(\ref{eq:lem2.1}) and (\ref{eq:thm2.1}). The proof is quite similar to Theorem~\ref{thm2.1}, except for the fact that eq.~(\ref{eq:lem2.1}) and (\ref{eq:thm2.1}) measure the upper bound by one-round eTE while eq.~(\ref{eq1:thm2.2}) and (\ref{eq2:thm2.2}) uses averaged multiple-round eTE, thus
\[
\frac{1}{K}\sum_{q=1}^{K}\mathcal{R}_{n_t}(b_{train}|Y_{t}^q,X_{t}^q).
\]
As shown in Theorem~\ref{thm2.1},
\begin{eqnarray}
        & & \mathbf{P} \{ \mathcal{R}_{n_s}(b_{train}|Y_{s},X_{s})
        \leqslant
        \frac{\mathcal{R}_{n_t}(b_{train}|Y_{t},X_{t})}{1-\sqrt{\epsilon}} + \varsigma \}
        \geqslant \varpi(1-1/n_t) \notag \\
        & \iff & \mathbf{P} \{ \mathcal{R}_{n_s}(b_{train}|Y_{s},X_{s})
        - \frac{\mathcal{R}_{n_t}(b_{train}|Y_{t},X_{t})}{1-\sqrt{\epsilon}}
        \leqslant \varsigma \}
        \geqslant \varpi(1-1/n_t)
\end{eqnarray}
As a result, in each round of cross validation, $\forall q\in[1,K]$,
\begin{eqnarray}
        & & \mathbf{P} \{ \mathcal{R}_{n_s}(b_{train}|Y_{s}^q,X_{s}^q)
        - \frac{\mathcal{R}_{n_t}(b_{train}|Y_{t}^q,X_{t}^q)}{1-\sqrt{\epsilon}}
        \leqslant \varsigma \}
        \geqslant \varpi(1-1/n_t)
\end{eqnarray}
Here we define $T_q = \mathcal{R}_{n_s}(b_{train}|Y_{s}^q,X_{s}^q) - \frac{\mathcal{R}_{n_t}(b_{train}|Y_{t}^q,X_{t}^q)}{1-\sqrt{\epsilon}}$, which implies that the mean of $T_q - \mathbb{E}[T_q]$ is zero and
\begin{eqnarray}
        & & \mathbf{P} \{ \mathcal{R}_{n_s}(b_{train}|Y_{s}^q,X_{s}^q)
        - \frac{\mathcal{R}_{n_t}(b_{train}|Y_{t}^q,X_{t}^q)}{1-\sqrt{\epsilon}}
        \leqslant \varsigma \}
        \geqslant \varpi(1-1/n_t) \notag \\
        & \iff & \mathbf{P} \{ T_q -\mathbb{E}[T_q]   \leqslant \varsigma -\mathbb{E}[T_q]\}
        \geqslant \varpi(1-1/n_t)
\end{eqnarray}

\begin{description}

        \item[no-tail or light-tail] If the $\mathcal{R}_{n_t}(b_{train}|Y_{t}^q,X_{t}^q) \in \left(0,B\right]$ or $\mathbb{E}(\vert T_q \vert^m) \leqslant m!B^{m-2}\mathrm{var}(T_q)/2, \forall m \geqslant 2$, by Bernstein inequality,
        \begin{eqnarray}
               & & \mathbf{P} \left\{ \frac{1}{K}\sum^{K}_{q=1} \mathcal{R}_{n_s}(b_{train}|Y_{s}^q,X_{s}^q)
               \leqslant \frac{1}{K}\sum^{K}_{q=1}
               \frac{\mathcal{R}_{n_t}(b_{train}|Y_{t}^q,X_{t}^q)}{1-\sqrt{\epsilon}}
                + \varsigma \right\} \notag \\
               & \geqslant &\mathbf{P} \left\{ \sum^{K}_{q=1} T_q -\mathbb{E}[T_q] \leqslant \vert \sum^{K}_{q=1} T_q -\mathbb{E}[T_q] \vert
               \leqslant K\varsigma - K\mathbb{E}[T_q] \right\} \notag \\
               & \geqslant &
               1 - 2\exp\left\{-\frac{1}{2}
               \frac{(K\varsigma - K\mathbb{E}[T_q])^2}
               {\mathrm{var}(\sum^{K}_{q=1}T_q) + B(K\varsigma - K\mathbb{E}[T_q])/3}\right\} \notag \\
               & = &
               1 - 2\exp\left\{-\frac{1}{2}
               \frac{(\varsigma - \mathbb{E}[T_q])^2}
               {\mathrm{var}(T_q)/K + B(\varsigma - \mathbb{E}[T_q])/(3K)}\right\}
        \end{eqnarray}

        \item[heavy-tail] If the $\mathcal{R}_{n_t}(b_{train}|Y_{t}^q,X_{t}^q)$ is $\mathcal{F}$-measurable but heavy-tailed, the Bernstein inequality fails for convolution and we cannot approximate the convoluted probability with Gaussian function. Hence, we need to narrow the category of `heavy-tailed distribution' to the major subclass of heavy-tailed distributions, the sub-exponential distributions.

        If $T_q$ is sub-exponential variable, all the sub-exponential distributions, by definition, satisfy the following condition
        \[
               \mathbf{P}\left\{ \frac{1}{K} \sum^{K}_{q = 1} T_q > \varsigma \right\}
               \sim K\mathbf{P}\left\{ T_1 > K\varsigma \right\}.
        \]
        As a result,
        \begin{eqnarray}
               & & \mathbf{P} \left\{ \frac{1}{K}\sum^{K}_{q=1} \mathcal{R}_{n_s}(b_{train}|Y_{s}^q,X_{s}^q)
               \leqslant \frac{1}{K}\sum^{K}_{q=1}
               \frac{\mathcal{R}_{n_t}(b_{train}|Y_{t}^q,X_{t}^q)}{1-\sqrt{\epsilon}}
               + \varsigma \right\} \notag \\
               & \geqslant & \mathbf{P} \left\{ \frac{1}{K} \sum^{K}_{q=1} T_q
               \leqslant \frac{1}{K} \sum^{K}_{q=1} \vert T_q \vert
               \leqslant \varsigma \right\} \notag \\
               & \geqslant & 1 - \mathbf{P} \left\{ \frac{1}{K} \sum^{K}_{q=1}
               \vert T_q \vert \geqslant \varsigma \right\} \notag \\
               & \sim & 1- K\mathbf{P} \left\{ \vert T_1 \vert \geqslant K\varsigma \right\} \notag \\
               & = & 1 - K(1-(1-2\tau^\nu\cdot
               \frac{\left(\mathbb{E}\left[Q(b_{train}\vert y,\mathbf{x})\right]\right)^\nu}
        {(K\varsigma)^\nu\cdot n_s^{\nu-1}})(1-1/n_t)) \notag \\
        & \geqslant & 1 - 2\tau^\nu\cdot
               \frac{\left(\mathbb{E}\left[Q(b_{train}\vert y,\mathbf{x})\right]\right)^\nu}
        {\varsigma^\nu\cdot n^{\nu}}-K/n_t
        \end{eqnarray}
        Hence, when $ \tau\cdot\mathbb{E}\left[Q(b_{train}\vert y,\mathbf{x})\right]/ \varsigma \leqslant 1 $ and large $n$, we can approximately have the following probabilistic bound
        \begin{eqnarray}
               & &\mathbf{P} \left\{ \frac{1}{K}\sum^{K}_{q=1} \mathcal{R}_{n_s}(b_{train}|Y_{s}^q,X_{s}^q)
               \leqslant \frac{1}{K}\sum^{K}_{q=1}
               \frac{\mathcal{R}_{n_t}(b_{train}|Y_{t}^q,X_{t}^q)}{1-\sqrt{\epsilon}}
               + \varsigma \right\} \notag \\
               & \geqslant & \left(1 - 2\tau^\nu\cdot
               \frac{\left(\mathbb{E}\left[Q(b_{train}\vert y,\mathbf{x})\right]\right)^\nu}
        {\varsigma^\nu\cdot n^{\nu}}-K/n_t \right)^+,
        \end{eqnarray}
        or in words, if the loss function is heavy-tailed and sub-exponential, with relatively large $\varsigma$ and relatively large $n$, the upper bound for cross validation can be approximately established.
\end{description}

\end{proof}
%%%%%%%%%%%%%%%%%%%%%%%%%%%%%%%%%%%%%%%%%%%%%%%%%%%%%%%%%%%%%%%%%%%%%%%%%%%%%%%%%%%%%%%%%%%%%%%%%%%%%%%%%%

%%%%%% Proposition 3.1 proof %%%%%%%%%%%%%%%%%%%%%%%%%%%%%%%%%%%%%%%%%%%%%%%%%%%%%%%%%%%%%%%%%%%%%%%%%%%%%
\begin{proof}
\textbf{Proposition~\ref{prop3.1}}
Given A1--A6, the true DGP is
\[
	y_{i} = x_{i}^{T}\beta + u_{i}, \quad i=1,\ldots,n.
\]
Proving that the true DGP has the lowest eGE is equivalent to proving, for any test set, that
\begin{equation}
    \frac{\sum_{i=1}^n\left(y_{i}-x_{i}^{T}\beta\right)^{2}}{n}
    \leqslant
    \frac{\sum_{i=1}^n\left(y_{i}-x_{i}^{T}b\right)^{2}}{n},
    \label{a1p1}
\end{equation}
which is equivalent to proving that
\begin{align*}
	0 & \leqslant \frac{1}{n} \sum_{i=1}^n\left[\left(y_{i}-x_{i}^{T}b\right)^{2}-
           \left(y_{i}-x_{i}^{T}\beta\right)^{2}\right] \\
	\iff
	0 & \leqslant \frac{1}{n}
	  \sum_{i=1}^n\left(y_{i}-x_{i}^{T}b+y_{i}-x_{i}^{T}\beta\right)
	\left(y_{i}-x_{i}^{T}b-y_{i}+x_{i}^{T}\beta\right) \\
	\iff
	0 & \leqslant \frac{1}{n}
	  \sum_{i=1}^n\left(y_{i}-x_{i}^{T}b+y_{i}-x_{i}^{T}\beta\right)
	  \left(x_{i}^{T}\beta-x_{i}^{T}b\right).
\end{align*}
Defining $\delta=\beta-b$, it follows,
\begin{align*}
	0      & \leqslant \frac{1}{n}
	\sum_{i=1}^n\left(2y_{i}-x_{i}^{T}b-x_{i}^{T}\beta\right)
	\left(x_{i}^{T}\delta\right) \\
	\iff 0 & \leqslant \frac{1}{n}
    \sum_{i=1}^n\left(2y_{i}-x_{i}^{T}\beta+x_{i}^{T}\beta-x_{i}^{T}b-x_{i}^{T}\beta\right)
	\left(x_{i}^{T}\delta\right) \\
	\iff 0 & \leqslant \frac{1}{n}
	\sum_{i=1}^n\left(2y_{i}-2x_{i}^{T}\beta+x_{i}^{T}\delta\right)
	\left(x_{i}^{T}\delta\right) \\
	\iff 0 & \leqslant \frac{1}{n}
	\sum_{i=1}^n\left(2u_{i}+x_{i}^{T}\delta\right)\left(x_{i}^{T}\delta\right)
\end{align*}
Hence, proving Proposition~\ref{prop3.1} is equivalent to proving that
\[
	\frac{1}{n} \sum_{i=1}^n\left(2u_{i}+x_{i}^{T}\delta\right)\left(x_{i}^{T}\delta\right) \geqslant 0
\]
Since $\mathbb{E}(X^T u)=\mathbf{0}$ from A2, it follows that
\[
 	\frac{1}{n}\sum_{i=1}^n u_{i} x_{i}\overset{\mathbf{P}}{\rightarrow}\mathbf{0}
	\iff
	\frac{1}{n}\sum_{i=1}^n\left(u_{i} x_{i}^{T}\right)\beta\overset{\mathbf{P}}{\rightarrow}0
	\quad\mbox{and}\quad
	\frac{1}{n}\stackrel[i=1]{n}{\sum}\left(u_{i} x_{i}^{T}\right)b\rightarrow 0
\]
Hence, asymptotically
\[
	\frac{1}{n}\sum_{i=1}^n\left(2u_{i}+x_{i}^{T}\delta\right)\left(x_{i}^{T}\delta\right)=
	\frac{1}{n}\sum_{i=1}^n 2\delta u_{i}x_{i}^{T}+
	\frac{1}{n}\sum_{i=1}^n\left(x_{i}^{T}\delta\right)^{2}
	\overset{\mathbf{P}}{\rightarrow}\mathbb{E}\left(x_{i}^{T}\delta\right)^{2}\geqslant 0
\]
\end{proof}
%%%%%%%%%%%%%%%%%%%%%%%%%%%%%%%%%%%%%%%%%%%%%%%%%%%%%%%%%%%%%%%%%%%%%%%%%%%%%%%%%%%%%%%%%%%%%%%%%%%%%%%%%%

\bigskip
%%%%%% Proposition 3.2 proof %%%%%%%%%%%%%%%%%%%%%%%%%%%%%%%%%%%%%%%%%%%%%%%%%%%%%%%%%%%%%%%%%%%%%%%%%%%%%
\begin{proof}
\textbf{Proposition~\ref{prop3.2}}
The proof of Proposition~\ref{prop3.2} is very straightforward. When $\lambda = 0$, $b_{\lambda} = b_{OLS}$. Hence, as $n\rightarrow \infty$, $b_{\lambda=0} = b_{OLS} \overset{\mathcal{L}_2}{\rightarrow} \beta$. Hence, there exists at least one $\lambda$ that guarantees $\mathcal{L}_2$-consistency. This guarantees that when $n\rightarrow\infty$, $\beta \in \{b_{\lambda}\}$, or the true DGP is in the list of alternative $b_{\lambda}$.
\end{proof}

%%%%%%%%%%%%%%%%%%%%%%%%%%%%%%%%%%%%%%%%%%%%%%%%%%%%%%%%%%%%%%%%%%%%%%%%%%%%%%%%%%%%%%%%%%%%%%%%%%%%%%%%%%

\bigskip
%%%%%%% Lemma 3.1 proof %%%%%%%%%%%%%%%%%%%%%%%%%%%%%%%%%%%%%%%%%%%%%%%%%%%%%%%%%%%%%%%%%%%%%%%%%%%%%%
\begin{proof}
\textbf{Lemma~\ref{lem3.1}}
Eq.~(\ref{eq1:lem3.1}) and (\ref{eq2:lem3.1}) are the direct application of eqs.~(\ref{eq:thm2.1}) and (\ref{eq1:thm2.2}) and (\ref{eq2:thm2.2}). Thus, we only need to focus on the last term of the RHS, $\varsigma$. Since the error term $u$ in classical regression analysis is distrbuted $N(0,\sigma^2)$, the loss function of OLS
\[
        Q(b_{OLS}) \sim \sigma^2 \chi^2(1).
\]
Hence in eq.~(\ref{eq:thm2.1}) the last RHS term is
\[
        \varsigma = \frac{2\sigma^4}{n_s\sqrt{1-\varpi}}
\]
By substituting the above values for $\varsigma$ into eqs.~(\ref{eq:thm2.1}) and (\ref{eq1:thm2.2}) and (\ref{eq2:thm2.2}), we have eq.~(\ref{eq1:lem3.1}) and (\ref{eq2:lem3.1}).

\end{proof}
%%%%%%%%%%%%%%%%%%%%%%%%%%%%%%%%%%%%%%%%%%%%%%%%%%%%%%%%%%%%%%%%%%%%%%%%%%%%%%%%%%%%%%%%%%%%%%%%%%%%%%

\bigskip
%%%%%% Corollary 3.1 proof %%%%%%%%%%%%%%%%%%%%%%%%%%%%%%%%%%%%%%%%%%%%%%%%%%%%%%%%%%%%%%%%%%%%%%%%%%%%%%%
\begin{proof}
\textbf{Corollary~\ref{cor3.1}}
The optimal $K$ or $n_t/n_s$ can be obtained by finding the smallest expectation of the RHS for eq.~(\ref{eq1:lem3.1}) and (\ref{eq2:lem3.1}). Since the error term $u$ in classical regression analysis is distrbuted $N(0,\sigma^2)$, the loss function of OLS
\[
	Q(b_{OLS}) \sim \sigma^2 \chi^2(1).
\]
As a result,
\[
	\mathcal{R}_{n_s}(b_{OLS}\vert Y,X) \sim \frac{\sigma^2}{n/K}\;\chi^2(n/K).
\]
On the RHS of eq.~(\ref{eq1:thm2.2}) and (\ref{eq2:thm2.2}),
\[
	\frac{1}{K}\sum_{q=1}^{K}\mathcal{R}_{n_t}(b_{train}|Y_{t}^q,X_{t}^q) 	
	\sim
	\frac{\sigma^2}{n(K-1)/K}\;\mathrm{Gamma}\left(\frac{n^2(K-1)^2}{2K^2},\frac{2K}{(K-1)n}\right)
\]
Hence, the expectation of the RHS for eq.~(\ref{eq1:thm2.2}) and (\ref{eq2:thm2.2}) is
\[
	\frac{\sigma^2}{1-\sqrt{\epsilon}} + \frac{2\sigma^4}{\sqrt{1-\varpi}(n/K)}
\]
and
\[
	K^{*} = \argmin_{K}\frac{\sigma^2}{1-\sqrt{\epsilon}} + \frac{2\sigma^4}{\sqrt{1-\varpi}(n/K)}
\]
\end{proof}
%%%%%%%%%%%%%%%%%%%%%%%%%%%%%%%%%%%%%%%%%%%%%%%%%%%%%%%%%%%%%%%%%%%%%%%%%%%%%%%%%%%%%%%%%%%%%%%%%%%%%%%%%%

\bigskip
%%%%%% Proposition 3.1 proof %%%%%%%%%%%%%%%%%%%%%%%%%%%%%%%%%%%%%%%%%%%%%%%%%%%%%%%%%%%%%%%%%%%%%%%%%%%%%
\begin{proof}
\textbf{Proposition~\ref{prop3.3}}
In the proof, $b_{OLS}$ is the OLS estimate learned from the training set $(Y_t,X_t)$ in validation and $b_{OLS}^q$ is the OLS estimate learned from the $q$th training set $(Y_t^q,X_t^q)$ in cross-validation.

\begin{description}
	
    \item[\textbf{Validation.}] As shown in Lemma~\ref{lem3.1}, eq.~(\ref{eq1:lem3.1}) holds with at least probability $\varpi(1-1/n_t)$,
		\begin{equation}
			\mathcal{R}_{n_s}(b_{OLS}|Y_{s},X_{s}) \leqslant
			\frac{\mathcal{R}_{n_t}(b_{OLS}|Y_{t},X_{t})}{1-\sqrt{\epsilon}}
			+ \varsigma	
		\end{equation}
        Also, the validation algorithm guarantees that, among all the $b\in\{ b_{\lambda}\}$, $b_{Lasso}$ has the lowest eGE on the test set,
		\begin{equation}
			\mathcal{R}_{n_s}(b^*|Y_{s},X_{s})
			\leqslant
			\mathcal{R}_{n_s}(b_{OLS}|Y_{s},X_{s})
		\end{equation}
		we have
		\begin{equation}
			\frac{1}{n_s}\left\Vert Y_{s}-X_{s}b^*\right\Vert _{2}^{2} \leqslant
			\frac{\frac{1}{n_t}(\Vert Y_{t}-X_{t}b_{OLS}\Vert_{2})^{2}}
			{1-\sqrt{\epsilon}} + \varsigma
		\end{equation}
		By defining $\Delta=b_{OLS}-b^*$, $Y_{t}-X_{t}b_{OLS}=e_{t}$ and $Y_{s}-X_{s}b_{OLS}=e_{s}$,
		\begin{eqnarray}
		\frac{1}{n_s}(\Vert Y_{s}-X_{s}b^*\Vert_{2})^{2}
		  & = & \frac{1}{n_s}(\Vert Y_{s}-X_{s}b_{OLS}+X_{s}\Delta\Vert_{2})^{2}
		  \notag  \\
		  & = & \frac{1}{n_s}(\Vert e_{s}+X_{s}\Delta\Vert_{2})^{2}
		  \notag     \\
		  & = & \frac{1}{n_s}(e_{s}+X_{s}\Delta)^{T}(e_{s}+X_{s}\Delta)
		  \notag  \\
		  & = & \frac{1}{n_s}(\left\Vert e_{s}\right\Vert_{2}^{2}+2e_{s}^{T}X_{s}\Delta+
			\Delta^{T}X_{s}^{T}X_{s}\Delta)
		\end{eqnarray}
		Hence,
		\begin{equation}
			\frac{1}{n_s} (\Vert Y_{s}-X_{s}b^*\Vert_{2})^{2}
			\leqslant
			\frac{\frac{1}{n_t}(\Vert Y_{t}-X_{t}b_{OLS} \Vert_{2})^{2}}{1-\sqrt{\epsilon}}
			+ \varsigma
		\end{equation}
		implies
		\begin{equation}
			\frac{1}{n_s}(\Vert e_{s} \Vert_{2})^{2}
			+ \frac{2}{n_s}e_{s}^{T}X_{s}\Delta
			+ \frac{1}{n_s}\Delta^{T}X_{s}^{T}X_{s}\Delta
			\leqslant
			\frac{\frac{1}{n_t} (\Vert e_{t} \Vert_{2})^{2}}{1-\sqrt{\epsilon}}
			+ \varsigma.
		\end{equation}
		It follows that
		\begin{equation}
			\frac{1}{n_s} (\Vert X_{s}\Delta \Vert_{2})^{2}
			\leqslant
			\left( \frac{1}{n_t}\frac{ ( \Vert e_{t} \Vert_{2})^{2}}{1-\sqrt{\epsilon}}-
			\frac{1}{n_s} ( \Vert e_{s} \Vert_{2})^{2} \right)-
			\frac{2}{n_s}e_{s}^{T}X_{s}\Delta
			+\varsigma.
		\end{equation}
		By the Holder inequality,
		\begin{equation}
			-e_{s}^{T} X_{s} \Delta
			\leqslant
			\vert e_{s}^{T} X_{s} \Delta \vert
			\leqslant
			\Vert e_{s}^{T} X_{s} \Vert_{\infty}
			\Vert \Delta \Vert_{1}.
		\end{equation}
		It follows that
		\begin{equation}
			\frac{1}{n_s} (\Vert X_{s} \Delta \Vert_{2})^{2}
			\leqslant
			\left( \frac{1}{n_t} \frac{ (\Vert e_{t} \Vert_{2})^{2}}{1-\sqrt{\epsilon}}
			- \frac{1}{n_s} (\Vert e_{s} \Vert_{2})^{2} \right)
			+ \frac{2}{n_s} \Vert e_{s}^{T} X_{s} \Vert_{\infty} \Vert \Delta \Vert _{1}
			+\varsigma.
		\end{equation}
		Also, since $\Vert b^*\Vert_{1} \leqslant \Vert b_{OLS} \Vert_{1}$
		\begin{eqnarray}
			\Vert \Delta \Vert _{1}
			& = & \Vert b_{OLS}-b^* \Vert_{1}
			\notag     \\
			& \leqslant &  \Vert b^* \Vert_{1} + \Vert b_{OLS} \Vert_{1}
			\notag  \\
			& \leqslant & 2 \Vert b_{OLS} \Vert_{1}		
		\end{eqnarray}
		As a result, we have
		\begin{equation}
			\frac{1}{n_s} (\Vert X_{s}\Delta \Vert _{2})^{2}
			\leqslant
			\left( \frac{1}{n_t} \frac{ ( \Vert e_{t} \Vert_{2})^{2}}{1-\sqrt{\epsilon}}
			- \frac{1}{n_s} ( \Vert e_{s} \Vert_{2})^{2} \right)
			+ \frac{4}{n_s} \Vert e_{s}^{T }X_{s} \Vert_{\infty} \Vert b_{OLS} \Vert_{1}
			+\varsigma
		\end{equation}

    \item[\textbf{$K$-fold cross-validation.}] If penalized regression is implemented by $K$-fold cross-validation, then based on Lemma~\ref{lem3.1}, the following bound holds with probability at least $(1-1/K)\varpi$
		\begin{equation}
			\frac{1}{K} \sum_{q=1}^{K} \mathcal{R}_{n_s}(b_{OLS}^q|X_s^q,Y_s^q)
			\leqslant
			\frac{1}{K} \sum_{q=1}^{K}
            \frac{\mathcal{R}_{n_t}(b_{OLS}^q|X_t^q,Y_t^q)}{1-\sqrt{\epsilon}}
			+ \varsigma.
		\end{equation}
        Since $b^*$ minimizes $(1/K)\sum_{q=1}^K\mathcal{R}_{n_s}(b|X_s^q,Y_s^q)$ among $\{b_{\lambda}\}$,
		\begin{eqnarray}
			\frac{1}{K}\sum_{q=1}^K\mathcal{R}_{n_s}(b^{*q}|X_s^q,Y_s^q)
			\leqslant
			\frac{1}{K}\sum_{q=1}^K\mathcal{R}_{n_s}(b_{OLS}^q|X_s^q,Y_s^q),
		\end{eqnarray}
	    it follows that
		\begin{equation}
			\frac{1}{K}\sum_{q=1}^K\mathcal{R}_{n_s}(b^{*q}|X_s^q,Y_s^q)
			\leqslant
			\frac{\mathcal{R}_{n_t}(b_{OLS}^q|X_t^{q},Y_t^{q})}{1-\sqrt{\epsilon}}
			+\varsigma.
		\end{equation}
        By defining $\Delta^q = b_{OLS}^q - b^{*q}$ and $e_s^q = Y_s^q-X_s^q b_{OLS}^q$ we have
		\begin{eqnarray}
			\frac{1}{n_s} (\Vert Y_s^q - X_s^q b^{*q} \Vert_{2})^{2}
		  	& = & \frac{1}{n_s}
            ( \Vert Y_s^q - X_s^q b_{OLS}^q + X_s^q \Delta^q \Vert_{2}) ^{2} \notag \\
		  	& = & \frac{1}{n_s} ( \Vert e_s^q + X_s^q \Delta^q \Vert_{2})^{2} \notag \\
		  	& = & \frac{1}{n_s} ( e_s^q + X_s^q \Delta^q )^{T} (e_s^q + X_s^q \Delta^q) \notag \\
		  	& = & \frac{1}{n_s} \left[ (\Vert e_s^q \Vert_{2})^{2} + 2(e_s^q)^{T} X_s^q \Delta^q +
		  	(\Delta^q)^{T} (X_s^q)^{T} X_s^q \Delta^q \right].
		\end{eqnarray}
	    Hence,
		\begin{equation}
			\frac{1}{K}\sum_{q=1}^K \left( \frac{1}{n_s}
            (\Vert Y_s^q - X_s^q b^* \Vert_{2})^{2}\right)
			\leqslant
			\frac{1}{n_t} \frac{\left\Vert Y_t^{q}-X_t^{q}b_{OLS}^q\right\Vert_{2}^{2}}
			{1-\sqrt{\epsilon}}
			+ \varsigma
		\end{equation}
	    implies
		\begin{eqnarray}
			 \frac{1}{K}\sum_{q=1}^K \frac{1}{n_s} (\Vert e_s^q \Vert_{2})^{2}
			+\frac{1}{K}\sum_{q=1}^K \frac{2}{n_s} (e_s^q)^{T} X_{s} \Delta
			& + &
			\frac{1}{K}\sum_{q=1}^K \frac{1}{n_s}
            (\Delta^q)^{T} (X_s^q)^{T} (X_s^q) \Delta^q \notag \\
			& \leqslant &
			\frac{1}{K}\sum_{q=1}^K \frac{\frac{1}{n_t} (\Vert e_t^q \Vert_{2})^{2}}
			{1-\sqrt{\epsilon}} + \varsigma.
		\end{eqnarray}
	    It follows that
		\begin{eqnarray}
			\frac{1}{K}\sum_{q=1}^K\frac{1}{n_s}\left\Vert X_s^q\Delta\right\Vert_{2}^{2}
			 \leqslant
            &\frac{1}{K}\sum_{q=1}^K \frac{\frac{1}{n_t} (\Vert e_t^q \Vert_{2})^{2}}{1-\sqrt{\epsilon}}
			 -\frac{1}{K}\sum_{q=1}^K\frac{\left\Vert e_s^q \right\Vert_{2}^{2}}{n_s} \notag \\			
			&-\frac{1}{K}\sum_{q=1}^K\frac{2}{n_s}\left(e_s^q\right)^{T}X_{s}^q\Delta^q
			+\varsigma.
		\end{eqnarray}
	    By the Holder inequality,
		\begin{equation}
			-1 (e_s^q)^{T} X_s^q \Delta^q
			\leqslant
			\vert ( e_s^q )^{T} X_s^q \Delta^q \vert
			\leqslant \Vert (e_s^q)^{T} X_s^q \Vert_{\infty} \Vert \Delta^q \Vert_{1}.
		\end{equation}
	    It follows that
		\begin{eqnarray}
			\frac{1}{K}\sum_{q=1}^K \frac{1}{n_s} \left(\Vert X_s^q \Delta \Vert_{2}\right)^{2}
			\leqslant
			& \left\vert \frac{1}{K}\sum_{q=1}^K \frac{\frac{1}{n_t}
            (\Vert e_t^q \Vert_{2})^{2}}{1-\sqrt{\epsilon}}
			- \frac{1}{K}\sum_{q=1}^K \frac{ (\Vert e_s^q \Vert_{2})^{2}}{n_s} \right\vert \notag \\
			& +\frac{1}{K}\sum_{q=1}^K \frac{2}{n_s} \Vert (e_s^q)^{T} X_{s}^q \Vert_{\infty}
			\Vert \Delta^q \Vert_{1}
			+ \varsigma.
		\end{eqnarray}
	    Also, since $\Vert b^*\Vert_{1} \leqslant \Vert b_{OLS} \Vert_{1}$
		\begin{eqnarray}
			\Vert \Delta^q \Vert _{1}
			& = & \Vert b_{OLS}^q -b^{*q} \Vert_{1}   \notag   \\
			& \leqslant & \Vert b^{*q} \Vert_{1} + \Vert b_{OLS}^q \Vert_{1} \notag  \\
			& \leqslant & 2\Vert b_{OLS}^q \Vert_{1}
		\end{eqnarray}
	    Therefore, we have
		\begin{eqnarray}
			\frac{1}{K}\sum_{q=1}^K \frac{1}{n_s} \left(\Vert X_s^q \Delta \Vert_{2}\right)^{2}
			\leqslant
			& \left\vert \frac{1}{K}\sum_{q=1}^K \frac{\frac{1}{n_t}
            (\Vert e_t^q \Vert_{2})^{2}}{1-\sqrt{\epsilon}}
			- \frac{1}{K}\sum_{q=1}^K \frac{ (\Vert e_s^q \Vert_{2})^{2}}{n_s} \right\vert \notag \\
			& +\frac{1}{K}\sum_{q=1}^K \frac{4}{n_s} \Vert (e_s^q)^{T} X_{s}^q \Vert_{\infty}
			\Vert b_{OLS}^q \Vert_{1}
			+ \varsigma.
		\end{eqnarray}

\end{description}
%%%%%%%%%%%%%%%%%%%%%%%%%%%%%%%%%%%%%%%%%%%%%%%%%%%%%%%%%%%%%%%%%%%%%%%%%%%%%%%%%%%%%%%%%%%%%%%%%%%%%%%%%%
	
\end{proof}
\bigskip{}

%%%%%%%%%%%%%%%%%%%%%%%%%%%%%%%%%%%%%%%%%%%%%%%%%%%%%%%%%%%%%%%%%%%%%%%%% Theorem 3.1 proof %%%%%%%%%%%%%%
\begin{proof}
\textbf{Theorem~\ref{thm3.1}}
The proof follows from Proposition~\ref{prop3.3}.

\begin{description}
	\item[\textbf{Validation.}] For OLS, $(1/n)\left\Vert X_{s}\Delta\right
       \Vert _{2}^{2}\geqslant\rho\left\Vert \Delta\right\Vert _{2}^{2}$, where $\rho$ is the minimal eigenvalue for $(X_s)^{T}X_s$. Hence,
	   \begin{eqnarray}
		  \rho (\Vert \Delta \Vert_{2})^{2}
		  & \leqslant &
		  \frac{1}{n_s} (\Vert X_{s} \Delta \Vert _{2})^{2} \notag \\
		  & \leqslant &
            \left\vert \frac{1}{n_t} \frac{ (\Vert e_t \Vert_{2})^{2}}{1-\sqrt{\epsilon}}
            - \frac{ \Vert e_s \Vert_{2}^{2}}{n_s} \right\vert \notag \\
		  &           &
            +  \frac{4}{n_s} \Vert (e_s)^{T} X_{s} \Vert_{\infty} \Vert b_{OLS} \Vert_{1} + \varsigma.
	   \end{eqnarray}
	   By the Minkowski inequality, the above can be simplified to
	   \begin{eqnarray}
		  \left\Vert b_{train}-b_{Lasso}\right\Vert _{2}
		  & \leqslant &		
		  \sqrt{\left\vert \frac{1}{n_t \rho} \frac{ (\Vert e_t \Vert_{2})^{2}}{1-\sqrt{\epsilon}}
		  - \frac{ \Vert e_s \Vert_{2}^{2}}{n_s \rho} \right\vert} \notag \\
		  & & +  \sqrt{\frac{4}{n_s \rho} \Vert (e_s)^{T} X_{s} \Vert_{\infty}\Vert b_{OLS} \Vert_{1}}
		  + \sqrt{\frac{\varsigma}{\rho}}.
	   \end{eqnarray}	
	
	\item[\textbf{$K$-fold cross-validation.}] For the OLS estimate from the $q$th round,
	   $(1/n)\left\Vert X_{s}^q \Delta\right\Vert _{2}^{2}\geqslant\rho
        \left\Vert \Delta\right\Vert _{2}^{2}$,	where $\rho_q$ is the minimal eigenvalue for $(X_s^q)^{T}X_s^q$ in the $q$th round. Hence, if we define the minimum of all the minimal round-by-round eigenvalues from all $K$ rounds,
	   \[
		  {\rho^*} = \min\{\rho_q \vert \forall q \in [1,K] \},
	   \]
	   then
	   \begin{eqnarray}
		  \frac{1}{K}\sum_{q=1}^K {\rho^*} (\Vert \Delta^q \Vert_{2})^{2}
		  & \leqslant &
		  \frac{1}{K}\sum_{q=1}^K \frac{1}{n_s} (\Vert X_{s}^q \Delta^q \Vert _{2})^{2} \notag \\
		  & \leqslant & \left\vert \frac{1}{K}\sum_{q=1}^K \frac{\frac{1}{n_t}
            (\Vert e_t^q \Vert_{2})^{2}}{1-\sqrt{\epsilon}}
		- \frac{1}{K}\sum_{q=1}^K \frac{ (\Vert e_s^q \Vert_{2})^{2}}{n_s} \right\vert \notag \\
		& & +\frac{1}{K}\sum_{q=1}^K \frac{4}{n_s} \Vert (e_s^q)^{T} X_{s}^q \Vert_{\infty}
		\Vert b_{OLS}^q \Vert_{1} + \varsigma.
	   \end{eqnarray}
	   Hence,
	   \begin{eqnarray}
		  \frac{1}{K}\sum_{q=1}^K \left(\left\Vert b_{OLS}^q-b^{*q}\right\Vert _{2}\right)^2
		  & \leqslant &
		  \frac{1}{K}\sum_{q=1}^K \left|\frac{1}{{\rho^*} n_t}
          \frac{\left\Vert e_{t}^q \right\Vert_{2}^{2}}{\left(1-\sqrt{\epsilon}\right)}
		-\frac{1}{K}\sum_{q=1}^K \frac{1}{{\rho^*} n_s}
        \left\Vert e_{s}^q \right\Vert _{2}^{2}\right| \notag \\
		& &
		+\frac{1}{K}\sum_{q=1}^K \frac{4}{{\rho^*} n_s}
        \left\Vert (e_{s}^q)^{T}X_{s} \right\Vert_{\infty}
		\left\Vert b_{OLS}^q\right\Vert _{1} +\frac{\varsigma}{{\rho^*}}
	\end{eqnarray}
	
\end{description}
\end{proof}
%%%%%%%%%%%%%%%%%%%%%%%%%%%%%%%%%%%%%%%%%%%%%%%%%%%%%%%%%%%%%%%%%%%%%%%%%%%%%%%%%%%%%%%%%%%%%%%%%%%%%%%%%%

\bigskip{}
%%%%%%%%%%%%%%%%%%%%%%%%%%%%%%%%%%%%%%%%%%%%%%%%%%%%%%%%%%%%%%%%%%%%%%%%%%% Corollary 3.2 proof %%%%%%%%%%
\begin{proof}
\textbf{Corollary~\ref{cor3.2}}
($\mathcal{L}_2$-consistency of  $b^*$)
	
\begin{description}
	
		\item[\textbf{Validation.}] In Theorem~\ref{thm3.1},
			\begin{eqnarray}
				\left\Vert b_{train}-b_{Lasso}\right\Vert _{2}
				& \leqslant &		
				\sqrt{\left\vert \frac{1}{n_t \rho} \frac{ (\Vert e_t \Vert_{2})^{2}}{1-\sqrt{\epsilon}}
				- \frac{ \Vert e_s \Vert_{2}^{2}}{n_s \rho} \right\vert} \notag \\
				& & +  \sqrt{\frac{4}{n_s \rho} \Vert (e_s)^{T} X_{s} \Vert_{\infty}
				\Vert b_{OLS} \Vert_{1}} + \sqrt{\frac{\varsigma}{\rho}}.
			\end{eqnarray}	
		Since
		\[
			\lim_{\tilde{n}/p \rightarrow \infty}
			\frac{1}{n_t} \frac{(\Vert e_t \Vert_{2})^{2}}{1-\sqrt{\epsilon}}
			=
			\lim_{\tilde{n}/p \rightarrow \infty}
            \frac{(\Vert e_s \Vert_{2})^{2}}{n_s}
			=
			\frac{(\Vert u \Vert_{2})^{2}}{n_t},
		\]
		\[
			\lim_{\tilde{n}/\log(p) \rightarrow \infty}\frac{1}{n_s}\Vert (e_s)^{T} X_{s}\Vert_{\infty}
			=	0
			\mbox{ if } u \sim N(0,\sigma^2),
		\]
		and
		\[
			\lim_{\tilde{n}/p \rightarrow \infty}\varsigma = 0,
		\]
		as a result, $\Vert b^* - \beta\Vert_2 \rightarrow 0$.
		
		\item[\textbf{$K$-fold cross-validation.}] In Theorem~\ref{thm3.1},
			\begin{eqnarray}
				\frac{1}{K}\sum_{q=1}^K (\left\Vert b_{OLS}^q-b^{*q}\right\Vert _{2})^2
				& \leqslant &
				\frac{1}{K}\sum_{q=1}^K
                \left|\frac{1}{{\rho^*} n_t}\frac{\left\Vert e_{t}^q \right\Vert_{2}^{2}}
				{\left(1-\sqrt{\epsilon}\right)}
				-\frac{1}{K}\sum_{q=1}^K \frac{1}{{\rho^*} n_s}
                \left\Vert e_{s}^q \right\Vert _{2}^{2}\right| \notag \\
				& &
				+ \frac{1}{K}\sum_{q=1}^K \frac{4}{{\rho^*} n_s}
                \left\Vert (e_{s}^q)^{T}X_{s} \right\Vert_{\infty}
				\left\Vert b_{train}\right\Vert _{1} + \frac{\varsigma}{{\rho^*}}				
			\end{eqnarray}	
		Since
		\[
			\lim_{\tilde{n}/p \rightarrow \infty}
			\frac{1}{n_t} \frac{(\Vert e_t^q \Vert_{2})^{2}}{1-\sqrt{\epsilon}}
			=
			\lim_{\tilde{n}/p \rightarrow \infty}
			\frac{(\Vert e_s^q \Vert_{2})^{2}}{n_s}
			=
			\frac{(\Vert u \Vert_{2})^{2}}{n_t},
		\]
		\[
			\lim_{\tilde{n}/\log(p) \rightarrow \infty}
			\frac{1}{n_s} \Vert (e_s^q)^{T} X_{s}^q \Vert_{\infty}
			=	0
			~\mbox{if u} \sim N(0,\sigma^2),
		\]
		and
		\[
			\lim_{\tilde{n}/p \rightarrow \infty}\varsigma = 0,
		\]
        as a result, $(1/K)\sum_{q=1}^K \left(\left\Vert b_{OLS}^q-b^{*q}\right\Vert _{2}\right)^2 \rightarrow 0$.
\end{description}
\end{proof}
%%%%%%%%%%%%%%%%%%%%%%%%%%%%%%%%%%%%%%%%%%%%%%%%%%%%%%%%%%%%%%%%%%%%%%%%%%%%%%%%%%%%%%%%%%%%%%%%%%%%%%%%%%

\bigskip{}
%%%%%%%%%%%%%%%%%%%%%%%%%%%%%%%%%%%%%%%%%%%%%%%%%%%%%%%%%%%%%%%%%%%%%%% Proposition 3.4 proof %%%%%%%%%%%%
\begin{proof}
\textbf{Proposition~\ref{prop3.4}}
As shown in the discussion above Proposition~\ref{prop3.4}, while Proposition~\ref{prop3.3} is valid for the $p>n$ case, we cannot derive the $\mathcal{L}_2$-difference between $b_{FSR}$ and $b^*$ because $X\Delta$ is no longer strongly convex. As a result, to derive the upper bound of
$\Vert b_{FSR}-b^*\Vert_2$, we need to use the restricted eigenvalue condition \citep{bickeletal09,meinshausenyu09,zhang09}.

	\begin{description}
    \item[Restricted eigenvalue condition.] For some integer $1\leqslant s \leqslant p$ and a positive number $k_0$, for both FSR and Lasso satisfies the following condition
		\[
			\min_{J_0 \subset \{1,\ldots,p\},\vert J_0 \vert \leqslant s}
			~\min_{\Vert \Delta_{J_0^c} \Vert_1 \leqslant k_0\Vert \Delta_{J_0} \Vert_1}
			\frac{\Vert X\Delta\Vert_2}{\sqrt{n}\Vert \Delta_{J_0}\Vert_2} = \rho_{re} >0
		\]
    where $\Delta_{J_0}$ stands for the difference between two vectors with at most $J_0$ non-zero vectors, and $J_0^c$ is the complement set of $J_0$. Also $J_0$ can be treated as the support of $\Vert \Delta\Vert_0$.
	\end{description}
	
	As a result,
	\begin{description}
	   \item[Validation.] For FSR, $(1/n)\left\Vert X_{s}b_{FSR} - X_{s}b^*\right\Vert_{2}^{2}=
        (1/n)\left\Vert X_{s}\Delta\right\Vert _{2}^{2}\geqslant \rho_{re}
        \left\Vert \Delta\right\Vert _{2}^{2}$,	where $\rho$ is the minimal eigenvalue for $(X_s)^{T}X_s$. Hence, the restricted eigenvalue condition implies
	   \begin{eqnarray}
		  \rho_{re} (\Vert \Delta \Vert_{2})^{2}
		  & \leqslant &
		  \frac{1}{n_s} (\Vert X_{s} \Delta \Vert _{2})^{2}	\notag \\
		  & \leqslant & \left\vert \frac{1}{n_t} \frac{ (\Vert e_t \Vert_{2})^{2}}{1-\sqrt{\epsilon}}
		  - \frac{ \Vert e_s \Vert_{2}^{2}}{n_s} \right\vert \notag \\
		  & & +  \frac{4}{n_s} \Vert (e_s)^{T} X_{s} \Vert_{\infty}\Vert b_{FSR} \Vert_{1} + \varsigma.
	\end{eqnarray}
	By the Minkowski inequality, the above can be simplified to
	\begin{eqnarray}
		\left\Vert b_{train}-b_{Lasso}\right\Vert _{2}
		& \leqslant &		
		\sqrt{\left\vert \frac{1}{n_t \rho_{re}}
		\frac{ (\Vert e_t \Vert_{2})^{2}}{1-\sqrt{\epsilon}}
		- \frac{ \Vert e_s \Vert_{2}^{2}}{n_s \rho_{re}} \right\vert} \notag \\
		& & + \sqrt{\frac{4}{n_s \rho_{re}} \Vert (e_s)^{T} X_{s} \Vert_{\infty}\Vert b_{FSR} \Vert_{1}}
		+ \sqrt{\frac{\varsigma}{\rho_{re}}}.
	\end{eqnarray}	
	
    \item[$K$-fold cross-validation.] For the FSR estimate in $q$th round, the restricted eigenvalue value condition implies that $(1/n)\left\Vert X_{s}^q \Delta\right\Vert_{2}^{2}	\geqslant\rho_{re}^q\left\Vert \Delta\right\Vert _{2}^{2}$,	where $\rho_{re}^q$ is the minimal restricted eigenvalue for $(X_s^q)^{T}X_s^q$ in the $q$th round. Hence, if we define the minimum of all the minimal round-by-round eigenvalues from all $K$ rounds,
	   \[
		  {\rho^*}_{re} = \min\{\rho_{re}^q \vert \forall q \in [1,K] \},
	   \]
	   then
	   \begin{eqnarray}
		  \frac{1}{K}\sum_{q=1}^K {\rho^*}_{re} (\Vert \Delta^q \Vert_{2})^{2}
		  & \leqslant &
		  \frac{1}{K}\sum_{q=1}^K \frac{1}{n_s} (\Vert X_{s}^q \Delta^q \Vert _{2})^{2} \notag \\
		  & \leqslant & \left\vert \frac{1}{K}\sum_{q=1}^K \frac{\frac{1}{n_t}
            (\Vert e_t^q \Vert_{2})^{2}}{1-\sqrt{\epsilon}}
		  - \frac{1}{K}\sum_{q=1}^K \frac{ (\Vert e_s^q \Vert_{2})^{2}}{n_s} \right\vert \notag \\
		  & & +\frac{1}{K}\sum_{q=1}^K \frac{4}{n_s} \Vert (e_s^q)^{T} X_{s}^q \Vert_{\infty}
		  \Vert b_{FSR}^q \Vert_{1}	+ \varsigma.
	   \end{eqnarray}
	   Hence,
	   \begin{eqnarray}
		\frac{1}{K}\sum_{q=1}^K \left(\left\Vert b_{OLS}^q-b^{*q}\right\Vert _{2}\right)^2
		& \leqslant &
		\frac{1}{K}\sum_{q=1}^K \left|\frac{1}{{\rho^*}_{re} n_t}
		\frac{\left\Vert e_{t}^q \right\Vert_{2}^{2}}
		{\left(1-\sqrt{\epsilon}\right)}
		-\frac{1}{K}\sum_{q=1}^K \frac{1}{{\rho^*}_{re} n_s}
		\left\Vert e_{s}^q \right\Vert _{2}^{2}\right| \notag \\
		& &
		+\frac{1}{K}\sum_{q=1}^K \frac{4}{{\rho^*}_{re} n_s}
        \left\Vert(e_{s}^q)^{T}X_{s} \right\Vert_{\infty}
		\left\Vert b_{FSR}^q\right\Vert _{1} +\frac{\varsigma}{{\rho^*}_{re}}
	   \end{eqnarray}	
	\end{description}
\end{proof}
\newpage{}
\doublespacing
%%%%%%%%%%%%%%%%%%%%%%%%%%%%%%%%%%%%%%%%%%%%%%%%%%%%%%%%%%%%%%%%%%%%%%%%%%%%%%%%%%%%%%%%%%%%%%%%%%%%%%%%%%%
%%%%%%%%%%%%%
%%%%%%%%%%%%%            APPENDIX B
%%%%%%%%%%%%%
%%%%%%%%%%%%%%%%%%%%%%%%%%%%%%%%%%%%%%%%%%%%%%%%%%%%%%%%%%%%%%%%%%%%%%%%%%%%%%%%%%%%%%%%%%%%%%%%%%%%%%%%%%%

%%%%%%%%%%%%%%%%%%%%%%%%%%%%%%%%%%%%%%%%%%%%%%%%%%%%%%%%%%%%%%%%%%%%%%%%%%%%%%%%%%%%%%%%%%%%%%%%%%%%%%%%%%%%

%%%%%%%%%%%%%%%%%%%%%%%%%%%%%%%%%%%%%%%%%%%%%%%%%%%%%%%%%%%%%%%%%%%%%%%%%%%%%%%%%%%%%%%%%%%%%%%%%%%%%%%%%%%%

%%%%%%%%%%%%%%%%%%%%%%%%%%%%%%%%%%%%%%%%%%%%%%%%%%%%%%%%%%%%%%%%%%%%%%%%%%%%%%%%%%%%%%%%%%%%%%%%%%%%%%%%%%%%
%%%%%%%%%%%%%%%%%%%%%%%%%%%%%%%%%%%%%%%%%%%%%%%%%%%%%%%%%%%%%%%%%%%%%%%%%%%%%%%%%%%%%%%%%%%%%%%%%%%%%%%%%%%%

\end{document}